\documentclass{article}

\usepackage{microtype}
\usepackage{graphicx}
\usepackage{subcaption}
\usepackage{booktabs} % for professional tables

\usepackage{hyperref}

\graphicspath{ {./images/} }
\usepackage{tabularx}
\usepackage[table]{xcolor}
\usepackage{xcolor}
\usepackage{soul}
\usepackage{tikz}
\usetikzlibrary{arrows.meta,positioning}
\usepackage{algorithmic}
\renewcommand{\algorithmicrequire}{\textbf{Input:}}
\renewcommand{\algorithmicensure}{\textbf{Output:}}
\renewcommand{\algorithmiccomment}[1]{\hfill{\color{gray}$\triangleright$~#1}}
\providecommand{\RETURN}{\STATE \textbf{return} }
\usepackage[accepted]{icml2026}

\usepackage{amsmath}
\usepackage{amssymb}
\usepackage{mathtools}
\usepackage{amsthm}

\allowdisplaybreaks
\usepackage[capitalize,noabbrev]{cleveref}

\theoremstyle{plain}

\theoremstyle{definition}

\theoremstyle{remark}

\usepackage[textsize=tiny]{todonotes}

\icmltitlerunning{Robust Filter Attention}

\begin{document}

\twocolumn[
\icmltitle{Robust Filter Attention: Self-Attention as Precision-Weighted State Estimation}

\begin{icmlauthorlist}
\icmlauthor{Peter Racioppo}{ind}
\end{icmlauthorlist}

\icmlaffiliation{ind}{Independent Researcher, Los Angeles, CA, USA}

\icmlcorrespondingauthor{Peter Racioppo}{pcracioppo@gmail.com}

\icmlkeywords{Transformers, Attention Mechanisms, Robust Estimation, Stochastic Dynamical Systems, Long-Context Modeling}

\vskip 0.3in

]

\printAffiliationsAndNotice{} 

\begin{abstract}
We introduce Robust Filter Attention (RFA), a formulation of self-attention as a robust state estimator. Each token is treated as a noisy observation of a latent trajectory governed by a linear stochastic differential equation (SDE), and attention weights are determined by consistency under this model rather than static feature similarity. Under isotropic noise and decay assumptions, RFA matches the computational complexity of standard attention. On language modeling benchmarks, RFA matches or improves on RoPE's perplexity within the training window while remaining stable under zero-shot extrapolation to longer contexts, where RoPE degrades sharply. The framework also provides a dynamical interpretation of standard positional mechanisms, connecting rotational embeddings and recency biases to transport and uncertainty propagation induced by dynamics.
\end{abstract}

\section{Introduction}
\label{sec:intro}

Self-attention has become the dominant paradigm for sequence modeling due to its parallelism and scalability \citep{vaswani2023attentionneed}. Unlike recurrent architectures \citep{Elman_RNNs}, however, it does not explicitly propagate latent states through shared temporal dynamics — each token attends independently to all others, with no constraint that states evolve consistently over time. Temporal structure is therefore encoded through positional mechanisms rather than state evolution.

We introduce Robust Filter Attention (RFA), which grounds such structure in an
explicit dynamical model: past tokens are propagated to the query position under
learned linear dynamics, and attention weights are determined by their predicted
reliability under this model, coupling transport and uncertainty propagation. 

In RFA, each token is modeled as a noisy observation of a latent trajectory
governed by a linear SDE. The query token serves as a reference observation of the current state, while past keys are propagated to the query position, each yielding a prediction of the latent state with a precision given by the Differential Lyapunov Equation (DLE). Attention weights
are then computed from prediction errors measured under these precisions, so
that tokens are weighted according to their consistency with the dynamical
model. Depending on the learned balance between process and measurement
uncertainty, different heads may specialize into distinct temporal filtering
behaviors, ranging from short-range recency-biased filtering to stable long-range
integration.

Under a purely Gaussian model, the Kalman filter \citep{originalkalmanfilter}
would be optimal, and a compressed state would summarize the history. RFA
instead allows the forcing realized over each interval to depart from its
expected magnitude, so the reliability of every transported prediction must be
re-evaluated at each query, which a recursion cannot do once an observation has
been absorbed. RFA therefore performs estimation independently at each query
position, yielding a parallel batch estimator. The dynamical model governs the
propagation of information and uncertainty, while attention retains direct
access to historical observations.

Common positional encodings can be understood as imposing implicit dynamical models on how information evolves across tokens. RoPE \citep{su2023roformerenhancedtransformerrotary} encodes transport through norm-preserving rotations but assigns no explicit reliability to tokens as a function of distance. ALiBi \citep{press2022trainshorttestlong} imposes a distance-based reliability penalty but carries no corresponding transport operator. In RFA, transport and reliability are not independent design choices but consequences of a shared dynamical model. RFA under noiseless, decay-free dynamics recovers RoPE, while pure diffusion with no decay yields a logarithmic distance penalty analogous to ALiBi.

Finally, we introduce Spectrally-Coupled RFA (SC-RFA). Standard rotational encodings such as RoPE apply no decay, so high-frequency components persist indefinitely. As a result, tokens far apart in the sequence can appear positionally similar to nearby ones, degrading long-context discrimination. SC-RFA addresses this by partitioning the frequency spectrum across attention heads and coupling each head's decay rate to its maximum frequency, so that high-frequency heads act as short-range filters while low-frequency heads behave as stable long-range integrators.

Our contributions are as follows:

\textbf{(i)} A derivation of self-attention as a tractable robust batch estimator for a linear SDE.

\textbf{(ii)} A scalable isotropic formulation with analytic uncertainty modeling at standard attention cost.

\textbf{(iii)} A recovery of RoPE and ALiBi as limiting cases, corresponding respectively to noiseless, zero-decay dynamics and Brownian diffusion.

\textbf{(iv)} A spectrally-coupled decay prior enabling multi-resolution temporal filtering across heads, with principled suppression of long-range phase interference.

\section{Related Work}
\label{sec:RelatedWork}

\subsection{Probabilistic and Kernel Views of Attention}

The Transformer architecture \citep{vaswani2023attentionneed} computes attention scores via a scaled dot-product between queries and keys. While originally motivated by its efficiency and ability to capture long-range dependencies, a growing body of work has sought to interpret these weights as probabilities derived from latent statistical models.

Probabilistic attention has been connected to mixture-model inference
\citep{gabbur2021probabilistic,nguyen2022improving}, energy-based associative
retrieval \citep{ramsauer2021hopfieldnetworksneed}, positional priors
\citep{bianchessi2026bayesian}, and correlated Gaussian process inference
\citep{bui2025revisitingkernelattentioncorrelated}. These approaches generally
use static similarity or fixed covariance structure, rather than uncertainty
propagated by a learned dynamical model.

Other work has interpreted attention through the lens of kernel regression, identifying softmax attention with the classical Nadaraya-Watson estimator \citep{tsai2019transformerdissectionunifiedunderstanding}.  Subsequent work has explored alternative kernels and weighting schemes to improve robustness, expressivity, or statistical efficiency \citep{han2023designingrobusttransformersusing,liu2020kalmanfilteringattentionuser}. Local Linear Attention \citep{zuo2026local} replaces the local
constant fit underlying softmax attention with a local linear one, improving the
bias--variance trade-off within the test-time regression view.  

These approaches treat attention as a kernel smoother whose behavior is determined by the choice of kernel. In contrast, RFA begins from a latent dynamical state estimation problem. The resulting attention kernel is not specified a priori, but emerges from uncertainty propagation and consistency under a stochastic dynamical model.

\subsection{Filtering, Continuous Dynamics, and SSMs}

Continuous-time sequence models parameterize latent dynamics using Neural ODEs
\citep{chen2019neuralordinarydifferentialequations} and their stochastic
extensions \citep{li2020scalablegradientsstochasticdifferential,
shen2025neuralsdesunifiedapproach}. Other work integrates neural networks with classical filtering frameworks by learning components of the Kalman filter, such as gains, noise models, or update rules \citep{jahanshahi2026,Revach_2022,Liu_2023,Cohen_2025,shen2025kalmanformer}. Recent work has also shown that Transformer architectures are expressive enough to approximate a Kalman filter \citep{goel2024transformerrepresentkalmanfilter}. RFA approaches the connection from the opposite direction: rather than showing that attention can represent a filter, we formulate self-attention itself as a batch state estimator.

State space models (SSMs) provide another approach to sequence modeling, using
structured linear recurrences that can be evaluated in parallel across the
sequence. Frameworks such as HiPPO \citep{gu2020hipporecurrentmemoryoptimal} and S4 \citep{gu2022efficientlymodelinglongsequences} achieve efficiency by restricting the dynamics to structured forms (e.g., diagonalizable or diagonal-plus-low-rank), reducing the cost of state updates from $\mathcal{O}(d^2)$ to $\mathcal{O}(d)$ and enabling efficient parallel implementations. Recent work shows that causal linear attention can be viewed as a special case of LTI convolution \citep{dao2024transformersssmsgeneralizedmodels}.

Concurrent with our work, Kalman Linear Attention (KLA) \citep{shaj2026kalmanlinearattentionparallel} develops a probabilistic filtering interpretation for linear attention and state-space sequence models using parallelized information-form Kalman updates. Whereas KLA and recurrent SSMs propagate a compressed hidden state through recursive updates, RFA instead formulates self-attention as a parallel precision-weighted batch estimation problem, where pairwise uncertainties are analytically derived from the dynamical model rather than recursively propagated through posterior state updates, while preserving explicit content-based token interactions through attention.

% Like S4 \citep{gu2022efficientlymodelinglongsequences},RFA relies on diagonalizable LTI dynamics. However, whereas deterministic SSMs propagate only the state mean, RFA propagates second-order statistics via the DLE, yielding a time-dependent precision kernel that defines a prior over attention weights. While modern SSMs such as Mamba \citep{gu2024mambalineartimesequencemodeling} achieve context sensitivity through data-dependent gating of the recurrent state, RFA preserves content-based routing through attention and regularizes it using a prior on predicted uncertainty.

\subsection{Positional Encodings}

Modeling relative temporal structure in Transformers has been approached with several geometric methods. RoPE \citep{su2023roformerenhancedtransformerrotary} encodes relative position through deterministic complex rotations of queries and keys, but introduces no explicit notion of decay or uncertainty. ALiBi \citep{press2022trainshorttestlong} applies a linear distance-based bias
to attention logits, improving length extrapolation by suppressing distant
interactions, and KERPLE \citep{chi2022kerple} generalizes this to learnable
power-law and logarithmic biases. xPos \citep{sun2022lengthextrapolatabletransformer, sun2023retentivenetworksuccessortransformer} generalizes RoPE by combining rotations with dimension-wise decay to stabilize long-range behavior.

These methods impose useful geometric or monotonic structure, but are not derived from an explicit model of latent state evolution and measurement. As a result, transport and reliability are treated as separate design choices. Each method can be interpreted as corresponding to an implicit dynamical assumption. RoPE implements deterministic phase transport without uncertainty accumulation, corresponding to a noiseless LTI system with zero decay. ALiBi’s linear distance penalty approximates RFA’s DLE-derived bias at short lags under pure Brownian diffusion, giving it an implicit noise model but no transport operator. xPos introduces decay without deriving it from a covariance model, leaving transport and reliability decoupled.

RFA derives both the transport operator and the precision weighting from the same underlying dynamical model, so that these are coupled by construction rather than introduced independently. This also implies a rotate--aggregate--counter-rotate structure on the value stream, required for aggregation to correspond to fusion of state estimates in a shared temporal frame. While methods such as RoPER \citep{harik2022roper} have applied value rotations previously, RFA derives this structure as a necessity of the dynamical model rather than a geometric heuristic.

Methods such as YaRN \citep{peng2023yarnefficientcontextwindow} address RoPE's extrapolation failure by compressing position indices at inference, mitigating out-of-distribution phase rotations post-hoc while introducing a train-inference mismatch. SC-RFA instead prevents the failure by coupling each head's decay rate to its maximum frequency during training, so that high-frequency modes attenuate before accumulating spurious phase matches. Selective RoPE \citep{movahedi2026selective} also combines rotation with decay via a spectral leakage analogy, but lacks a precision kernel, value rotation structure, and principled frequency-decay coupling — components that in RFA are necessary consequences of the dynamical model rather than design choices.

\section{Methods}
\label{sec:method}

RFA models each token as a noisy observation of a latent trajectory governed by
a linear SDE. Keys are transported to the query position under the
deterministic part of the dynamics, while the Differential Lyapunov Equation
provides an analytic estimate of the uncertainty accumulated during transport.
Attention weights are then determined by consistency under this
uncertainty model, yielding a robust precision-weighted state estimator. Under
isotropic noise and decay assumptions, the uncertainty model reduces to a scalar
function of temporal lag, recovering standard $\mathcal{O}(N^2 d)$ attention
complexity. Appendix~\ref{sec:Derivation} provides additional details on the
derivation of the estimator, and Appendix~\ref{sec:RFA_Mechanism} on its
formulation as an efficient attention mechanism.

\subsection{Attention as Batch State Estimation}
\label{sec:Setup}

\paragraph{Dynamical model.}

We model each token embedding $\boldsymbol z_i\in\mathbb R^d$ as a noisy
observation of a latent state evolving under a linear SDE:
\begin{equation}
\label{eq:SDE}
\begin{aligned}
d\boldsymbol x(t) &= \boldsymbol A\boldsymbol x(t)\,dt
    + \boldsymbol G\,d\boldsymbol W(t),\\
\boldsymbol z_i &= \boldsymbol C\bigl(\boldsymbol x(t_i)
    + \boldsymbol\varepsilon_i\bigr),
\end{aligned}
\end{equation}
where $\boldsymbol W$ is a Wiener process,
$\boldsymbol Q=\boldsymbol G\boldsymbol G^\top$,
$\boldsymbol\varepsilon_i\sim\mathcal N(\boldsymbol 0,\boldsymbol R)$,
$\boldsymbol A$ is Hurwitz, and $\boldsymbol C$ is invertible. Over an
interval $[t_j,t_i]$, with $\Delta t_{ij}=t_i-t_j$, the state evolves as
\[
\boldsymbol x(t_i)=e^{\boldsymbol A\Delta t_{ij}}\boldsymbol x(t_j)+\boldsymbol F_{ij},
\qquad
\boldsymbol F_{ij}\sim\mathcal N\big(\boldsymbol 0,\boldsymbol V(\Delta t_{ij})\big),
\]
where the accumulated forcing $\boldsymbol F_{ij}$ is independent of
$\boldsymbol x(t_j)$ and $\boldsymbol V$ solves the Differential Lyapunov
Equation (DLE)
\begin{equation}
\label{eq:DLE}
\dot{\boldsymbol V}(s)=\boldsymbol A\boldsymbol V(s)+\boldsymbol V(s)\boldsymbol A^\top+\boldsymbol Q,
\qquad
\boldsymbol V(0)=\boldsymbol 0 .
\end{equation}

\paragraph{Transported measurements.}
Our goal is to estimate $\boldsymbol x_i:=\boldsymbol x(t_i)$ from
$\{\boldsymbol z_j\}_{j\le i}$. We know nothing about the current state in
advance, so we give it a flat prior \citep{durbin_koopman_2012time}, and its
estimation is maximum likelihood. Each token gives an estimate
$\hat{\boldsymbol x}_j:=\boldsymbol C^{-1}\boldsymbol z_j
=\boldsymbol x_j+\boldsymbol\varepsilon_j$ of its own state, which we
transport to position $i$:
\[
\hat{\boldsymbol x}_{ij}:=e^{\boldsymbol A\Delta t_{ij}}\hat{\boldsymbol x}_j
=\boldsymbol x_i-\boldsymbol F_{ij}+e^{\boldsymbol A\Delta t_{ij}}\boldsymbol\varepsilon_j .
\]
Under the flat prior, $\boldsymbol x_i$ carries no information about the
forcing, so each transported estimate is an unbiased measurement of the
current state,
$\hat{\boldsymbol x}_{ij}\mid\boldsymbol x_i\sim\mathcal N(\boldsymbol x_i,\boldsymbol\Sigma_{ij})$, with
\begin{equation}
\label{eq:measurement}
\boldsymbol\Sigma_{ij}
=\underbrace{\boldsymbol V(\Delta t_{ij})}_{\text{forcing}}
+\underbrace{e^{\boldsymbol A\Delta t_{ij}}\boldsymbol R e^{\boldsymbol A^\top\Delta t_{ij}}}_{\text{transported measurement noise}}
+\gamma^2\boldsymbol I ,
\end{equation}
where $\gamma^2\boldsymbol I$ is an isotropic noise floor added for
stability. The covariance $\boldsymbol\Sigma_{ij}$ specifies how reliable
each measurement is as a function of lag.

\paragraph{Latent scales.}

The Gaussian model is a nominal description: the forcing realized over an
interval may exceed its budget $\boldsymbol V$, and individual observations
may be atypical. Rather than modeling the forcing further, we give each
measurement its own latent scale,
\[
\hat{\boldsymbol x}_{ij}\mid\boldsymbol x_i,s_{ij}
\sim\mathcal N\big(\boldsymbol x_i,\,s_{ij}\boldsymbol\Sigma_{ij}\big),
\]
where $s_{ij}\sim\mathrm{Inv\text{-}Gamma}(\nu/2,\nu/2)$, drawn independently for each pair, so that $s_{ij}$ measures the reliability of that measurement. The scales are a robustness device on each pairwise
comparison rather than a model of the forcing process. Under the nominal
model alone, the filtered state would be a sufficient statistic and the
Kalman filter optimal. With unobserved scales, it is not, since a recursion
cannot re-evaluate an observation once it has been absorbed into the state.
We therefore estimate independently at each query position, yielding a
parallel batch estimator.

\paragraph{Robust state estimation.}

The measurements are correlated through shared forcing, so the exact joint
likelihood requires their full covariance. To obtain a tractable
attention-style estimator, we replace it with a composite likelihood
\citep{lindsay1988composite},
\[
p\big(\{\hat{\boldsymbol x}_{ij}\}_{j\le i}\mid\boldsymbol x_i\big)
\approx\prod_{j\le i}p(\hat{\boldsymbol x}_{ij}\mid\boldsymbol x_i).
\]
Each factor is exact, but the product treats the measurements as independent,
overstating the accumulated precision by an amount that grows with context
length (Appendix~\ref{sec:appendix_robust_estimation}). Define the Mahalanobis distance $d_{ij}^2(\boldsymbol x)
=(\boldsymbol x-\hat{\boldsymbol x}_{ij})^\top\boldsymbol P_{ij}(\boldsymbol x-\hat{\boldsymbol x}_{ij})$, where $\boldsymbol P_{ij}=\boldsymbol\Sigma_{ij}^{-1}$. Conditional on the scales $\boldsymbol s_i:=\{s_{ij}\}_{j\le i}$, the
negative log composite likelihood is the precision-weighted quadratic
\begin{equation}
\label{eq:conditional_objective}
\mathcal L(\boldsymbol x\mid\boldsymbol s_i)
=\tfrac12\sum_{j\le i}s_{ij}^{-1}\,d_{ij}^2(\boldsymbol x).
\end{equation}
Marginalizing the
scales makes each measurement Student-$t$, giving
\begin{equation}
\label{eq:marginal_objective}
\mathcal L(\boldsymbol x)
=\frac{\nu+d}{2}\sum_{j\le i}\log\!\Big(1+\frac{d_{ij}^2(\boldsymbol x)}{\nu}\Big).
\end{equation}
Setting the gradient to zero gives
\[
\sum_{j\le i}w_{ij}\boldsymbol P_{ij}(\boldsymbol x-\hat{\boldsymbol x}_{ij})=\boldsymbol 0,
\qquad
w_{ij}=\frac{\nu+d}{\nu+d_{ij}^2(\boldsymbol x)},
\]
and hence the robust precision-weighted estimator
\begin{equation}
\label{eq:precision_weighted_average}
\bar{\boldsymbol x}_i
=\Big(\sum_{j\le i}w_{ij}\boldsymbol P_{ij}\Big)^{-1}
\sum_{j\le i}w_{ij}\boldsymbol P_{ij}\hat{\boldsymbol x}_{ij}.
\end{equation}

The weights are themselves evaluated at the solution, so
Eq.~\ref{eq:precision_weighted_average} is implicit. It is solved by
expectation--maximization over the scales \citep{Lange1989RobustSM}. Given a
current estimate $\boldsymbol x^{(k)}$, the E-step computes the posterior mean
inverse scale $w_{ij}=\mathbb E[s_{ij}^{-1}\mid d_{ij}^2(\boldsymbol x^{(k)})]$,
which by conjugacy is exactly the weight above, and the M-step evaluates
Eq.~\ref{eq:precision_weighted_average} with these weights held fixed. A
single attention layer performs one such iteration, evaluated at the query
$\hat{\boldsymbol x}_i=\boldsymbol C^{-1}\boldsymbol z_i$, so the weights
depend on the residual
$\boldsymbol r_{ij}:=\hat{\boldsymbol x}_i-\hat{\boldsymbol x}_{ij}$ through
\begin{equation}
\label{eq:mahalanobis_distance}
d_{ij}^2=\boldsymbol r_{ij}^\top\boldsymbol P_{ij}\boldsymbol r_{ij},
\end{equation}
which replaces dot-product similarity with a consistency test under the
uncertainty predicted by the dynamical model. Appendix~\ref{sec:EM}
interprets depth as successive iterations.

Since this weight tends to $1$ as $\nu\to\infty$, we raise it to an exponent
$\kappa$ that controls how sharply influence falls off,
\begin{equation}
\label{eq:robust_weights}
w_{ij}\propto\Big(1+d_{ij}^2/\nu\Big)^{-\kappa},
\qquad
\kappa=\frac{\nu+d}{d}.
\end{equation}
In the isotropic case, $w_{ij}P_{ij}$ is then the Student-$t$ likelihood at
temperature $d/2$, normalizing the logits by $d$ as in standard attention,
and recovering the exponential kernel $\exp(-d_{ij}^2/d)$ as $\nu\to\infty$
(Appendix~\ref{sec:appendix_robust_estimation}). The weight is generated by a
bounded, redescending penalty, so the estimator is equivalently an
M-estimator \citep{huber1964},
\[
\bar{\boldsymbol x}_i=\arg\min_{\boldsymbol x}\sum_{j\le i}\rho_\kappa\big(d_{ij}^2\big),
\qquad
w=\rho_\kappa'.
\]

\paragraph{Closed-form parallel computation.}

To evaluate $\boldsymbol{P}_{ij}$ in closed form for all token pairs
simultaneously, we require that the system matrices be simultaneously
diagonalizable by some $\boldsymbol{S} \in \mathbb{C}^{d \times d}$:
$\boldsymbol{A} = \boldsymbol{S}\boldsymbol{\Lambda}\boldsymbol{S}^{-1}, \,
\boldsymbol{Q} = \boldsymbol{S}\boldsymbol{\Lambda}_{Q}\boldsymbol{S}^{\dagger}, \,
\boldsymbol{R} = \boldsymbol{S}\boldsymbol{\Lambda}_{R}\boldsymbol{S}^{\dagger}.$
Under this assumption, the DLE decouples into independent scalar ODEs for each
eigenmode $\boldsymbol{\Sigma}_{ij} = \boldsymbol{S}
\boldsymbol{\Lambda}_{\Sigma, ij} \boldsymbol{S}^\dagger$, yielding a diagonal
covariance,
\begin{align*}
\boldsymbol{\Lambda}_{\Sigma,ij}
&=
 \boldsymbol{\Lambda}_{Q} 
\frac{\boldsymbol{1} - e^{2\,\mathrm{Re}(\boldsymbol{\Lambda})\,\Delta t_{ij}}}
{-2\,\mathrm{Re}(\boldsymbol{\Lambda})}
+ \boldsymbol{\Lambda}_{R}  
e^{2\,\mathrm{Re}(\boldsymbol{\Lambda})\,\Delta t_{ij}}
+ \gamma^2 \boldsymbol{I},
\end{align*}
and the precision is obtained by diagonal inversion,
$\boldsymbol{\Lambda}_{P,ij} = \boldsymbol{\Lambda}_{\Sigma,ij}^{-1}$. The Mahalanobis distance then decomposes into independent scalar contributions
per mode. Letting
$\hat{\boldsymbol{x}}_{s,j} = \boldsymbol{S}^{-1}\boldsymbol{C}^{-1}
\boldsymbol{z}_j$ and
$\hat{\boldsymbol{x}}_{s,ij} = e^{\boldsymbol{\Lambda}\Delta t_{ij}}
\hat{\boldsymbol{x}}_{s,j}$, the residual is
\[
\boldsymbol r_{s,ij}
=
\hat{\boldsymbol{x}}_{s,i}-\hat{\boldsymbol{x}}_{s,ij},
\qquad
d_{ij}^2
=
\sum_{k=1}^{d}
\lambda_{P,ij,k}
\big|r_{s,ij,k}\big|^2 ,
\]
where $\lambda_{P,ij,k}$ and $r_{s,ij,k}$ denote the $k$th
entries of $\boldsymbol{\Lambda}_{P,ij}$ and
$\boldsymbol r_{s,ij}$. The influence weights $w_{ij}=w(d_{ij}^2)$ can
therefore be evaluated for all token pairs directly in this basis, and the
robust precision-weighted aggregation takes the form
\begin{equation}
\label{eq:estimator}
\begin{aligned}
\bar{\boldsymbol{x}}_{s,i}
&=
\sum_{j \le i}
\mathcal{A}_{ij}
\odot
\hat{\boldsymbol{x}}_{s,ij},
\\
\mathcal{A}_{ij}
&:=
\Big(
\sum_{j' \le i}
w_{ij'}
\boldsymbol{\Lambda}_{P,ij'}
\Big)^{-1}
w_{ij}
\boldsymbol{\Lambda}_{P,ij}.
\end{aligned}
\end{equation}

Hence, the estimate can be computed cheaply in the eigenbasis. We can then
recover an estimated measurement via
$\bar{\boldsymbol{z}}_{i} = \boldsymbol{C}\boldsymbol{S}
\bar{\boldsymbol{x}}_{s,i}$. When $\mu=0$, the pairwise transport
$e^{\boldsymbol{\Lambda}\Delta t_{ij}}$ reduces to element-wise rotations, as in
RoPE. The estimator in Eq.~\ref{eq:estimator} can be viewed as a heteroskedastic extension of the Nadaraya--Watson estimator, in which the kernel width and weight are both functions of the temporal lag, and distances are measured after dynamical transport.

\subsection{Robust Filter Attention Mechanism}
\label{sec:RFA_main}

We instantiate the robust state estimator as a complex-valued attention layer by
identifying the diagonalization matrices with learned linear projections. The input projections
$\boldsymbol{W}_q, \boldsymbol{W}_k, \boldsymbol{W}_v \in \mathbb{C}^{d \times d}$
absorb $\boldsymbol{S}^{-1}\boldsymbol{C}^{-1}$, mapping into the eigenbasis,
while $\boldsymbol{W}_o$ absorbs $\boldsymbol{C}\boldsymbol{S}$, mapping the
filtered estimates back:
\[
\boldsymbol{Q} = \boldsymbol{W}_q \,\boldsymbol{Z}, \quad
\boldsymbol{K} = \boldsymbol{W}_k \,\boldsymbol{Z}, \quad
\boldsymbol{V} = \boldsymbol{W}_v \,\boldsymbol{Z} \quad
\in \mathbb{C}^{d \times N} .
\]
Under a single observation
model, the three projections coincide; allowing them to be learned separately
relaxes this. Separate query/key projections can be justified as arising from separate observation models (Appendix~\ref{sec:Observation_Models}).

To preserve the $\mathcal{O}(N^2 + Nd)$ memory complexity of standard attention, we impose isotropic decay and noise in the learned eigenbasis (per head):
\[
\boldsymbol{\Lambda} = -\mu \boldsymbol{I} + i \boldsymbol{\Lambda}_{\Omega} , \, \, \,
\boldsymbol{\Lambda}_{Q} = \sigma^{2} \boldsymbol{I}, \, \, \,
\boldsymbol{\Lambda}_{R} = \eta^{2} \boldsymbol{I},
\]
where $\mu, \sigma^2, \eta^2 \in \mathbb{R}^+$ and $\boldsymbol{\Lambda}_{\Omega} \in \mathbb{R}^{d \times d}$ is diagonal with $k$th diagonal entry $\omega_k$. This removes the ability to model dimension-dependent noise, but preserves the temporal dependence of uncertainty on lag.

Under isotropic decay and noise, each eigenmode follows independent exponentially decaying rotations with decay rate $\mu$ and angular frequency $\omega_k$. This yields simple element-wise rotation factors for forward/backward propagation, and a decay kernel that depends only on the time lag $\Delta t_{ij}$:
\[
\tilde{\boldsymbol{\Phi}}^{-}[k,i] := e^{-i \omega_k t_i} , \,  \tilde{\boldsymbol{\Phi}}^{+}[k,i] := e^{i \omega_k t_i}, \, 
\boldsymbol{E}[i,j] := e^{-\mu \Delta t_{ij}}.
\]
We define rotated queries, keys, and values:
\[
\tilde{\boldsymbol{Q}}
:=
\tilde{\boldsymbol{\Phi}}^{-} \odot \boldsymbol{Q}, \quad
\tilde{\boldsymbol{K}}
:=
\tilde{\boldsymbol{\Phi}}^{-} \odot \boldsymbol{K}, \quad
\tilde{\boldsymbol{V}}
:=
\tilde{\boldsymbol{\Phi}}^{-} \odot \boldsymbol{V}
.
\]
The isotropic constraints cause the variance to become independent of the feature dimension:
\begin{equation}
\label{eq:isotropic_covariance}
\begin{aligned}
\boldsymbol{\Sigma}_{\Delta t}[i,j] := \tilde{\sigma}^2 \big( 1 - \boldsymbol{E}^2[i,j] \big) + \eta^{2} \boldsymbol{E}^2[i,j] + \gamma^2.
\end{aligned}
\end{equation}
Here, $\tilde{\sigma}^2 := \frac{\sigma^{2}}{2\mu}$, $\eta^2$, $\gamma^2$ are learned scalar parameters (per head), corresponding, respectively, to steady-state process uncertainty, measurement noise, and a noise floor ($\gamma^2$ can alternatively be interpreted as query-side measurement noise). Note that $\boldsymbol{\Sigma}_{\Delta t}$ denotes the scalar variance as a function of lag rather than a covariance matrix. Collecting terms,
\[
\boldsymbol{\Sigma}_{\Delta t}[i,j]
=
\alpha e^{-2\mu \Delta t_{ij}} + \beta,
\quad
\alpha := \eta^2 - \tilde{\sigma}^2,
\quad
\beta := \gamma^2 + \tilde{\sigma}^2.
\]
The behavior of the variance with temporal lag is governed by the sign of $\alpha$. When $\alpha < 0$, process noise dominates and precision decreases monotonically, yielding a \emph{diffusive regime}. When $\alpha > 0$, measurement noise dominates and precision increases with lag, corresponding to an \emph{integrative regime}. Because $\alpha$ and $\beta$ are determined by learned scalar parameters per head, different heads can specialize into distinct filtering behaviors.

Letting $\boldsymbol{P}_{\Delta t}[i,j] := \boldsymbol{\Sigma}^{-1}_{\Delta t}[i,j]$, the isotropic constraint allows the Mahalanobis distance for all pairs $(i,j)$ to be computed by element-wise multiplying a matrix of scalar precisions by a matrix of squared residual norms $\|\boldsymbol{R}_{qk}[i,j]\|^2$:
\begin{equation*}
\begin{aligned}
\boldsymbol{D}^2[i,j] = \boldsymbol{P}_{\Delta t}[i,j] \cdot \big\| \boldsymbol{R}_{qk}[i,j] \big\|^{2},
\end{aligned}
\end{equation*}
where the $ij$th residual is:
\[
\boldsymbol{R}_{qk}[i,j] := \tilde{\boldsymbol{Q}}_i - \boldsymbol{E}[i,j] \cdot \tilde{\boldsymbol{K}}_j ,
\]
which decomposes as
% \begin{equation}
% \label{eq:residual_norm}
% \begin{split}
% \big\|\boldsymbol{R}_{qk}[i,j]\big\|^{2} 
% & := \sum_{k} \Bigl(
%     \bigl|\boldsymbol{Q}[k,i]\bigr|^{2} \\ 
% &\quad + \boldsymbol{E}[i,j]^{2}\, \bigl|\boldsymbol{K}[k,j]\bigr|^{2} \\
%     &- 2\, \boldsymbol{E}[i,j]\, 
%       \mathrm{Re}\bigl(\tilde{\boldsymbol{Q}}^{*}[k,i]\, \tilde{\boldsymbol{K}}[k,j]\bigr)
% \Bigr).
% \end{split}
% \end{equation}
\begin{equation}
\label{eq:residual_norm}
\begin{aligned}
\big\|\boldsymbol{R}_{qk}[i,j]\big\|^{2}
&=
\|\boldsymbol{Q}_{i}\|^{2}
+
\boldsymbol{E}[i,j]^{2} \cdot\|\boldsymbol{K}_{j}\|^{2} \\
&\quad
- 2\,\boldsymbol{E}[i,j] \cdot 
\mathrm{Re}\!\left(\tilde{\boldsymbol{Q}}_{i}^{\dagger}\tilde{\boldsymbol{K}}_{j}\right).
\end{aligned}
\end{equation}
Setting
\(
\kappa=\frac{\nu+d}{d},
\,
\nu=\nu_s d,
\) with
\(
\nu_s\in\mathbb R^+
\)
yields dimension-normalized attention logits. Choosing the exponential kernel with scale $\nu_s$ yields logits
\[
\boldsymbol L
=
\log(\boldsymbol P_{\Delta t})
-
\frac{1}{\nu_s d}
\boldsymbol P_{\Delta t}
\odot
\|\boldsymbol R_{qk}\|^2,
\]
while the power-law form yields
\[
\boldsymbol L
=
\log(\boldsymbol P_{\Delta t})
-
(\nu_s+1)
\log\! \Big(
1+
\frac{1}{\nu_s d}
\boldsymbol P_{\Delta t}
\odot
\|\boldsymbol R_{qk}\|^2
\Big).
\]
The attention matrix is then $\hat{\boldsymbol{A}} = \boldsymbol{A} \odot \boldsymbol{E} $, where:
\[
\boldsymbol{A} = \mathrm{softmax}_{j} \bigl(\beta_s \boldsymbol{L} + \boldsymbol{M}_{\text{causal}}\bigr),
\]
where $\beta_s$ is an additional inverse temperature parameter and $\boldsymbol{M}_{\text{causal}} \in \{0,-\infty\}^{N \times N}$ is a causal mask.

The value estimate is computed by aggregating the rotated values and then rotating back into the original value frame:
\[
\bar{\boldsymbol{V}}
=
\tilde{\boldsymbol{\Phi}}^{+}
\odot
\bigl(
    \tilde{\boldsymbol{V}} \,
    \hat{\boldsymbol{A}}^{\top}
\bigr).
\]
This rotate--aggregate--counter-rotate structure is required by the dynamical model: values from different time steps must be brought into a common temporal frame before aggregation, and the counter-rotation restores the output to the original frame.

The estimate is projected back and added to the residual stream, as in
standard attention:
\[
\boldsymbol{Z}^{+}=\boldsymbol{Z}+\boldsymbol{W}_o\bar{\boldsymbol{V}}.
\]
The estimation view suggests adding only the correction toward the estimate,
$\boldsymbol W_o(\bar{\boldsymbol V}-\boldsymbol V)$; the two differ by
$\boldsymbol W_o\boldsymbol V$, a linear function of the current token, and
we use the standard form for simplicity
(Appendix~\ref{sec:residual_connection}).

\subsection{Real-valued Implementation}
Although RFA is formulated over $\mathbb{C}^d$, all operations reduce to standard real arithmetic, as detailed in Appendix~\ref{sec:Complex-valued Computations}. The complex projections $\boldsymbol{W}_q, \boldsymbol{W}_k, 
\boldsymbol{W}_v \in \mathbb{C}^{d \times d}$ are implemented as real $d \times 2d$ 
matrices and $\boldsymbol{W}_o$ as a $2d \times d$ matrix. The complex rotations  $\tilde{\boldsymbol{\Phi}}^-$ and $\tilde{\boldsymbol{\Phi}}^+$ correspond exactly to RoPE rotation and counter-rotation. RoPE corresponds to zero decay ($\mu = 0$), so that the complex exponential $e^{\boldsymbol{\Lambda}\Delta t}$ reduces to pure rotation and the decay kernel $\boldsymbol{E}$ drops out.

Under the isotropic constraint, RFA preserves the asymptotic complexity of standard attention. The dominant operation remains a single $\mathcal{O}(N^2 d)$ matrix multiplication to compute the cross-term $\mathrm{Re}(\tilde{\boldsymbol{Q}}^\dagger \tilde{\boldsymbol{K}})$. The remaining components—the decay kernel $\boldsymbol{E}[i,j]$ and precision kernel $\boldsymbol{P}_{\Delta t}[i,j]$—are computed via elementwise operations with $\mathcal{O}(N^2)$ cost, and do not change the asymptotic complexity.

The complete implementation of the Isotropic RFA mechanism is summarized in Figure~\ref{fig:rfa_block} (Appendix~\ref{sec:RFA_Mechanism}) and formalized in Algorithm~\ref{alg:Isotropic_RFA} (Appendix~\ref{sec:Implementation}).
% \footnote{An anonymous multi-head implementation is available at \url{https://github.com/Anonymous18482/Robust-Filter-Attention}.}
\footnote{A multi-head implementation is available at \url{https://github.com/PCR-git/Robust-Filter-Attention}.}

\subsection{Layer Depth as Iterative State Estimation}

Each layer takes one step of an iteratively reweighted least-squares (IRLS)
procedure for the robust objective. Given a current state estimate, the
weights $w_{ij}=\rho'(d_{ij}^2)$ are evaluated at that estimate, and the
weighted least-squares problem is solved in closed form for the state estimate
$\bar{\boldsymbol x}_i$. A single attention layer performs one such update,
using the previous layer's output as the current estimate; residual
connections implement a partial step toward the resulting estimate. Stacking
layers therefore yields successive refinements of the latent state estimate.
For the Student-$t$ penalty, the weight is the posterior mean inverse scale
$\mathbb E[s_{ij}^{-1}\mid d_{ij}^2]$, and the iteration is also an EM
procedure for the marginal state likelihood (Appendix~\ref{sec:EM}).

\subsection{Filtering Behaviors}

The uncertainty model influences the attention weights through two coupled mechanisms. The additive bias $\boldsymbol{B}_{\Delta t} := \log(\boldsymbol{P}_{\Delta t})$ 
acts as a prior budget allocated to tokens at each lag, while the multiplicative gate 
$\boldsymbol{P}_{\Delta t}$ controls the selectivity of attention — how sharply the 
model discriminates between consistent and inconsistent tokens at that lag. Both are determined by the learned noise parameters $(\tilde{\sigma}^2, \eta^2, \gamma^2)$ through the scalar variance $\boldsymbol{\Sigma}_{\Delta t} = \alpha e^{-2\mu\Delta t} + \beta$.

The sign of $\alpha = \eta^2 - \tilde{\sigma}^2$ defines a phase transition between two qualitatively distinct filtering behaviors:

\textbf{Diffusive Regime} ($\alpha < 0$).
When process noise dominates ($\tilde{\sigma}^2 > \eta^2$), uncertainty accumulates monotonically with lag. The precision $\boldsymbol{P}_{\Delta t}$ acts as a \emph{closing 
gate}: selectivity is maximal near the diagonal and degrades as lag increases. The 
additive bias decays toward a floor of $-\log(\beta)$, implementing a forgetting prior 
that suppresses distant tokens. These heads implement a recency bias that is linear at short lags, as in
ALiBi, but levels off at this floor (Appendix~\ref{sec:pos_heuristics}).

\textbf{Integrative Regime} ($\alpha > 0$).
When measurement noise dominates ($\eta^2 > \tilde{\sigma}^2$), the precision acts as an \emph{opening gate}: 
selectivity is low near the diagonal and increases with lag as the initial measurement noise on the transported key dissipates under the stable dynamics. The additive bias correspondingly starts 
low and curves upward, implementing a settling prior that delays commitment until 
the transported observation becomes reliable. These heads function as lag-selective 
denoising filters, suppressing transient noise to identify stable historical structure.

Because $\alpha$ is determined by learned scalars per head, different heads can 
self-organize into different regimes during training. Explicit functional 
forms for the bias and gate in each regime are derived in Appendix~\ref{sec:pos_heuristics}.

\subsection{Recovery of Standard Positional Encodings}
\label{sec:physical_interpretation}

If the queries and keys are 
normalized, the Mahalanobis distance reduces to a dot product between queries and keys. In the zero-decay ($\mu = 0$), short-lag limit $(\Delta t \ll (\eta^2+\gamma^2)/\sigma^2)$, the RFA additive bias reduces to:
\[
\boldsymbol{B}_{\Delta t}
\approx
 - \frac{\sigma^2}{\eta^2+\gamma^2}\,\Delta t + \text{const},
\]
recovering a linear distance penalty consistent with the form of ALiBi's linear bias (Appendix~\ref{sec:alibi_derivation}).

Standard dot product attention with rotary positional encoding corresponds to the special case in which the dynamics are noiseless 
($\sigma^2 = 0 $), decay is absent ($\mu = 0$), query and key are normalized, and value rotation is omitted. Under these conditions, the state transition matrix reduces to a complex rotation and the precision kernel is uniform.

\subsection{Spectrally Coupled RFA (SC-RFA)}
\label{sec:SC_RFA}

Standard positional mechanisms such as RoPE utilize a fixed frequency bank across 
all heads, allowing high-frequency oscillations to persist indefinitely. At long 
horizons, this leads to phase wrap-around: tokens separated by a full oscillation 
period produce similar phase configurations, making them difficult to distinguish 
based on position alone.

To address this, we introduce Spectrally Coupled RFA (SC-RFA), in which decay rates 
are coupled to frequencies across heads. We partition a global frequency bank $\Omega$ 
monotonically across heads, assigning each head $h$ a spectral band 
$[\omega_{h,\min}, \omega_{h,\max}]$, and couple each head's decay rate to its maximum frequency:
\[
\mu_h = b \cdot \omega_{h,\max},
\]
where $b \in \mathbb{R}^+$ is a dimensionless damping coefficient (Appendix~\ref{sec:SC_RFA_appendix}). This enforces a 
fixed decay per oscillation cycle: over one period, the signal is attenuated by a factor of $e^{-2 \pi b}$, directly controlling the trade-off between spectral resolution 
and long-range stability.

Each attention head induces a prior temporal response given by the product of a decay term and a precision term:
\[
\boldsymbol{P}_{\Delta t} \cdot e^{-\mu \Delta t}
\;\propto\;
\frac{e^{-\mu \Delta t}}{\alpha e^{-2\mu \Delta t} + \beta},
\]
where $\alpha = \eta^2 - \tilde{\sigma}^2$ and $ \beta = \gamma^2 + \tilde{\sigma}^2$.

When $\alpha > \beta$, precision initially increases with lag as  key-side measurement noise dissipates, competing with exponential decay. The product peaks at a characteristic lag:
\[
\Delta t^* = \frac{1}{2\mu} \log\!\bigg(\frac{\alpha}{\beta}\bigg).
\]
Since $\mu_h$ varies across heads, peak locations $\Delta t_h^* \propto 1/\mu_h$ span 
a range of temporal scales, forming a bank of lag-selective filters. For sufficiently 
small $\mu$, the peak may lie beyond the training context length, causing the 
integrative profile to appear as a monotonically increasing function over the training 
window and inducing an anti-recency bias that extrapolates poorly. Diffusive heads 
($\alpha < 0$) provide a fallback by enforcing monotonic decay and stabilizing 
long-range behavior.

\section{Experimental Evaluation and Ablations}
\label{sec:Experiments}

We evaluate whether explicitly modeling uncertainty growth improves long-context stability while preserving short-range accuracy, comparing RFA against two widely used positional baselines derived from deterministic geometry (RoPE) \citep{su2023roformerenhancedtransformerrotary} and monotonic recency biasing (ALiBi) \citep{press2022trainshorttestlong}, respectively. The per-head scalar parameters ($\tilde{\sigma}^2, \eta^2, \gamma^2, \nu_s, \beta_s $) are learned entirely within the training window ($L=512$), and evaluated under zero-shot extrapolation at longer context lengths ($L \in \{512, 1024, 2048, 4096\})$.

\subsection{Experimental Setup}

\textbf{Architecture.} All models use a 6-layer Transformer with $h=8$ heads and embedding dimension $d=256$. To ensure comparable model capacity, we apply identical $d \to 2d \to d$ projections in both RFA and the RoPE/ALiBi baselines, which represent the mapping between the real and complex domains (Appendix~\ref{sec:Complex-valued Computations}). RFA introduces only a small number of additional scalar parameters per head for noise and robustness, increasing total parameter count by approximately 0.02\%. We employ a pre-norm architecture with an FFN expansion factor of 4. Models are trained for 15 epochs until convergence using Adam with a cosine learning rate schedule.

\textbf{Datasets.}
We evaluate on WikiText-103, a large-scale word-level language modeling benchmark derived from Wikipedia articles and used to measure perplexity and long-context extrapolation \citep{merity2016pointersentinelmixturemodels}, and on BabyLM-2025 (Strict), a curated English language modeling corpus used as a complementary benchmark under the same training and evaluation protocol \citep{charpentier-etal-2025-findings}.

\textbf{Ablations.} We compare against standard positional baselines: RoPE (B1) and ALiBi (B2), and include two geometry-only decay variants to isolate the effect of damping in rotational embeddings: Decayed RoPE (B3), which applies exponential decay with distance, as in RFA, and SC-RoPE (B4), which couples decay rates to head-wise frequency bands, as in SC-RFA. These baselines test whether decay and spectral coupling alone can explain extrapolation gains, without modeling uncertainty.

We evaluate RFA (M1) and two variants of SC-RFA: one optimized for in-window performance (M2), with $b=0.05$, and one optimized for long-context extrapolation via increased damping (M3), with $b=5.0$. We also include structural ablations relative to M2, designed to isolate the effect of its components when removed:  the power-law robust weight $w_{ij}$, replacing it with an exponential weight (M2.1); the DLE-derived precision kernel (M2.2); the multiplicative gating term $\boldsymbol{P}_{\Delta t}$ (M2.3); value rotations (M2.4); all rotations (M2.5); a purely rotational, zero decay and zero noise variant, analogous to RoPER (M2.6); SC rotation and the additive DLE bias, equivalent to partitioned RoPE with a logarithmic bias (M2.7); and finally, M2.6 without value rotation, recovering rotary dot-product attention (M2.8).

Full architectural and ablation details are provided in 
Appendix~\ref{sec:Experimental_Details}. Analyses of attention maps and noise parameters are provided in Appendix~\ref{sec:Additional_Results}.

\subsection{Results on Wikitext-103}

We evaluate extrapolation by measuring test perplexity on WikiText-103 at increasing context lengths, after training all models with a fixed context window of 512 tokens. Results are shown in Table~\ref{tab:extrapolation_wikitext}.

\begin{table}[t]
\centering
\caption{Long-context extrapolation on WikiText-103 (Test PPL). All models were trained with a fixed context window of 512 tokens.}
\label{tab:extrapolation_wikitext}
\small
\renewcommand{\arraystretch}{1.1}
\begin{tabularx}{\columnwidth}{l *{4}{>{\centering\arraybackslash}X}}
\toprule
\textbf{Model} & \textbf{L=512} & \textbf{L=1024} & \textbf{L=2048} & \textbf{L=4096} \\
\midrule
RoPE (B1)            & 27.93 & 40.55 & 100.63  & 202.36 \\
ALiBi (B2)          & 28.59   & 27.30 & 26.54 & \textbf{26.30} \\
Decayed RoPE (B3)   & 27.86 & 34.60 & 67.43 & 126.86 \\
SC-RoPE (B4)        & 28.02 & 32.45 & 48.56  & 87.44 \\
\midrule
\rowcolor[gray]{0.95} RFA (M1)     & 28.01 & 27.58 & 29.99 & 38.46 \\
\rowcolor[gray]{0.95} \textbf{SC-RFA (M2)} & \textbf{27.54} & 26.73 & 29.46 & 37.19 \\
% \rowcolor[gray]{0.95} \textbf{SC-RFA (M3)} & 28.09 & 26.98 & 26.76 & 27.92 \\
\rowcolor[gray]{0.95} \textbf{SC-RFA (M3)} & 27.91 & 26.68 & \textbf{26.37} & 28.16 \\
\midrule
\multicolumn{5}{l}{\textit{Structural Ablations (Relative to M2)}} \\ 
Exp. Weight (M2.1) & 27.98 & 27.16 & 28.95 & 33.51 \\
Flat Prior (M2.2) & 27.69 & 28.71 & 38.11 & 62.83 \\
No Mult. Gate (M2.3) & 27.65 & 29.01 & 39.18 & 57.30 \\
No Value Rot. (M2.4) & 27.80 & \textbf{26.55} & 26.69 & 28.56 \\
No Rotations (M2.5) & 28.58 & 27.25 & 26.61 & 26.83 \\
Pure Rotation (M2.6) & 27.97 & 35.59 & 69.39 & 131.29 \\
Minimal SC-RFA (M2.7) & 27.88 & 31.24 & 43.88 & 68.40 \\
Rotary Limit (M2.8) & 28.00 & 39.77 & 78.58 & 133.61 \\
\bottomrule
\end{tabularx}
\end{table}

SC-RFA (M2) achieves the lowest in-window perplexity ($27.54$ vs.\ $27.93$ for
RoPE) and degrades far more gracefully beyond the training window ($37.19$
vs.\ $202.36$ at $L=4096$). Without spectral coupling, RFA (M1) matches RoPE
in-window ($28.01$) but retains most of the extrapolation benefit ($38.46$).
M2 outperforms M1 at every context length.

Unlike RoPE, which degrades monotonically outside the training window, RFA exhibits a non-monotonic extrapolation profile: perplexity initially decreases beyond the training horizon before increasing at longer context lengths.

% This indicates that additional context can improve state estimation over a finite range beyond the training window, rather than being uniformly detrimental, corresponding to an integrative regime in which distant tokens contribute useful signal before eventual degradation.

With higher damping coefficient, SC-RFA (M3) further improves long-context stability, achieving nearly flat perplexity at up to $L=4096$ while maintaining competitive performance within the training window. This behavior emerges under a fixed training protocol without requiring length-dependent scaling rules or curriculum schedules.

Decay (B3) slows RoPE's long-range degradation, and spectral coupling (B4)
slows it further ($87.44$ vs.\ $126.86$ at $L=4096$), but both trail SC-RFA at
every context length. Decay and spectrum alone therefore do not
explain RFA's gains without explicit uncertainty modeling.

% This behavior does not require fine-tuning, curriculum schedules, or manual scaling rules: the transition from high-resolution local modeling to low-resolution long-range integration emerges directly from the learned noise and decay parameters as a zero-shot consequence of the latent stochastic dynamics.

The exponential weight variant of SC-RFA (M2.1) under-performs the power-law robust weighting at the training horizon, but achieves lower perplexity at extreme extrapolation lengths. This is consistent with Gaussian likelihoods imposing stronger quadratic penalties on residuals, which suppress extreme deviations more aggressively but reduce sensitivity to small errors when uncertainty is low.

Removing the DLE-derived precision kernel (M2.2) leads to degradation at long horizons, with perplexity increasing to $62.83$ at $L=4096$, demonstrating that explicit uncertainty propagation via the DLE is necessary for stable long-context behavior. Removing the precision term $\boldsymbol{P}_{\Delta t}$ from the Mahalanobis
distance (M2.3) slightly worsens in-window perplexity and worsens
extrapolation ($57.30$ at $L=4096$), suggesting that the multiplicative
precision term also contributes to stability.

Removing value rotation (M2.4) slightly worsens in-window perplexity
($27.80$) but extrapolates more stably ($28.56$ at $L=4096$), so value-frame
alignment appears to aid in-window modeling, while stability comes from decay
and uncertainty. Removing all rotations (M2.5) degrades short-context
performance but yields strong long-range stability, reducing RFA to a
logarithmic distance bias that performs similarly to ALiBi.

Without decay or noise (M2.6), perplexity rises sharply with context, as for RoPE, so value rotation alone does not prevent long-range degradation.
Reducing SC-RFA to partitioned RoPE with a learned logarithmic distance bias
(M2.7) ties RoPE in-window ($27.88$) and recovers only part of SC-RFA's
stability ($68.40$ vs.\ $37.19$ at $L=4096$), so the additive precision term
does not explain SC-RFA's gains. Finally, removing value rotation from M2.6 yields RoPE implemented within RFA (M2.8), which matches RoPE in-window within noise ($28.00$ vs.\ $27.93$).

% The structural ablations show that stable extrapolation requires the joint presence of: (i) state propagation via the matrix exponential, (ii) DLE-derived uncertainty accumulation, and (iii) value-frame alignment. Removing any of these components leads to rapid degradation at long context.

\begin{table}[t]
\centering
\caption{Sensitivity analysis of the damping coefficient $b$ in SC-RFA (M2), with RoPE (B1) and ALiBi (B2) as baselines. Results show Test PPL on WikiText-103 across increasing context lengths.}
\label{tab:b_sensitivity}
\small
\renewcommand{\arraystretch}{1.2}
\begin{tabularx}{\columnwidth}{l *{4}{>{\centering\arraybackslash}X}}
\toprule
\textbf{Damping ($b$)} & \textbf{L=512} & \textbf{L=1024} & \textbf{L=2048} & \textbf{L=4096} \\
\midrule
RoPE (B1)            & 27.93 & 40.55 & 100.63  & 202.36 \\
ALiBi (B2)         & 28.59   & 27.30 & 26.54 & \textbf{26.30} \\
\midrule
$5 \times 10^{-4}$ & 27.60 & 28.88 & 37.34 & 51.48 \\
$5 \times 10^{-3}$ & 27.60 & 28.71 & 35.35 & 43.90 \\
$5 \times 10^{-2}$ & \textbf{27.54} & 26.73 & 29.46 & 37.19 \\
$5 \times 10^{-1}$ & 27.61 & \textbf{26.38} & \textbf{26.37} & 29.72 \\
$5 \times 10^{0}$ & 27.91 & 26.68 & \textbf{26.37} & 28.16 \\
\bottomrule
\end{tabularx}
\end{table}

Compared to ALiBi, SC-RFA achieves lower perplexity at intermediate context lengths, $L=512$ and $L=1024$. At longer horizons, ALiBi attains lower perplexity: its penalty grows without bound, so the total weight on distant tokens stays bounded. In damped RFA heads, the penalty levels off, so each distant token keeps a fixed share of precision. This overcounts distant evidence as the context grows (Section~\ref{sec:Setup}), a cost of integrating distant context rather
than suppressing it.

Table~\ref{tab:b_sensitivity} shows that this tradeoff is continuously controlled in SC-RFA by the damping coefficient $b$. Smaller values of $b$ yield slower decay, improving short-context performance but leading to faster degradation as context increases. Larger values of $b$ produce stronger attenuation and more stable long-range behavior at the cost of reduced short-range performance. Notably, for sufficiently strong damping ($b \geq 0.5 $), SC-RFA outperforms ALiBi at intermediate horizons (up to $L=2048$), with ALiBi retaining an advantage only at the largest tested context length.

\subsection{Results on BabyLM-2025}

We use the same architectures, hyperparameters, and training protocol as on WikiText-103. On BabyLM-2025, where language modeling performance is more strongly dominated by short-range context, differences between positional mechanisms are smaller at short context lengths. All RFA variants roughly match RoPE within the training window and outperform it at longer contexts ($22.25$ vs.\ $52.86$ at $L=4096$ for SC-RFA). SC-RFA also beats ALiBi at $L=1024$, while ALiBi remains strongest
at the longest horizons due to its strict recency bias. These results mirror the trade-off observed on WikiText-103: precision weighting improves robustness over purely rotational embeddings while retaining stronger short- and mid-range performance than aggressively local recency biases.

\begin{table}[t]
\centering
\caption{Long-context extrapolation on BabyLM-2025 (Test PPL). All models were trained with a fixed context window of 512 tokens.}
\label{tab:extrapolation_babylm}
\small
\renewcommand{\arraystretch}{1.1}
\begin{tabularx}{\columnwidth}{l *{4}{>{\centering\arraybackslash}X}}
\toprule
\textbf{Model} & \textbf{L=512} & \textbf{L=1024} & \textbf{L=2048} & \textbf{L=4096} \\
\midrule
RoPE (B1)           & 17.46 & 20.71 & 32.14  & 52.86 \\
ALiBi (B2)          & 17.70   & 17.20 & \textbf{17.06} & \textbf{17.51} \\
\midrule
\rowcolor[gray]{0.95} RFA (M1) & 17.51 & 17.71 & 20.61 & 31.04 \\
\rowcolor[gray]{0.95} \textbf{SC-RFA (M2)} & \textbf{17.36} & \textbf{16.99} & 18.33 & 22.25 \\
\rowcolor[gray]{0.95} \textbf{SC-RFA (M3)} & 17.51 & 17.07 & 17.26 & 18.74 \\
\bottomrule
\end{tabularx}
\end{table}

\subsection{Learning Dynamics and Head Specialization}

SC-RFA (M2) reaches lower validation perplexity than RoPE and ALiBi at every
epoch, while M1 and M3 track RoPE closely, so their extrapolation gains come at no in-window cost (Appendix~\ref{sec:training_dynamics}). Analysis of learned noise, decay, and robustness parameters reveals systematic specialization across heads into distinct uncertainty and selectivity regimes (Appendix~\ref{app:m1_dynamics}).

Attention maps at long context lengths show how the two models place their
bands (Appendix~\ref{sec:attn_mats}). RoPE attends to many bands at fixed
lags, wherever the phases of its rotations realign, and these persist across
the entire context. In RFA, some heads instead exhibit an integrative regime,
in which attention is suppressed near the diagonal and peaks at a
characteristic lag.

SC-RFA sharpens this structure through spectral coupling: each decaying head
attends to a single band, whose lag shrinks as the head's decay rate grows,
so the heads form an ordered bank of lag-selective filters. Since decay alone
falls monotonically with lag, it cannot place a band away from the diagonal;
the peak comes from precision that rises with lag
(Section~\ref{sec:SC_RFA}). Together, these patterns support the
interpretation of RFA as learning a structured, uncertainty-aware temporal
filter bank rather than relying solely on geometric positional bias.

\section{Conclusion}
\label{sec:Conclusion}

We reformulate self-attention as a tractable approximation to state estimation under a linear stochastic dynamical model, yielding Robust Filter Attention (RFA). In this formulation, attention weights reflect uncertainty-aware agreement between dynamically transported representations rather than static feature similarity, while preserving the computational structure of standard attention. RFA recovers existing positional mechanisms as limiting cases and improves temporal consistency and long-context behavior.

Tractability requires linear dynamics, diagonalizability, and isotropic noise and decay per head. A central challenge for future work is relaxing these constraints while retaining tractable covariance propagation and parallel computation.

The flat prior treats each transported token as a full-strength measurement
of the current state, so the composite likelihood overcounts evidence from
distant tokens as the context grows. Priors that make this evidence decay
with lag are a natural next step.

Other important directions include understanding how these dynamical and
uncertainty-based mechanisms interact with architectural components such as
normalization and depth, and incorporating similar uncertainty-aware
formulations into other sequence modeling architectures.

\section*{Impact Statement}

This work introduces a dynamical formulation of self-attention. We do not identify ethical concerns beyond those generally associated with advances in machine learning.

% \section*{Acknowledgements}

% I would like to thank Dae Woong Kim for helpful discussions and feedback on this work.

\bibliography{references}

@inproceedings{gu2020hipporecurrentmemoryoptimal,
author = {Gu, Albert and Dao, Tri and Ermon, Stefano and Rudra, Atri and R\'{e}, Christopher},
title = {{HiPPO}: recurrent memory with optimal polynomial projections},
year = {2020},
isbn = {9781713829546},
publisher = {Curran Associates Inc.},
address = {Red Hook, NY, USA},
booktitle = {Proceedings of the 34th International Conference on Neural Information Processing Systems},
articleno = {125},
numpages = {14},
location = {Vancouver, BC, Canada},
series = {NIPS '20}
}

@inproceedings{
gu2022efficientlymodelinglongsequences,
title={Efficiently Modeling Long Sequences with Structured State Spaces},
author={Albert Gu and Karan Goel and Christopher Re},
booktitle={International Conference on Learning Representations},
year={2022},
url={https://openreview.net/forum?id=uYLFoz1vlAC}
}

@inproceedings{vaswani2023attentionneed,
 author = {Vaswani, Ashish and Shazeer, Noam and Parmar, Niki and Uszkoreit, Jakob and Jones, Llion and Gomez, Aidan N and Kaiser, \L ukasz and Polosukhin, Illia},
 booktitle = {Advances in Neural Information Processing Systems},
 editor = {I. Guyon and U. Von Luxburg and S. Bengio and H. Wallach and R. Fergus and S. Vishwanathan and R. Garnett},
 pages = {},
 publisher = {Curran Associates, Inc.},
 title = {Attention is All you Need},
 volume = {30},
 year = {2017}
}

@inproceedings{dao2024transformersssmsgeneralizedmodels,
author = {Dao, Tri and Gu, Albert},
title = {Transformers are {SSMs}: generalized models and efficient algorithms through structured state space duality},
year = {2024},
publisher = {JMLR.org},
booktitle = {Proceedings of the 41st International Conference on Machine Learning},
articleno = {399},
numpages = {31},
location = {Vienna, Austria},
series = {ICML'24}
}

@inproceedings{tsai2019transformerdissectionunifiedunderstanding,
    title = "Transformer Dissection: A Unified Understanding for {T}ransformer{'}s Attention via the Lens of Kernel",
    author = "Tsai, Yao-Hung Hubert  and
      Bai, Shaojie  and
      Yamada, Makoto  and
      Morency, Louis-Philippe  and
      Salakhutdinov, Ruslan",
    booktitle = "Proceedings of the 2019 Conference on Empirical Methods in Natural Language Processing and the 9th International Joint Conference on Natural Language Processing (EMNLP-IJCNLP)",
    month = nov,
    year = "2019",
    address = "Hong Kong, China",
    publisher = "Association for Computational Linguistics",
    doi = "10.18653/v1/D19-1443",
    pages = "4344--4353"
}

@article{su2023roformerenhancedtransformerrotary,
title = {{RoFormer}: Enhanced transformer with Rotary Position Embedding},
journal = {Neurocomputing},
volume = {568},
pages = {127063},
year = {2024},
issn = {0925-2312},
doi = {https://doi.org/10.1016/j.neucom.2023.127063},
author = {Jianlin Su and Murtadha Ahmed and Yu Lu and Shengfeng Pan and Wen Bo and Yunfeng Liu}}

@inproceedings{chen2019neuralordinarydifferentialequations,
author = {Chen, Ricky T. Q. and Rubanova, Yulia and Bettencourt, Jesse and Duvenaud, David},
title = {Neural ordinary differential equations},
year = {2018},
publisher = {Curran Associates Inc.},
address = {Red Hook, NY, USA},
booktitle = {Proceedings of the 32nd International Conference on Neural Information Processing Systems},
pages = {6572–6583},
numpages = {12},
location = {Montr\'{e}al, Canada},
series = {NIPS'18}
}

@misc{sun2023retentivenetworksuccessortransformer,
      title={{Retentive Network}: A Successor to {Transformer} for Large Language Models}, 
      author={Yutao Sun and Li Dong and Shaohan Huang and Shuming Ma and Yuqing Xia and Jilong Xue and Jianyong Wang and Furu Wei},
      year={2023},
      eprint={2307.08621},
      archivePrefix={arXiv},
      primaryClass={cs.CL},
      url={https://arxiv.org/abs/2307.08621}, 
}

@article{originalkalmanfilter ,
    Author = {Kalman, Rudolph Emil},
    Title = {A New Approach to Linear Filtering and Prediction Problems},
    Journal = {Transactions of the ASME--Journal of Basic Engineering},
    Volume = {82},
    Number = {Series D},
    Pages = {35--45},
    Year = {1960}
}

@InProceedings{goel2024transformerrepresentkalmanfilter,
  title = 	 {Can a {Transformer} represent a {K}alman filter?},
  author =       {Goel, Gautam and Bartlett, Peter},
  booktitle = 	 {Proceedings of the 6th Annual Learning for Dynamics \& Control Conference},
  pages = 	 {1502--1512},
  year = 	 {2024},
  editor = 	 {Abate, Alessandro and Cannon, Mark and Margellos, Kostas and Papachristodoulou, Antonis},
  volume = 	 {242},
  series = 	 {Proceedings of Machine Learning Research},
  month = 	 {15--17 Jul},
  publisher =    {PMLR},
  url = 	 {https://proceedings.mlr.press/v242/goel24a.html}
}

@inproceedings{
    ramsauer2021hopfieldnetworksneed,
    title={Hopfield Networks is All You Need},
    author={Hubert Ramsauer and Bernhard Sch{\"a}fl and Johannes Lehner and Philipp Seidl and Michael Widrich and Lukas Gruber and Markus Holzleitner and Thomas Adler and David Kreil and Michael K Kopp and G{\"u}nter Klambauer and Johannes Brandstetter and Sepp Hochreiter},
    booktitle={International Conference on Learning Representations},
    year={2021},
    url={https://openreview.net/forum?id=tL89RnzIiCd}
    }

@inproceedings{
gabbur2021probabilistic,
title={Probabilistic Attention for Interactive Segmentation},
author={Prasad Gabbur and Manjot Bilkhu and Javier Movellan},
booktitle={Advances in Neural Information Processing Systems},
editor={A. Beygelzimer and Y. Dauphin and P. Liang and J. Wortman Vaughan},
year={2021},
url={https://openreview.net/forum?id=JpDlWGTBHB}
}

@inproceedings{bui2025revisitingkernelattentioncorrelated,
author = {Bui, Long Minh and Huu, Tho Tran and Dinh, Duy and Nguyen, Tan Minh and Hoang, Trong Nghia},
title = {Revisiting kernel attention with correlated {Gaussian} process representation},
year = {2024},
publisher = {JMLR.org},
booktitle = {Proceedings of the Fortieth Conference on Uncertainty in Artificial Intelligence},
articleno = {22},
numpages = {21},
location = {Barcelona, Spain},
series = {UAI '24}
}

@article{Cohen_2025,
title = {Adaptive {K}alman-Informed {T}ransformer},
journal = {Engineering Applications of Artificial Intelligence},
volume = {146},
pages = {110221},
year = {2025},
issn = {0952-1976},
doi = {https://doi.org/10.1016/j.engappai.2025.110221},
author = {Nadav Cohen and Itzik Klein}
}

@article{shen2025kalmanformer,
  title={{KalmanFormer}: using {Transformer} to model the {Kalman} Gain in {Kalman} Filters},
  author={Shen, Siyuan and Chen, Jichen and Yu, Guanfeng and Zhai, Zhengjun and Han, Pujie},
  journal={Frontiers in Neurorobotics},
  volume={18},
  pages={1460255},
  year={2025},
  doi={10.3389/fnbot.2024.1460255}
}

@article{Liu_2023,
   title={Neural extended {Kalman} filters for learning and predicting dynamics of structural systems},
   volume={23},
   ISSN={1741-3168},
   DOI={10.1177/14759217231179912},
   number={2},
   journal={Structural Health Monitoring},
   publisher={SAGE Publications},
   author={Liu, Wei and Lai, Zhilu and Bacsa, Kiran and Chatzi, Eleni},
   year={2023},
   month=jun, pages={1037–1052} }

@article{Revach_2022,
   title={{KalmanNet}: Neural Network Aided {Kalman} Filtering for Partially Known Dynamics},
   volume={70},
   ISSN={1941-0476},
   DOI={10.1109/tsp.2022.3158588},
   journal={IEEE Transactions on Signal Processing},
   publisher={Institute of Electrical and Electronics Engineers (IEEE)},
   author={Revach, Guy and Shlezinger, Nir and Ni, Xiaoyong and Escoriza, Adria Lopez and van Sloun, Ruud J. G. and Eldar, Yonina C.},
   year={2022},
   pages={1532–1547} }

@inproceedings{
bianchessi2026bayesian,
title={{B}ayesian Attention Mechanism: A Probabilistic Framework for Positional Encoding and Context Length Extrapolation},
author={Arthur S. Bianchessi and Yasmin C. Aguirre and Rodrigo C. Barros and Lucas S. Kupssinsk{\"u}},
booktitle={The Fourteenth International Conference on Learning Representations},
year={2026},
url={https://openreview.net/forum?id=dXJB9O8fLd}
}

@article{jahanshahi2026,
author = {Jahanshahi, Hadi and Zhu, George},
year = {2026},
month = {02},
pages = {133134},
title = {Uncertainty Propagation Networks for Neural Ordinary Differential Equations},
volume = {677},
journal = {Neurocomputing},
doi = {10.1016/j.neucom.2026.133134}
}

@inproceedings{
press2022trainshorttestlong,
title={Train Short, Test Long: Attention with Linear Biases Enables Input Length Extrapolation},
author={Ofir Press and Noah Smith and Mike Lewis},
booktitle={International Conference on Learning Representations},
year={2022},
url={https://openreview.net/forum?id=R8sQPpGCv0}
}

@InProceedings{li2020scalablegradientsstochasticdifferential,
  title = 	 {Scalable Gradients and Variational Inference for
 Stochastic Differential Equations },
  author =       {Li, Xuechen and Wong, Ting-Kam Leonard and Chen, Ricky T. Q. and Duvenaud, David K.},
  booktitle = 	 {Proceedings of The 2nd Symposium on
 Advances in Approximate Bayesian Inference},
  pages = 	 {1--28},
  year = 	 {2020},
  editor = 	 {Zhang, Cheng and Ruiz, Francisco and Bui, Thang and Dieng, Adji Bousso and Liang, Dawen},
  volume = 	 {118},
  series = 	 {Proceedings of Machine Learning Research},
  month = 	 {08 Dec},
  publisher =    {PMLR},
  url = 	 {https://proceedings.mlr.press/v118/li20a.html}
}

@misc{shen2025neuralsdesunifiedapproach,
      title={Neural {SDE}s as a Unified Approach to Continuous-Domain Sequence Modeling}, 
      author={Macheng Shen and Chen Cheng},
      year={2025},
      eprint={2501.18871},
      archivePrefix={arXiv},
      primaryClass={cs.LG},
      url={https://arxiv.org/abs/2501.18871}, 
}

@inproceedings{sun2022lengthextrapolatabletransformer,
    title = "A Length-Extrapolatable {Transformer}",
    author = "Sun, Yutao  and
      Dong, Li  and
      Patra, Barun  and
      Ma, Shuming  and
      Huang, Shaohan  and
      Benhaim, Alon  and
      Chaudhary, Vishrav  and
      Song, Xia  and
      Wei, Furu",
    editor = "Rogers, Anna  and
      Boyd-Graber, Jordan  and
      Okazaki, Naoaki",
    booktitle = "Proceedings of the 61st Annual Meeting of the Association for Computational Linguistics (Volume 1: Long Papers)",
    month = jul,
    year = "2023",
    address = "Toronto, Canada",
    publisher = "Association for Computational Linguistics",
    doi = "10.18653/v1/2023.acl-long.816",
    pages = "14590--14604"
}

@inproceedings{
merity2016pointersentinelmixturemodels,
title={Pointer Sentinel Mixture Models},
author={Stephen Merity and Caiming Xiong and James Bradbury and Richard Socher},
booktitle={International Conference on Learning Representations},
year={2017},
url={https://openreview.net/forum?id=Byj72udxe}
}

@article{Huber1964,
  author       = {Huber, Peter J.},
  title        = {Robust Estimation of a Location Parameter},
  journal      = {The Annals of Mathematical Statistics},
  volume       = {35},
  number       = {1},
  pages        = {73--101},
  year         = {1964},
  month        = mar,
  doi          = {10.1214/aoms/1177703732}
}

@inproceedings{
han2023designingrobusttransformersusing,
title={Designing Robust {Transformers} using Robust Kernel Density Estimation},
author={Xing Han and Tongzheng Ren and Tan Minh Nguyen and Khai Nguyen and Joydeep Ghosh and Nhat Ho},
booktitle={Thirty-seventh Conference on Neural Information Processing Systems},
year={2023},
url={https://openreview.net/forum?id=BqTv1Mtuhu}
}

@inproceedings{nguyen2022improving,
  author={Tam Minh Nguyen and Tan Minh Nguyen and Dung D. D. Le and Duy Khuong Nguyen and Viet-Anh Tran and Richard G. Baraniuk and Nhat Ho and Stanley J. Osher},
  title={Improving {Transformers} with Probabilistic Attention Keys},
  year={2022},
  cdate={1640995200000},
  pages={16595-16621},
  url={https://proceedings.mlr.press/v162/nguyen22c.html},
  booktitle={ICML}
}

@inproceedings{liu2020kalmanfilteringattentionuser,
author = {Liu, Hu and Lu, Jing and Zhao, Xiwei and Xu, Sulong and Peng, Hao and Liu, Yutong and Zhang, Zehua and Li, Jian and Jin, Junsheng and Bao, Yongjun and Yan, Weipeng},
title = {Kalman filtering attention for user behavior modeling in {CTR} prediction},
year = {2020},
isbn = {9781713829546},
publisher = {Curran Associates Inc.},
address = {Red Hook, NY, USA},
booktitle = {Proceedings of the 34th International Conference on Neural Information Processing Systems},
articleno = {774},
numpages = {11},
location = {Vancouver, BC, Canada},
series = {NIPS '20}
}

@misc{harik2022roper,
    author = {Harik, Georges and Jayasiri, Varuna},
    title  = {Rotary Positional Embeddings with Relative distances},
    url    = {http://research.labml.ai/RoPER.html},
    year   = {2022}
}

@inproceedings{
movahedi2026selective,
title={Selective Rotary Position Embedding},
author={Sajad Movahedi and Timur Carstensen and Arshia Afzal and Frank Hutter and Antonio Orvieto and Volkan Cevher},
booktitle={The Fourteenth International Conference on Learning Representations},
year={2026},
url={https://openreview.net/forum?id=AQo1SEElNb}
}

@inproceedings{
peng2023yarnefficientcontextwindow,
title={Ya{RN}: Efficient Context Window Extension of Large Language Models},
author={Bowen Peng and Jeffrey Quesnelle and Honglu Fan and Enrico Shippole},
booktitle={The Twelfth International Conference on Learning Representations},
year={2024},
url={https://openreview.net/forum?id=wHBfxhZu1u}
}

@inproceedings{charpentier-etal-2025-findings,
    title = "Findings of the Third {B}aby{LM} Challenge: Accelerating Language Modeling Research with Cognitively Plausible Data",
    author = "Charpentier, Lucas  and
      Choshen, Leshem  and
      Cotterell, Ryan  and
      Gul, Mustafa Omer  and
      Hu, Michael Y.  and
      Liu, Jing  and
      Jumelet, Jaap  and
      Linzen, Tal  and
      Mueller, Aaron  and
      Ross, Candance  and
      Shah, Raj Sanjay  and
      Warstadt, Alex  and
      Wilcox, Ethan Gotlieb  and
      Williams, Adina",
    booktitle = "Proceedings of the First BabyLM Workshop",
    month = nov,
    year = "2025",
    address = "Suzhou, China",
    publisher = "Association for Computational Linguistics",
    doi = "10.18653/v1/2025.babylm-main.28",
    pages = "399--420"
}

@article{Elman_RNNs,
author = {Elman, Jeffrey L.},
title = {Finding Structure in Time},
journal = {Cognitive Science},
volume = {14},
number = {2},
pages = {179-211},
doi = {https://doi.org/10.1207/s15516709cog1402_1},
eprint = {https://onlinelibrary.wiley.com/doi/pdf/10.1207/s15516709cog1402_1},
year = {1990}
}

@inproceedings{shaj2026kalmanlinearattentionparallel,
    title={{Kalman Linear Attention}: Parallel {B}ayesian Filtering For Efficient Language Modeling and State Tracking},
    author={Vaisakh Shaj and Cameron Barker and Aidan Scannell and Andras Szecsenyi and Elliot J. Crowley and Amos Storkey},
    booktitle={Forty-third International Conference on Machine Learning},
    year={2026},
    url={https://openreview.net/forum?id=9h7sSJe4jN}
    }

@inproceedings{
zuo2026local,
title={{Local Linear Attention}: An Optimal Interpolation of Linear and Softmax Attention For Test-Time Regression},
author={Yifei Zuo and Yutong Yin and Zhichen Zeng and Ang Li and Banghua Zhu and Zhaoran Wang},
booktitle={The Fourteenth International Conference on Learning Representations},
year={2026},
url={https://openreview.net/forum?id=WGpzi489XY}
}

@incollection{lindsay1988composite,
  author    = {Lindsay, Bruce G.},
  editor    = {Prabhu, N. U.},
  title     = {Composite Likelihood Methods},
  booktitle = {Statistical Inference from Stochastic Processes},
  series    = {Contemporary Mathematics},
  volume    = {80},
  pages     = {221--239},
  year      = {1988},
  publisher = {American Mathematical Society}}

@article{varin2011overview,
  author  = {Varin, Cristiano and Reid, Nancy and Firth, David},
  title   = {An Overview of Composite Likelihood Methods},
  journal = {Statistica Sinica},
  volume  = {21},
  number  = {1},
  pages   = {5--42},
  year    = {2011}
}

@article{Lange1989RobustSM,
  title={Robust Statistical Modeling Using the t Distribution},
  author={Kenneth Lange and Roderick J. Little and Jeremy M. G. Taylor},
  journal={Journal of the American Statistical Association},
  year={1989},
  volume={84},
  pages={881-896},
  url={https://api.semanticscholar.org/CorpusID:122769764}
}

@book{durbin_koopman_2012time,
title = "Time series analysis by state space methods",
author = "J. Durbin and S.J. Koopman",
year = "2012",
language = "English",
publisher = "Oxford University press"
}

@inproceedings{
chi2022kerple,
title={{KERPLE}: Kernelized Relative Positional Embedding for Length Extrapolation},
author={Ta-Chung Chi and Ting-Han Fan and Peter Ramadge and Alexander Rudnicky},
booktitle={Advances in Neural Information Processing Systems},
editor={Alice H. Oh and Alekh Agarwal and Danielle Belgrave and Kyunghyun Cho},
year={2022},
url={https://openreview.net/forum?id=hXzOqPlXDwm}
}
\bibliographystyle{icml2026}

%%%%%%%%%%%%%%%%%%%%%%%%%%%%%%%%%%%%%%%%%%%%%%%%%%%%%%%%%%%%%%%%%%%%%%%%%%%%%%%
%%%%%%%%%%%%%%%%%%%%%%%%%%%%%%%%%%%%%%%%%%%%%%%%%%%%%%%%%%%%%%%%%%%%%%%%%%%%%%%
% APPENDIX
%%%%%%%%%%%%%%%%%%%%%%%%%%%%%%%%%%%%%%%%%%%%%%%%%%%%%%%%%%%%%%%%%%%%%%%%%%%%%%%
%%%%%%%%%%%%%%%%%%%%%%%%%%%%%%%%%%%%%%%%%%%%%%%%%%%%%%%%%%%%%%%%%%%%%%%%%%%%%%%
\newpage
\appendix
\onecolumn

\section*{Appendix Table of Contents}

\begin{enumerate}

\item
\noindent\textbf{Appendix A: {Attention from a Batch State Estimation Formulation}} \hfill\pageref{sec:Derivation}

\begin{itemize}
\item \textit{Derives a precision-weighted attention mechanism from a batch state estimation formulation.}
\end{itemize}

\item
\noindent\textbf{Appendix B: Robust Filter Attention Mechanism}\hfill\pageref{sec:RFA_Mechanism}

\begin{itemize}
\item \textit{Covers the simplification from the full anisotropic tensor formulation of RFA to the scalable isotropic variant.}
\end{itemize}

\item
\noindent\textbf{Appendix C: Recovery of Positional Encodings}\hfill\pageref{sec:recovery_pos_encodings}

\begin{itemize}
\item \textit{Provides a physical interpretation of common positional embeddings, showing how RoPE and ALiBi arise as limiting cases of RFA.}
\end{itemize}

\item
\noindent\textbf{Appendix D: Real-Valued Realization and Initialization}\hfill\pageref{sec:Implementation}

\begin{itemize}
\item \textit{Provides the complex-to-real isomorphism for hardware-efficient computation and full pseudocode for the Isotropic RFA layer.}
\end{itemize}

\item
\noindent\textbf{Appendix E: Extensions}\hfill\pageref{sec:Extensions}

\begin{itemize}
\item \textit{Describes several natural extensions of the RFA framework.}
\end{itemize}

\item
\noindent\textbf{Appendix F: Experimental Details and Ablations}\hfill\pageref{sec:Experimental_Details}

\begin{itemize}
\item \textit{Defines all ablation variants, training setup, and hyperparameters.}
\end{itemize}

\item
\noindent\textbf{Appendix G: Additional Experimental Results}\hfill\pageref{sec:Additional_Results}

\begin{itemize}
\item \textit{Analyzes learned noise parameters, head specialization, and long-context attention maps to illustrate how RFA implements multi-scale filtering behavior in practice.}
\end{itemize}

\end{enumerate}

%%%%%%%%%%%%%%%%%%%%%%%%%%%%%%%%%%%%%%%%%%%

\newpage

\section{Attention from a Batch State Estimation Formulation}
\label{sec:Derivation}

Here, we derive the RFA mechanism introduced in Section~\ref{sec:method} in greater detail.

\subsection{Latent Dynamics}
\label{sec:latent_dynamics}

\begin{figure}[H]
     \centering
     \begin{subfigure}[b]{0.45\linewidth}
         \centering
         \includegraphics[width=\linewidth]{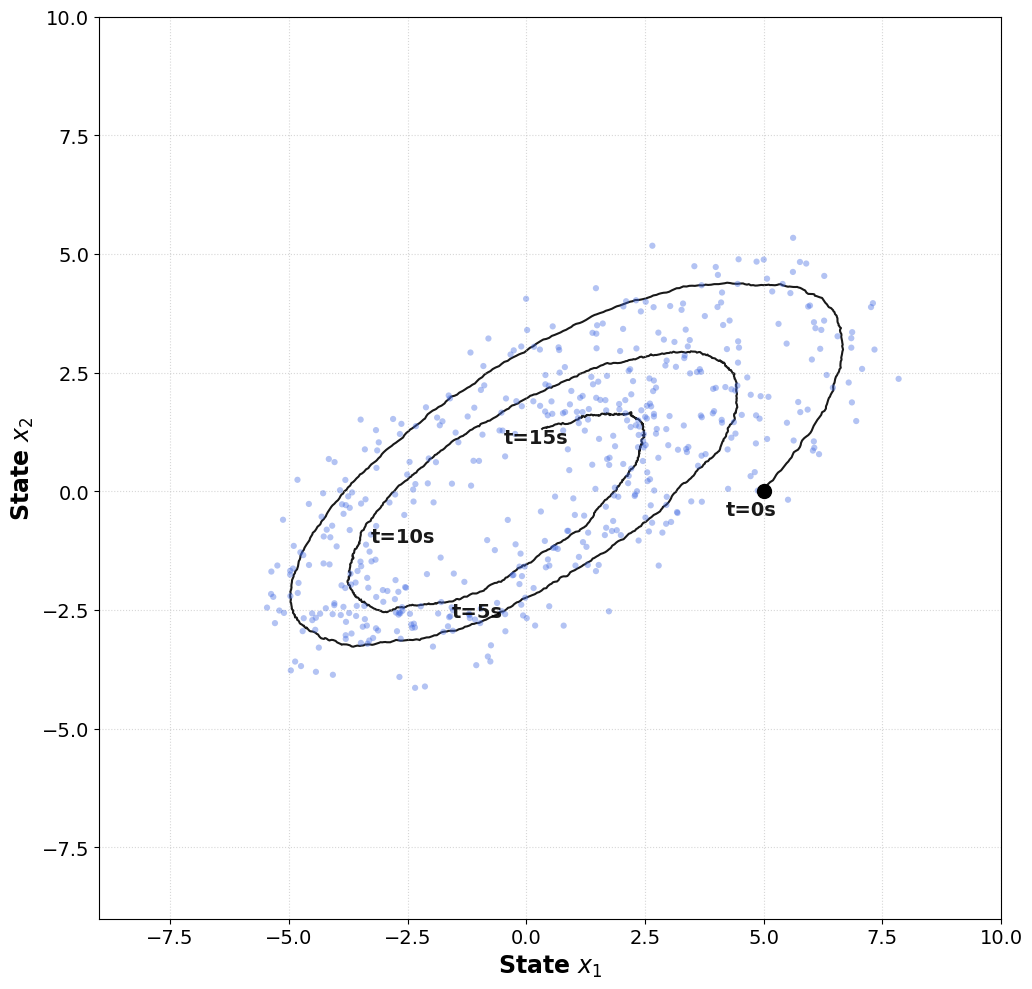}
         \caption{Latent trajectory and noisy observations}
         \label{fig:measurements_raw}
     \end{subfigure}
     \hfill
     \begin{subfigure}[b]{0.45\linewidth}
         \centering
         \includegraphics[width=\linewidth]{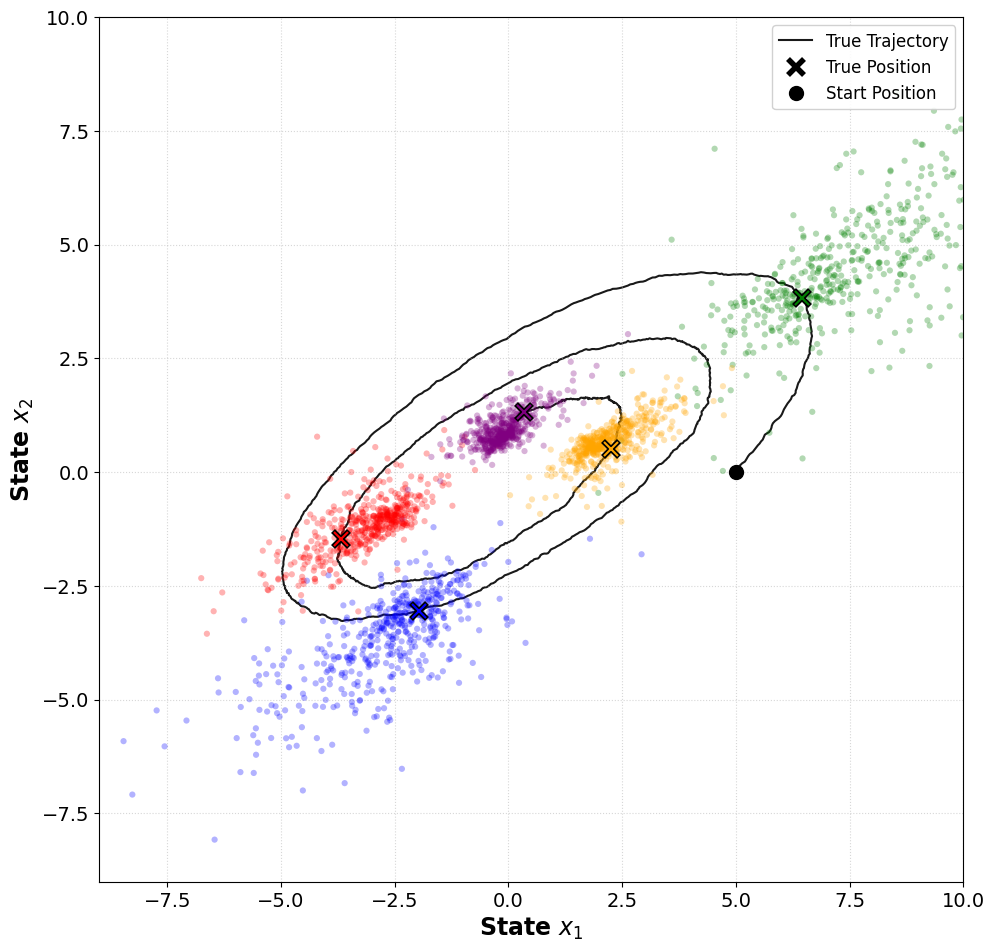}
         \caption{Propagated measurements at 5 target points}
         \label{fig:estimates_causal}
     \end{subfigure}
     \caption{Illustration of an LTI SDE in 2D. \textbf{(a)} The true trajectory (black) is observed through additive Gaussian measurement noise $\boldsymbol{\varepsilon}_i \sim \mathcal{N}(0, \boldsymbol{R})$; noisy measurements $\boldsymbol{z}_{i}$ are shown as blue points. \textbf{(b)} For five target times $t_i$, the plot shows the ensemble of transported keys $\hat{\boldsymbol{x}}_{ij}$ mapped through the transition $e^{\boldsymbol{A}\Delta t_{ij}}$ from all noisy measurements $\boldsymbol{z}_{j}$ (where we have used $\boldsymbol{C}=\boldsymbol{I}$). Both forward and backward transported $\hat{\boldsymbol{x}}_{ij}$ are shown for visualization; estimation is causal and uses only $j \le i$.}
     \label{fig:state_ests}
\end{figure}

The model of Eq.~\ref{eq:SDE} is a special case of a linear time-varying
system, which we develop here before specializing to the time-invariant case
used in the main text:
\[
\dot{\boldsymbol x}(t)=\boldsymbol A(t)\boldsymbol x(t)+\boldsymbol f(t),
\qquad
\boldsymbol z_i=\boldsymbol C_i\big(\boldsymbol x(t_i)+\boldsymbol\varepsilon_i\big),
\qquad
\boldsymbol\varepsilon_i\sim\mathcal N(\boldsymbol 0,\boldsymbol R_i),
\]
where $\boldsymbol x(t)$ is the latent state, $\boldsymbol A(t)$ the nominal
dynamics, $\boldsymbol\varepsilon_i$ Gaussian measurement noise, 
and $\boldsymbol f(t)$ is unmodeled forcing: whatever drives the state but is
not captured by the linear dynamics, such as nonlinear effects or new
information introduced by each token. Let $\boldsymbol\Phi(t,s)$ denote the
state-transition matrix associated with $\boldsymbol A(t)$, satisfying
$\partial_t\boldsymbol\Phi(t,s)=\boldsymbol A(t)\boldsymbol\Phi(t,s)$ and
$\boldsymbol\Phi(s,s)=\boldsymbol I$. By variation of constants, the state
propagated from time $s$ to time $t$ is
\[
\boldsymbol x(t)=\boldsymbol\Phi(t,s)\boldsymbol x(s)+\boldsymbol F(t,s),
\qquad
\boldsymbol F(t,s)=\int_s^t\boldsymbol\Phi(t,\tau)\boldsymbol f(\tau)\,d\tau ,
\]
so transporting under the nominal dynamics alone incurs the accumulated
forcing $\boldsymbol F(t,s)$ as prediction error. Its realization is unavailable, so we describe only its typical size. Taking
$\boldsymbol f$ to be white noise,
$\boldsymbol f(\tau)\,d\tau=\boldsymbol G(\tau)\,d\boldsymbol W(\tau)$, with
$\boldsymbol W$ a Wiener process and
$\boldsymbol Q(t)=\boldsymbol G(t)\boldsymbol G(t)^\top$, makes the model an
It\^o SDE, under which the accumulated forcing is Gaussian,
\[
\boldsymbol F(t,s)\sim\mathcal N\big(\boldsymbol 0,\boldsymbol V(t,s)\big),
\qquad
\boldsymbol V(t,s)=\int_s^t\boldsymbol\Phi(t,\tau)\boldsymbol Q(\tau)\boldsymbol\Phi(t,\tau)^\top d\tau .
\]
Thus, $\boldsymbol V(t,s)$ is the budgeted uncertainty accumulated when
transporting a state from time $s$ to time $t$. The latent scales of
Section~\ref{sec:Setup} then account for departures of the realized forcing
from this budget. The main text specializes to the linear time-invariant case
$\boldsymbol A(t)=\boldsymbol A$, $\boldsymbol Q(t)=\boldsymbol Q$,
$\boldsymbol C_i=\boldsymbol C$, $\boldsymbol R_i=\boldsymbol R$, for which
$\boldsymbol\Phi(t,s)=e^{\boldsymbol A(t-s)}$. For token positions $j\le i$,
the transported prediction is therefore
$\hat{\boldsymbol x}_{ij}=e^{\boldsymbol A\Delta t_{ij}}\hat{\boldsymbol x}_j$.
Its error relative to the current state consists of the forcing accumulated
over the lag and the transported measurement noise, with covariance
$\boldsymbol\Sigma_{ij}$ given by Eq.~\ref{eq:measurement}.
Figure~\ref{fig:state_ests} illustrates this for a stable
two-dimensional LTI system.
Appendix~\ref{sec:content_dependent_dynamics} returns to the time-varying
case, where the transition and the accumulated covariance are obtained from
the corresponding discrete recursions rather than in closed form in the lag.

\subsection{Analytical Solution of the Differential Lyapunov Equation (DLE)}
\label{sec:DLESolution}

For parallel aggregation across all token pairs, we require an efficient
evaluation of $\boldsymbol{\Sigma}_{ij}$. Under the LTI assumptions above,
the covariance satisfies the differential Lyapunov equation
\[
\frac{d\boldsymbol{V}(\Delta t)}{d\Delta t}
=
\boldsymbol{A}\boldsymbol{V}(\Delta t)
+
\boldsymbol{V}(\Delta t)\boldsymbol{A}^{\top}
+
\boldsymbol{Q},
\qquad
\boldsymbol{V}(0)=\boldsymbol{0},
\]
whose solution is
\[
\boldsymbol{V}(\Delta t)
=
\int_0^{\Delta t}
e^{\boldsymbol{A}s}
\boldsymbol{Q}
e^{\boldsymbol{A}^{\top}s}
\,ds.
\]
To obtain an analytically tractable expression, we assume the system matrices
are simultaneously diagonalizable by an invertible
$\boldsymbol{S}\in\mathbb{C}^{d\times d}$, such that
$\boldsymbol{A}=\boldsymbol{S}\boldsymbol{\Lambda}\boldsymbol{S}^{-1}$ and
$\boldsymbol{Q}=\boldsymbol{S}\boldsymbol{\Lambda}_Q\boldsymbol{S}^{\dagger}$.
This corresponds to learning the dynamics in a basis of decoupled modes. Transforming to this eigenbasis gives
\(
\boldsymbol{V}(\Delta t)
=
\boldsymbol{S}
\boldsymbol{\Lambda}_V(\Delta t)
\boldsymbol{S}^{\dagger},
\)
where each diagonal entry satisfies
\[
\lambda_{V,k}(\Delta t)
=
\lambda_{Q,k}
\int_0^{\Delta t}
e^{(\lambda_k+\lambda_k^*)s}\,ds
=
\lambda_{Q,k}
\int_0^{\Delta t}
e^{2\operatorname{Re}(\lambda_k)s}\,ds.
\]
Thus, each mode accumulates uncertainty according to its real decay rate
$\operatorname{Re}(\lambda_k)$. Evaluating the integral gives
\[
\boldsymbol{\Lambda}_V(\Delta t)
=
\operatorname{diag}
\!\left(
\varphi(\lambda_k,\lambda_{Q,k},\Delta t)
\right)_{k=1}^{d},
\qquad
\varphi(\lambda,\lambda_Q,\Delta t)
=
\begin{cases}
\lambda_Q
\dfrac{1-e^{2\operatorname{Re}(\lambda)\Delta t}}
{-2\operatorname{Re}(\lambda)},
&
\operatorname{Re}(\lambda)\neq 0,
\\[6pt]
\lambda_Q\,\Delta t,
&
\operatorname{Re}(\lambda)=0.
\end{cases}
\]

\subsection{Robust Precision-Weighted State Estimation}
\label{sec:appendix_robust_estimation}

This section expands the derivation of the robust estimator of
Section~\ref{sec:Setup}.

\paragraph{Composite likelihood.}
The measurements are correlated through shared forcing. For $j<j'\le i$, the
forcing over $[t_j,t_i]$ contains the forcing over $[t_{j'},t_i]$,
$\boldsymbol F_{ij}=\boldsymbol\Phi(t_i,t_{j'})\boldsymbol F(t_{j'},t_j)+\boldsymbol F_{ij'}$,
so under the flat prior,
\[
\mathrm{Cov}\big(\hat{\boldsymbol x}_{ij},\hat{\boldsymbol x}_{ij'}\mid\boldsymbol x_i\big)
=\boldsymbol V_{ij'} .
\]
This covariance does not decay with lag: two distant measurements share
nearly all of their forcing. The exact joint likelihood accounts for this
through the full covariance of the measurements. We instead approximate it by
the product of the exact marginals,
\[
p\big(\{\hat{\boldsymbol x}_{ij}\}_{j\le i}\mid\boldsymbol x_i\big)
\approx\prod_{j\le i}p(\hat{\boldsymbol x}_{ij}\mid\boldsymbol x_i),
\]
a composite likelihood \citep{lindsay1988composite}. Each factor is exact,
but the product treats the measurements as independent, so it misallocates
weight among correlated measurements and overstates the accumulated
precision. Since the number of measurements sharing a given interval grows
with the context, so does the overcount. Since each measurement is unbiased for $\boldsymbol x_i$, the overcount costs
efficiency rather than validity: any normalized weighting with fixed weights
remains unbiased \citep{varin2011overview}.

\paragraph{The conditional problem.}
Conditional on the latent scales $\boldsymbol s_i=\{s_{ij}\}_{j\le i}$, each
factor is Gaussian, and the negative log composite likelihood is the
precision-weighted quadratic
\[
\mathcal L(\boldsymbol x\mid\boldsymbol s_i)
=\tfrac12\sum_{j\le i}s_{ij}^{-1}d_{ij}^2(\boldsymbol x)+\text{const},
\qquad
d_{ij}^2(\boldsymbol x)
=(\boldsymbol x-\hat{\boldsymbol x}_{ij})^\top\boldsymbol P_{ij}(\boldsymbol x-\hat{\boldsymbol x}_{ij}),
\]
whose minimizer is the Gaussian consensus estimate
\[
\bar{\boldsymbol x}_i
=\bigg(\sum_{j\le i}s_{ij}^{-1}\boldsymbol P_{ij}\bigg)^{-1}
\sum_{j\le i}s_{ij}^{-1}\boldsymbol P_{ij}\hat{\boldsymbol x}_{ij}.
\]
The scales are unobserved, so this estimator is not available. A scale shared
across measurements would cancel in the normalization; it is the variation
across pairs that produces the reweighting.

\paragraph{Marginalization.}
Integrating each scale out under
$s_{ij}\sim\mathrm{Inv\text{-}Gamma}(\nu/2,\nu/2)$,
\[
p(\hat{\boldsymbol x}_{ij}\mid\boldsymbol x)
=\int_0^\infty\mathcal N\big(\hat{\boldsymbol x}_{ij};\boldsymbol x,s\boldsymbol\Sigma_{ij}\big)\,
\mathrm{Inv\text{-}Gamma}(s;\nu/2,\nu/2)\,ds
\;\propto\;
|\boldsymbol\Sigma_{ij}|^{-1/2}\bigg(1+\frac{d_{ij}^2(\boldsymbol x)}{\nu}\bigg)^{-(\nu+d)/2},
\]
a $d$-dimensional Student-$t$ with $\nu$ degrees of freedom. The marginal
objective is therefore
\[
\mathcal L(\boldsymbol x)
=\frac{\nu+d}{2}\sum_{j\le i}\log\!\bigg(1+\frac{d_{ij}^2(\boldsymbol x)}{\nu}\bigg)+\text{const}.
\]
Since
$\nabla_{\boldsymbol x}d_{ij}^2=2\boldsymbol P_{ij}(\boldsymbol x-\hat{\boldsymbol x}_{ij})$,
\[
\nabla_{\boldsymbol x}\mathcal L
=\sum_{j\le i}w_{ij}\boldsymbol P_{ij}(\boldsymbol x-\hat{\boldsymbol x}_{ij}),
\qquad
w_{ij}=\frac{\nu+d}{\nu+d_{ij}^2(\boldsymbol x)},
\]
and setting this to zero gives Eq.~\ref{eq:precision_weighted_average}.

\paragraph{Relation to expectation--maximization.}
The weight $w_{ij}$ is exactly the posterior mean inverse scale. By
conjugacy,
\[
p(s\mid\hat{\boldsymbol x}_{ij},\boldsymbol x)
\;\propto\;
\underbrace{s^{-d/2}e^{-d_{ij}^2(\boldsymbol x)/2s}}_{\text{likelihood}}
\cdot
\underbrace{s^{-\nu/2-1}e^{-\nu/2s}}_{\text{prior}}
\;\propto\;
s^{-\frac{\nu+d}{2}-1}\exp\!\bigg(-\frac{\nu+d_{ij}^2(\boldsymbol x)}{2s}\bigg),
\]
so $s_{ij}\mid\hat{\boldsymbol x}_{ij},\boldsymbol x
\sim\mathrm{Inv\text{-}Gamma}\big(\tfrac{\nu+d}{2},\tfrac{\nu+d_{ij}^2(\boldsymbol x)}{2}\big)$
and $\mathbb E[s_{ij}^{-1}\mid d_{ij}^2(\boldsymbol x)]=w_{ij}$. Because
Eq.~\ref{eq:conditional_objective} is linear in $s_{ij}^{-1}$, the E-step of
an EM algorithm for the marginal likelihood replaces each $s_{ij}^{-1}$ by
this posterior mean, evaluated at the current estimate
$\boldsymbol x^{(k)}$, and the M-step solves the resulting weighted
least-squares problem \citep{Lange1989RobustSM}. The fixed points of the
iteration are the stationary points of Eq.~\ref{eq:marginal_objective}. A
single attention layer performs one iteration, initialized at the query
$\hat{\boldsymbol x}_i$; Appendix~\ref{sec:EM} interprets depth as successive
iterations.

\paragraph{Sharpness exponent and dimension normalization.}
As $\nu\to\infty$, the EM weight tends to $1$, and the Gaussian consensus
estimate has no residual-dependent weighting. We therefore generalize the
weight to
\(
w_{ij}\propto(1+d_{ij}^2/\nu)^{-\kappa},
\)
where $\kappa$ controls how sharply influence falls off with the residual.
The exponent is fixed by requiring that the logits not grow with the feature
dimension, the consideration that motivates the $\sqrt d$ scaling in standard
attention. For large $\nu$ the logits are of order $\kappa d_{ij}^2/\nu$,
while $d_{ij}^2$ accumulates $d$ independent contributions, and setting
$\kappa=(\nu+d)/d$ normalizes them by $d$. With this choice,
\[
\lim_{\nu\to\infty}\bigg(1+\frac{d_{ij}^2}{\nu}\bigg)^{-\frac{\nu+d}{d}}
=\exp\!\bigg(-\frac{d_{ij}^2}{d}\bigg),
\]
so $\nu$ interpolates between a power-law and an exponential kernel at fixed
logit scale. In the isotropic case $\boldsymbol\Sigma_{ij}=\Sigma_{ij}\boldsymbol I$,
the choice has a direct interpretation: raising the Student-$t$ likelihood to
the power $2/d$ gives
\[
p(\hat{\boldsymbol x}_{ij}\mid\boldsymbol x)^{2/d}
\;\propto\;
\Sigma_{ij}^{-1}\bigg(1+\frac{d_{ij}^2}{\nu}\bigg)^{-\frac{\nu+d}{d}}
=P_{ij}\,w_{ij},
\]
so the combined weight is the likelihood of the measurement at temperature
$d/2$. In the anisotropic case, the determinant contributes the geometric
mean of the per-mode precisions rather than $\boldsymbol\Lambda_{P,ij}$, and
the identity does not hold. A degenerate mixing density, $s_{ij}\equiv1$, is
a distinct limit: it removes the scales entirely, giving $w_{ij}\equiv1$ and
the unweighted Gaussian consensus.

\paragraph{Equivalence to M-estimation.}
The weight is the influence function of a robust penalty, so the estimator is
equivalently an M-estimator solved by iteratively reweighted least squares
\citep{huber1964}:
\[
\boldsymbol x_i^+
=\arg\min_{\boldsymbol x}\sum_{j\le i}\rho_\kappa\big(d_{ij}^2(\boldsymbol x)\big),
\qquad
w=\rho_\kappa'.
\]
Integrating the weight gives the penalty
\[
w(u)=\Big(1+\frac{u}{\nu}\Big)^{-\kappa},
\qquad
\rho_\kappa(u)=\frac{\nu}{\kappa-1}\Big[1-\Big(1+\frac{u}{\nu}\Big)^{1-\kappa}\Big],
\]
where $\nu>0$ controls the transition scale and $\kappa>1$ the tail decay.
With $\kappa=(\nu+d)/d$, this becomes
$\rho_\kappa(u)=d\big[1-(1+u/\nu)^{-\nu/d}\big]$. Integrating the limiting
kernel gives the exponential, or Welsch, penalty
\[
w(u)=\exp(-u/d),
\qquad
\rho_{\mathrm{exp}}(u)=d\big(1-\exp(-u/d)\big).
\]
Both penalties are bounded, so their influence redescends to zero: a
sufficiently inconsistent measurement is rejected rather than merely
downweighted. The power-law kernel redescends polynomially and retains
nonvanishing influence at large residuals, while the exponential kernel
discounts them faster than any power law.

\subsection{Parallel Aggregation via Diagonalization}

To obtain a scalable implementation, we transform the robust
precision-weighted average to the diagonalized basis, defining
\[
\hat{\boldsymbol{x}}_{s,i} := \boldsymbol{S}^{-1} \hat{\boldsymbol{x}}_i,
\qquad
\hat{\boldsymbol{x}}_{s,ij}
:= e^{\boldsymbol{\Lambda}\Delta t_{ij}}\, \hat{\boldsymbol{x}}_{s,j},
\qquad
\boldsymbol{r}_{s,ij} = \hat{\boldsymbol{x}}_{s,i} -
\hat{\boldsymbol{x}}_{s,ij}.
\]
Using the simultaneous diagonalization
\(
\boldsymbol{P}_{ij}
=
\boldsymbol{S}^{-\dagger}
\boldsymbol{\Lambda}_{P,ij}
\boldsymbol{S}^{-1},
\)
the Mahalanobis distance decomposes into independent scalar components,
\[
d_{ij}^2
=
\boldsymbol{r}_{s,ij}^{\dagger}
\boldsymbol{\Lambda}_{P,ij}
\boldsymbol{r}_{s,ij}
=
\sum_{k=1}^{d}
\lambda_{P,ij,k}\,
\big| r_{s,ij,k} \big|^2 ,
\]
where $\lambda_{P,ij,k}$ and $r_{s,ij,k}$ are the $k$th
components of $\boldsymbol{\Lambda}_{P,ij}$ and
$\boldsymbol{r}_{s,ij}$. The robust weights $w_{ij}=w(d_{ij}^2)$ can
then be computed for all token pairs in this basis. Since all matrices are diagonal, the aggregation reduces to element-wise
operations,
\[
\bar{\boldsymbol{x}}_{s,i}
=
\sum_{j \le i}
\mathcal{A}_{ij} \odot \hat{\boldsymbol{x}}_{s,ij},
\qquad
\mathcal{A}_{ij}
=
w_{ij} \boldsymbol{\Lambda}_{P,ij}
\oslash
\bigg(
\sum_{j' \le i} w_{ij'} \boldsymbol{\Lambda}_{P,ij'}
\bigg),
\]
where $\oslash$ denotes element-wise division, so that each $\mathcal{A}_{ij}$
is a vector rather than a scalar attention weight. The output is recovered by
\(
\bar{\boldsymbol{z}}_i = \boldsymbol{C} \boldsymbol{S}\,
\bar{\boldsymbol{x}}_{s,i}.
\)
All operations are $\mathcal{O}(d)$ per token pair, giving overall complexity
$\mathcal{O}(N^2 d)$ with no matrix inversions. Equivalently, the normalization may be written as a softmax over dimension-wise
logits,
\[
\mathcal{A}_{ij,k}
=
\frac{\exp(\mathcal{L}_{ij,k})}
{\sum_{j' \le i} \exp(\mathcal{L}_{ij',k})},
\qquad
\mathcal L_{ij,k}
=
\log\lambda_{P,ij,k}
+
\log w_{ij},
\]
which for the power-law and exponential kernels gives
\[
\mathcal{L}_{ij,k}
=
\log \lambda_{P,ij,k}
- \kappa \log\!\big(1 + d_{ij}^2/\nu \big),
\qquad
\mathcal{L}_{ij,k}
=
\log \lambda_{P,ij,k}
- d_{ij}^2/d ,
\]
respectively. Note that only the additive term carries a mode index; the
residual-dependent term is shared across modes.

\paragraph{Remark.} The normalization term
$\big(\sum_{j'\le i} w_{ij'}\boldsymbol{\Lambda}_{P,ij'}\big)^{-1}$ corresponds to the approximate posterior covariance of the state estimate in
the diagonalizing basis under the composite likelihood. Because the composite
likelihood neglects correlations from shared forcing, this covariance is
optimistic, and increasingly so as the context grows
(Appendix~\ref{sec:appendix_robust_estimation}). This quantity is a byproduct of the formulation that goes unused in standard attention, and, suitably calibrated, may be useful beyond weighting: for instance, to gate updates or to signal when the context provides insufficient evidence. We
leave exploration of such uses to future work.

\newpage

\section{Robust Filter Attention Mechanism}
\label{sec:RFA_Mechanism}

We instantiate the state estimation formulation as an attention mechanism, moving from the full anisotropic formulation to the memory-efficient isotropic variant used in practice.

\subsection{Anisotropic Tensor RFA}
\label{sec:Tensor_RFA}

Under diagonalizable dynamics, the most general RFA formulation propagates and weights each feature dimension independently, yielding an $\mathcal{O}(N^2 d)$ attention tensor. This is not memory-scalable but serves as the reference from which the isotropic variant is derived.

\paragraph{Learned Change-of-Basis Projections.}

The transformation to the decoupled eigenbasis is learned through complex-valued projections. We define:
\[
\boldsymbol{W}_q, \,
\boldsymbol{W}_k, \,
\boldsymbol{W}_v, \,
\boldsymbol{W}_o \in \mathbb{C}^{d \times d},
\]
where $d$ is the embedding dimension. The input projections
$\{\boldsymbol{W}_q, \boldsymbol{W}_k, \boldsymbol{W}_v\}$ 
parameterize the learned diagonalizing basis $\boldsymbol{S}^{-1} \boldsymbol{C}^{-1}$, mapping the input into the eigenbasis of $\boldsymbol{A}$ where the DLE is analytically solvable, while the output projection $\boldsymbol{W}_o$ parameterizes $\boldsymbol{C} \boldsymbol{S}$, mapping the filtered state estimate back into the original embedding space.

Given an input sequence $\boldsymbol{Z} \in \mathbb{R}^{d \times N}$, we obtain query, key, and value representations as usual:
\[
\boldsymbol{Q} = \boldsymbol{W}_q \boldsymbol{Z}, \quad
\boldsymbol{K} = \boldsymbol{W}_k \boldsymbol{Z}, \quad
\boldsymbol{V} = \boldsymbol{W}_v \boldsymbol{Z}.
\]

\paragraph{Key and Value Propagation.}

% There are two precision matrices, in the prior and in the Mahalanobis distance, which correspond to the value and the query/key spaces, respectively, so in principle we could define separate dynamics and noise parameters $\{\boldsymbol{\lambda}, \boldsymbol{\lambda}_Q, \boldsymbol{\lambda}_R, \boldsymbol{\lambda}_{\Gamma} \}$ for each. For simplicity and parameter efficiency, we instead use a shared embedding dimension $d$ and a unified set of parameters.

In this reference model, every feature $k$ possesses its own complex eigenvalue $\lambda_k$. We define the propagation tensors $\mathcal{E}$ and the resulting propagated keys and values:
\[
\mathcal{E}[k,i,j] = e^{\boldsymbol{\lambda}_{k} (t_i - t_j)}, \qquad
\hat{\mathcal{K}}[k,i,j] = \mathcal{E}[k,i,j] \cdot \boldsymbol{K}[k,j], \qquad
\hat{\mathcal{V}}[k,i,j] = \mathcal{E}[k,i,j] \cdot \boldsymbol{V}[k,j].
\]

\paragraph{Residuals \& Precision.}

We compute the residual tensor $\mathcal{R}_{qk}$:
\[
\mathcal{R}_{qk}[k,i,j] = \boldsymbol{Q}[k,i] - \hat{\mathcal{K}}[k,i,j],
\]
This is weighted by the analytic precision tensor $\mathcal{P}$, which is defined element-wise for each channel $k$ using the DLE solution, where $\mu_{k} := -\mathrm{Re}(\lambda_{k})$:
\[
\mathit{\Sigma}[k,i,j] =
\tilde{\sigma}_{k}^2 \big( 1 - e^{-2\mu_{k} \Delta t_{ij}} \big) 
+ \eta_{k}^2 e^{-2\mu_{k} \Delta t_{ij}} + \gamma^2,
\]
\[
\mathcal{P}[k,i,j] = \mathit{\Sigma}[k,i,j]^{-1}.
\]

\paragraph{Aggregation.}

Unlike standard attention, which applies a single scalar score per head, tensor RFA computes an attention tensor 
$\mathcal{A} \in \mathbb{R}^{d \times N \times N}$. The logit tensor is then:
\[
\mathcal{L}[k,i,j] = \log \big(\mathcal{P}[k,i,j] \big) - \kappa \log \left( 1 + \frac{1}{\nu} \sum_{k'} \mathcal{P}[k',i,j] \cdot \bigl|\mathcal{R}_{qk}[k',i,j]\bigr|^{2} \right).
\]
The estimate in the eigenbasis is computed via a row-wise softmax over the logits, followed by a weighted sum:
\[
\mathcal{A}[k,i,j] = \text{softmax}_j(\mathcal{L}[k,i,j]), \quad \bar{\boldsymbol{V}}[k,i] = \sum_{j \leq i} \mathcal{A}[k,i,j] \cdot \mathcal{\hat{V}}[k,i,j].
\]
The time complexity remains \(\mathcal{O}(N^2 d)\), but storing the propagated keys and values, residuals, and attention tensors requires \(\mathcal{O}(N^2 d)\) memory, limiting scalability. We therefore derive a memory-efficient implementation that avoids storing tensors.

\subsection{Factorization and Complexity Reduction}
\label{sec:Complexity_Reduction}

We introduce the following factorizations to simplify the computation.

\subsubsection{Toeplitz Kernel for Precision}

If the measurements occur at equal time intervals $\delta t$, the covariance tensor $\mathit{\Sigma}[k,i,j]$ depends only on the channel $k$ and the time lag $\tau = |i-j|$. This induces a Toeplitz structure along the temporal dimensions for each channel.

Letting $ \Delta t_{ij} = \tau \delta t$, we can thus pre-compute a 1D covariance kernel: 
$\mathcal{K}_{\Sigma} \in \mathbb{R}^{d \times N}$:
\[
\mathcal{K}_{\Sigma}[k, \tau] =
\tilde{\sigma}_{k}^{2} \, 
\big( 1 - e^{-2\mu_{k} \delta t \, \tau} \big)
+ \eta_{k}^{2} e^{-2\mu_{k} \delta t \, \tau} + \gamma^2 .
\]
The full precision tensor is then simply the element-wise inverse of this kernel:
\[
\mathcal{P}[k,i,j] = \mathcal{K}_{P}[k,|i-j|] := 1/\mathcal{K}_{\Sigma}[k,|i-j|] .
\]

\subsubsection{Transport Through a Common Reference Frame}

For an LTI system, the state transition depends only on the time difference,
\(
e^{\boldsymbol{\Lambda}(t_i-t_j)}
\). Using the fact that
\(
e^{\boldsymbol{\Lambda}(t_i-t_j)}
=
e^{\boldsymbol{\Lambda}t_i}
e^{-\boldsymbol{\Lambda}t_j},
\)
the propagation can be separated into two stages: a backward transport that maps every measurement into a common stationary reference frame ($t=0$), followed by a forward transport into the query time.

\begin{figure}[t]
    \centering
    \includegraphics[width=0.9\linewidth]{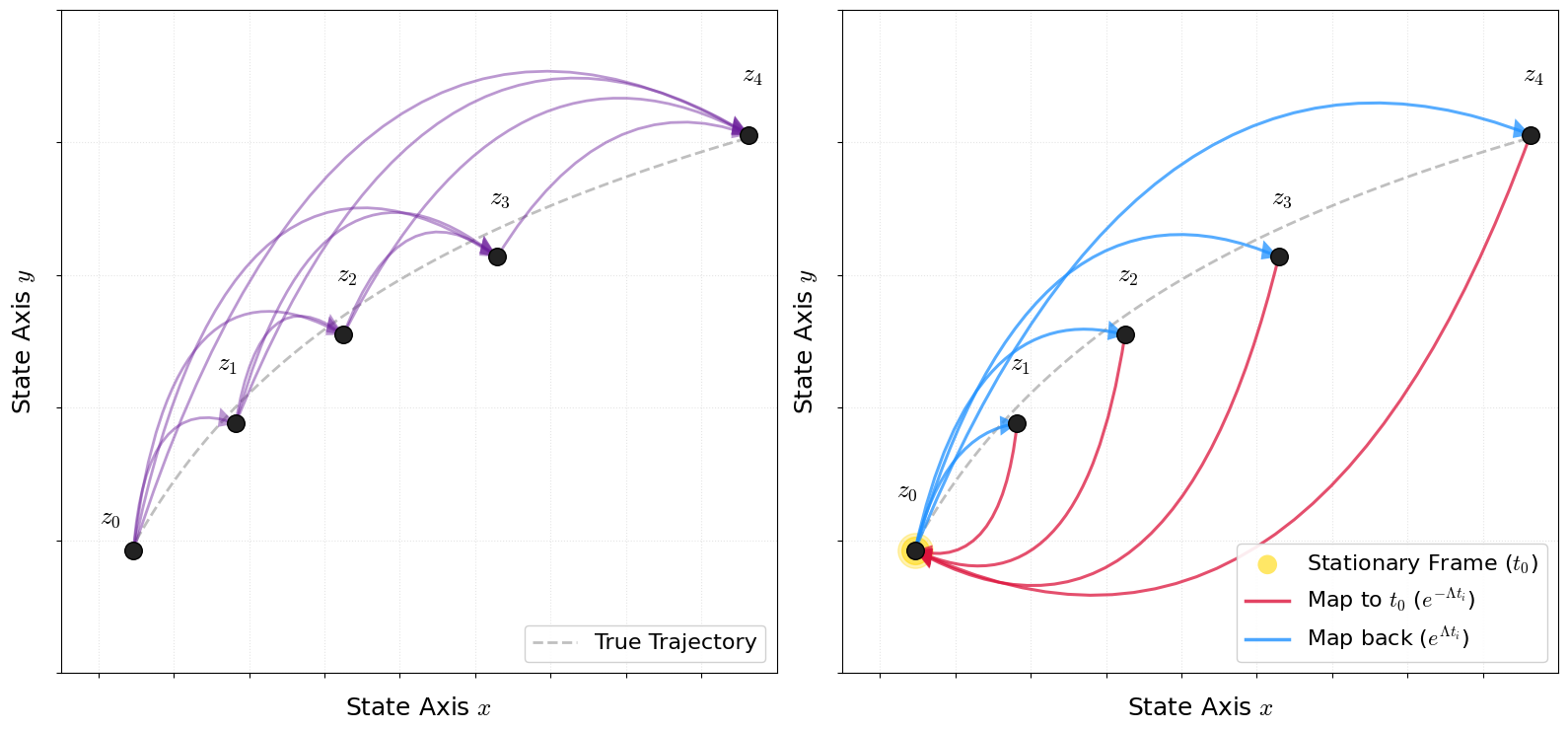}
    \caption{
    \textbf{Factorizing state propagation through a common reference frame.}
    \textbf{Left:} Direct propagation requires a distinct transport operator
    $e^{\boldsymbol{\Lambda}(t_i-t_j)}$
    for every source--destination pair $(i,j)$.
    \textbf{Right:}
    Using
    $
    e^{\boldsymbol{\Lambda}(t_i-t_j)}
    =
    e^{\boldsymbol{\Lambda}t_i}
    e^{-\boldsymbol{\Lambda}t_j},
    $
    each representation is first transported once into the stationary frame
    ($t=0$) via
    $
    e^{-\boldsymbol{\Lambda}t_j},
    $
    producing reusable stationary representations. Each query then applies its own forward transport
    $
    e^{\boldsymbol{\Lambda}t_i}
    $
    after attention aggregation to recover the output in the query frame. This replaces $N^2$ pairwise transport operators with two sets of $N$ transport factors.
    }
    \label{fig:transport_factorization}
\end{figure}

Accordingly, we define the forward and backward transport factors
\[
\boldsymbol{\Phi}^+[k,i]
:=
e^{\lambda_k t_i},
\qquad
\boldsymbol{\Phi}^-[k,j]
:=
e^{-\lambda_k t_j},
\]
so that the propagation tensor factorizes as
\[
\mathcal{E}[k,i,j]
=
\boldsymbol{\Phi}^+[k,i]\,
\boldsymbol{\Phi}^-[k,j].
\]
Rather than constructing a distinct transport operator for every pair $(i,j)$, each key, query, and value is transported only once into the stationary frame:
\[
\hat{\boldsymbol{Q}}[k,j]
:=
\boldsymbol{\Phi}^-[k,j]\,
\boldsymbol{Q}[k,j],
\qquad
\hat{\boldsymbol{K}}[k,j]
:=
\boldsymbol{\Phi}^-[k,j]\,
\boldsymbol{K}[k,j],
\qquad
\hat{\boldsymbol{V}}[k,j]
:=
\boldsymbol{\Phi}^-[k,j]\,
\boldsymbol{V}[k,j].
\]
The propagated keys and values are then recovered by applying the query-dependent forward transport,
\[
\hat{\mathcal{K}}[k,i,j]
=
\boldsymbol{\Phi}^+[k,i]\,
\hat{\boldsymbol{K}}[k,j],
\qquad
\hat{\mathcal{V}}[k,i,j]
=
\boldsymbol{\Phi}^+[k,i]\,
\hat{\boldsymbol{V}}[k,j].
\]
Since $\boldsymbol{\Phi}^+[k,i]$ is independent of the source index $j$, it can be pulled outside the attention aggregation:
\[
\bar{\boldsymbol{V}}[k,i]
=
\sum_{j\le i}
\mathcal{A}[k,i,j]\,
\hat{\mathcal{V}}[k,i,j]
=
\boldsymbol{\Phi}^+[k,i]
\sum_{j\le i}
\mathcal{A}[k,i,j]\,
\hat{\boldsymbol{V}}[k,j].
\]
Thus, transport is performed by mapping all representations into a common stationary frame, performing attention there, and applying a single forward transport to recover the output in the query frame. Figure~\ref{fig:transport_factorization} illustrates this decomposition.

\subsubsection{Memory Efficiency and Numerical Stability}
Recall that:
\[
\mathcal{R}_{qk}[k,i,j] = \boldsymbol{Q}[k,i] - \hat{\mathcal{K}}[k, i, j] 
\]
Plugging in the factorizations for $\hat{\boldsymbol{Q}}[k,j]$ and $\mathcal{\hat{K}}[k,i,j]$, the residual becomes:
\[
\mathcal{R}_{qk}[k,i,j] = \boldsymbol{Q}[k,i] - \boldsymbol{\Phi}^+[k,i] \cdot \hat{\boldsymbol{K}}[k,j] 
= \boldsymbol{\Phi}^+[k,i] \cdot 
\bigl( 
\hat{\boldsymbol{Q}}[k,i] - 
\hat{\boldsymbol{K}}[k,j] 
\bigr).
\]
The matrix of Mahalanobis distances now becomes:
\[
\begin{aligned}
\boldsymbol{D}^{2}[i,j] &= 
\sum_{k} 
\underbrace{\mathcal{K}_{P}[k,|i-j|]}_{\text{Precision kernel}} \cdot 
\underbrace{|\boldsymbol{\Phi}^+[k,i]|^2}_{\text{Forward decay}} \cdot 
\Bigg[
\underbrace{\bigl| \hat{\boldsymbol{Q}}[k,i] \bigr|^{2}}_{\text{Stationary Query}} 
+ 
\underbrace{\bigl| \hat{\boldsymbol{K}}[k,j] \bigr|^{2}}_{\text{Stationary Key}} 
- 
\underbrace{2\,\mathrm{Re} \bigl( \hat{\boldsymbol{Q}}^{*}[k,i] \, \hat{\boldsymbol{K}}[k,j] \bigr)}_{\text{Stationary Cross-term}} 
\Bigg]
\end{aligned}
\]
(where $*$ denotes the complex conjugate). The remaining bottleneck is the $k$-dependence of the precision kernel $\mathcal{K}_{P}$ in the evaluation of the cross-term:
\[
\sum_k \mathcal{K}_{P}[k,|i-j|]  \cdot 2\,\mathrm{Re} \bigl(\hat{\boldsymbol{Q}}^{*}[k,i]\,\hat{\boldsymbol{K}}[k,j]\bigr).
\]
In standard attention, scores are computed with a single matrix multiplication
\((\boldsymbol{Q}\boldsymbol{K}^\top)\). Here, however, the precision kernel \( \mathcal{K}^P[k,|i-j|] \) weights each feature differently as a function of time
lag, so the summation over \(k\) cannot be expressed as a single matmul. Achieving \(\mathcal{O}(N^2 + Nd)\) memory therefore requires the precision kernel to be independent of the feature index \(k\), allowing it to factor outside the summation.

A degenerate case occurs in the zero-noise limit, where the precision kernel is constant. This recovers a memory-efficient formulation with anisotropic (feature-wise) decay, similar to xPos.

However, for stable dynamics with \(\mu_k > 0\), the backward transition factor
\(
\boldsymbol{\Phi}^-[k,j] = e^{(\mu_k - i\omega_k) t_j}
\) grows exponentially with sequence length. Hence, when decay rates vary across features, the stationary representations \( \hat{\boldsymbol{Q}}, \hat{\boldsymbol{K}}, \hat{\boldsymbol{V}} \) grow exponentially with sequence length, making fully parallel computation numerically unstable because forward and backward factors cancel only after
multiplication, allowing intermediate values to overflow.

Therefore, retaining a non-constant precision kernel while ensuring numerical stability under extrapolation requires restricting decay to be isotropic within each head. This allows decay to be factored at the head level rather than per feature, enabling stable, fully parallel attention with \(\mathcal{O}(N^2 + Nd)\) memory. This motivates the Isotropic RFA variant introduced next.

\subsection{Isotropic RFA}
\label{sec:Isotropic_RFA}

\subsubsection{Isotropic Decay and Noise Assumptions}

All assumptions in this section are applied \emph{per attention head}. In particular, the real part of the eigenvalues within a head is taken to be a shared scalar \( -\mu \):
\[
\lambda_k = -\mu + i \omega_k, \qquad \mu \in \mathbb{R}^+, \omega_k \in \mathbb{R}
\]
This corresponds to a system with an isotropic plus skew-symmetric state matrix:
\[
\boldsymbol{A} = -\mu \boldsymbol{I} + \boldsymbol{\Omega} , \qquad \boldsymbol{\Omega} = -\boldsymbol{\Omega}^\top \in \mathbb{R}^{d \times d} .
\]
We also assume that the noise is isotropic, i.e. that the noise covariances are scalar multiples of identity:
\[
\boldsymbol{\Lambda}_{Q} = \sigma^2 \boldsymbol{I}, \quad \boldsymbol{\Lambda}_{R} = \eta^2 \boldsymbol{I}.
\]
Under this constraint, the covariance kernel simplifies to a scalar function of the lag $|i-j|$:
\[
 \boldsymbol{\Sigma}_{\Delta t}[i,j] := \Sigma(|i-j|) = \tilde{\sigma}^{2} \, \big( 1 - e^{-2 \mu \delta t |i-j|}\big) + \eta^{2}  e^{-2 \mu \delta t |i-j|} + \gamma^2.
\]
Hence, the precision kernel becomes a scalar function of the time lag \(\tau = |i-j|\), allowing it to be pulled outside the feature summation. Defining $ \boldsymbol{P}_{\Delta t}[i,j] := 1 / \boldsymbol{\Sigma}_{\Delta t}[i,j]$, the matrix of Mahalanobis distances become:
\[
\boldsymbol{D}^2[i,j] = \boldsymbol{P}_{\Delta t}[i,j] \cdot \Big( \sum_k \big| \mathcal{R}_{qk}[k, i, j] \big|^2 \Big)
=: \boldsymbol{P}_{\Delta t}[i,j] \cdot \big\| \boldsymbol{R}_{qk}[i,j]\big\|^2 .
\]
Note that $\big\| \boldsymbol{R}_{qk}\big\| $ denotes a matrix of vector norms, rather than a matrix norm.

\subsubsection{Simplifying the Squared Residual Norm}
The isotropic constraint allows the dynamics to be factored into a stable decay kernel and complex-valued forward/backward rotations:
\[
\boldsymbol{E}[i,j] = e^{-\mu |t_i - t_j|}, \quad 
\tilde{\boldsymbol{\Phi}}^{+}[k,i] := e^{i \omega_{k} t_i}, \quad 
\tilde{\boldsymbol{\Phi}}^{-}[k,i] := e^{-i \omega_{k} t_i},
\]
We can then define backward-rotated queries, keys, and values:
\[
\tilde{\boldsymbol{Q}} := \tilde{\boldsymbol{\Phi}}^{-} \odot \boldsymbol{Q}, \quad \tilde{\boldsymbol{K}} := \tilde{\boldsymbol{\Phi}}^{-} \odot \boldsymbol{K}, \quad \tilde{\boldsymbol{V}} := \tilde{\boldsymbol{\Phi}}^{-} \odot \boldsymbol{V}, \quad 
\]
Note that:
\[
\boldsymbol{\Phi}^+[k,i] = e^{-\mu t_i} \tilde{\boldsymbol{\Phi}}^+[k,i], \quad
\hat{\boldsymbol{Q}}[k,i] = e^{\mu t_i} \tilde{\boldsymbol{Q}}[k,i], \quad
\hat{\boldsymbol{K}}[k,j] = e^{\mu t_j} \tilde{\boldsymbol{K}}[k,j].
\]
Plugging this into the expression for the Mahalanobis distance, and using the fact that complex rotation preserves magnitude:
\[
\big\| \boldsymbol{R}_{qk}[i,j] \big\|^2
=
\sum_{k} 
e^{-2 \mu t_i}  \bigl| \tilde{\boldsymbol{\Phi}}^+[k,i] \bigr|^{2} 
\cdot 
\Big[ 
e^{2 \mu t_i}  \bigl| \tilde{\boldsymbol{Q}}[k,i] \bigr|^{2} 
+ 
e^{2 \mu t_j} \bigl| \tilde{\boldsymbol{K}}[k,j] \bigr|^{2} 
- 
2 e^{\mu t_i} e^{\mu t_j}  \,\mathrm{Re} \bigl( 
\tilde{\boldsymbol{Q}}^*[k,i] 
\, 
\tilde{\boldsymbol{K}}[k,j]
\bigr)
\Big].
\]
\[
=
\sum_{k}  
\Big[ 
\bigl| \tilde{\boldsymbol{Q}}[k,i] \bigr|^{2} 
+ 
e^{-2 \mu (t_i - t_j)} \bigl| \tilde{\boldsymbol{K}}[k,j] \bigr|^{2} 
- 
2 e^{-\mu (t_i - t_j)} \,\mathrm{Re} \bigl( 
\tilde{\boldsymbol{Q}}[k,i]^* 
\, 
\tilde{\boldsymbol{K}}[k,j]
\bigr).
\]
Or, in vectorized form:
\[
\begin{aligned}
\big\|\boldsymbol{R}_{qk}[i,j]\big\|^{2}
=
\underbrace{\|\boldsymbol{Q}_{i}\|^{2}}_{\text{Query Norm}}
+
\underbrace{\boldsymbol{E}[i,j]^{2} \cdot\|\boldsymbol{K}_{j}\|^{2}}_{\text{Decayed Key Norm}} 
- \underbrace{2\,\boldsymbol{E}[i,j] \cdot
\mathrm{Re}\!\left(\tilde{\boldsymbol{Q}}_{i}^{\dagger}\tilde{\boldsymbol{K}}_{j}\right)}_{\text{Cross-term}}
\end{aligned}
\]
(since $\|\boldsymbol{Q}_{i}\|^{2} = \|\tilde{\boldsymbol{Q}}_{i}\|^{2} $ and $\|\boldsymbol{K}_{j}\|^{2} = \|\tilde{\boldsymbol{K}}_{j}\|^{2} $).
% This sum can be computed efficiently using matrix multiplication and broadcasting:
% \[
% \big\| \boldsymbol{R}_{qk} \big\|^2
% =
% \big( \boldsymbol{1}_{\boldsymbol{d}_{qk}}^{\top} |\boldsymbol{Q}|^2 \big)^{\top} \boldsymbol{1}_m^{\top}
% +
% \boldsymbol{E}^{2} \odot 
% \boldsymbol{1}_m \big( \boldsymbol{1}_{\boldsymbol{d}_{qk}}^{\top} |\boldsymbol{K}|^2 \big)
% -
% 2 \boldsymbol{E} \odot 
% \mathrm{Re}\big( \tilde{\boldsymbol{Q}}^{\dagger} \tilde{\boldsymbol{K}} \big).
% \]
Hence, under the isotropic assumption, the cross-term $\mathrm{Re}(\tilde{\boldsymbol{Q}}^{\dagger} \tilde{\boldsymbol{K}})$ can be computed using one $\mathcal{O}(N^2 d)$ matrix multiplication, achieving the required memory efficiency.

\subsubsection{The Attention Matrix and Estimate}
\label{sec:isotropic_afa_attn_mat}

The logit matrix $\boldsymbol{L}$ is defined using the power-law/Student-$t$ weighting introduced in Section~\ref{sec:method}:
\[
\boldsymbol{L} = \log \big(\boldsymbol{P}_{\Delta t}\big) - (\nu_s+1) \log \Big( 1 + \frac{1}{\nu_s d} \boldsymbol{P}_{\Delta t} \odot \big\| \boldsymbol{R}_{qk} \big\|^2 \Big),
\]
where $\nu_s\in \mathbb{R}^+$ and $d$ is the head dimension.

Defining a causal mask $\boldsymbol{M}_{\text{causal}}$ and adding an inverse temperature parameter $\beta_s$, we can then express the row-normalization using row-wise softmax: 
\[
\boldsymbol{A}[i,j] = \text{softmax}_j \big( \beta_s \boldsymbol{L}[i,j] + \boldsymbol{M}_{\text{causal}} \big).
\]
The value aggregation is refactored for stability:
\[
\bar{\boldsymbol{V}}[k,i] = \left(e^{-\mu t_i} \tilde{\boldsymbol{\Phi}}^{+}[k,i]\right) \cdot \sum_{j \le i} \boldsymbol{A}[i,j] \cdot \left(e^{\mu t_j} \tilde{\boldsymbol{V}}[k,j]\right) \]
\[
= \tilde{\boldsymbol{\Phi}}^{+}[k,i] \cdot \sum_{j \le i} \Big( \boldsymbol{A}[i,j] \cdot \boldsymbol{E}[i,j] \Big) \cdot \tilde{\boldsymbol{V}}[k,j]
\]
Hence, defining a decayed attention matrix \( \hat{\boldsymbol{A}} := \boldsymbol{A} \odot\boldsymbol{E} \), the filtered estimate \(\bar{\boldsymbol{V}}\) is computed by transforming the aggregation back into the original frame:
\[
\bar{\boldsymbol{V}} = \tilde{\boldsymbol{\Phi}}^{+} \odot \big( \tilde{\boldsymbol{V}} \, \hat{\boldsymbol{A}}^{\top} \big).
\]
Or, in a form more typical for attention:
\[
\bar{\boldsymbol{V}}^\top = (\tilde{\boldsymbol{\Phi}}^{+})^\top \odot ( \hat{\boldsymbol{A}} \tilde{\boldsymbol{V}}^\top )
\]
This rotate–aggregate–counter-rotate structure aligns representations in a shared dynamical frame before aggregation. In practice, multi-head attention is implemented by applying this procedure independently across multiple heads, each with its own dynamical and uncertainty parameters. The resulting update steps are concatenated, projected back to the original basis, and added to the residual stream, $\boldsymbol Z^+=\boldsymbol Z+\boldsymbol W_o\bar{\boldsymbol V}$. Figure~\ref{fig:rfa_block} summarizes the complete Isotropic RFA head.

\subsubsection{Residual Connection and Output Projection}
\label{sec:residual_connection}

Our experiments use the standard residual update
$\boldsymbol Z^+=\boldsymbol Z+\boldsymbol W_o\bar{\boldsymbol V}$. Alternatively, the output of an attention layer may be interpreted as a correction relative to the current value representation:
\[
\Delta \boldsymbol{V}
:=
\bar{\boldsymbol{V}} - \boldsymbol{V}.
\]
The residual connection then applies a partial step toward this estimate:
\[
\boldsymbol{V}^{+}
=
\boldsymbol{V}
+
\Delta \boldsymbol{V},
\]
Mapping back to the original basis:
\[
\boldsymbol{Z}^{+}
=
\boldsymbol{W}_{o}\boldsymbol{V}^{+}
=
\boldsymbol{W}_{o}\boldsymbol{V}
+ \boldsymbol{W}_{o}
\Delta \boldsymbol{V}.
\]
The value projection maps residual representations into the latent dynamical
space, and the output projection maps the resulting corrections back, so at
convergence we expect
\(
\boldsymbol{W}_{o}\boldsymbol{W}_{v}
\approx
\boldsymbol{I} 
\)
(per head, the projector onto the head's observable subspace). We instead impose this directly, updating the residual stream as
\[
\boldsymbol{Z}^{+}
=
\boldsymbol{Z}
+
\Delta \boldsymbol{Z},
\qquad
\Delta \boldsymbol{Z} = \boldsymbol{W}_{o} \Delta \boldsymbol{V} .
\]
This substitutes the identity for a product that is only approximately the
identity early in training, keeping the residual path clean and improving
gradient flow through depth. Each attention layer then contributes an incremental correction to the current representation.

% \begin{figure}[H]
%     \centering
%     \includegraphics[width=0.7\linewidth]{images/explanations/RoPER_sandwich.png}
%     \caption{\textbf{Rotate, aggregate, counter-rotate structure of Isotropic RFA.} 
% Queries, keys, and values are rotated into a common frame to compute attention and aggregate values. The resulting estimate is then rotated back to the initial frame, yielding a state that preserves relative phase while remaining equivariant to absolute position.}
%     \label{fig:roper_sandwich}
% \end{figure}

\begin{figure}[H]
  \centering
  \resizebox{\textwidth}{!}{
\begin{tikzpicture}[
  font=\small, >={Stealth[length=2mm]},
  box/.style={draw, rounded corners=2pt, align=center, minimum height=1.05cm, inner sep=3pt, line width=0.6pt},
  std/.style={box, fill=gray!10, draw=gray!60},
  rfa/.style={box, fill=orange!16, draw=orange!75!black},
  dyn/.style={box, fill=blue!7, draw=blue!55!black},
  arr/.style={->, line width=0.7pt, draw=black!75},
  darr/.style={->, line width=0.7pt, draw=blue!55!black},
  skip/.style={->, dashed, line width=0.6pt, draw=black!45},
  lab/.style={font=\scriptsize, inner sep=1pt},
]
% ---------------- query / key path ----------------
\node[std, minimum height=0.8cm] (Z) at (0,-1.3) {$\boldsymbol Z$};
\node[std] (qk)    at (1.8,0)    {$\boldsymbol W_q,\boldsymbol W_k$\\{\scriptsize $d\!\to\!2d$}};
\node[std] (rotqk) at (3.85,0)   {rotate\\$\tilde{\boldsymbol\Phi}^{-}$};
\node[rfa] (res)   at (6.65,0)   {transported\\residual $\|\boldsymbol R_{qk}\|^2$};
\node[rfa] (logit) at (10.3,0)   {robust logits\\$\boldsymbol L=\log\boldsymbol P_{\Delta t}\,+$\\$\log w\big(\boldsymbol P_{\Delta t}\odot\|\boldsymbol R_{qk}\|^2\big)$};
\node[std] (sm)    at (13.9,0)   {normalize\\$\boldsymbol A=\mathrm{softmax}_j(\beta_s\boldsymbol L)$};
\node[rfa] (A)     at (16.85,0)  {decay\\$\hat{\boldsymbol A}=\boldsymbol A\odot\boldsymbol E$};
 
% ---------------- value path ----------------
\node[std] (v)     at (1.8,-2.6)   {$\boldsymbol W_v$\\{\scriptsize $d\!\to\!2d$}};
\node[rfa] (rotv)  at (3.85,-2.6)  {rotate\\$\tilde{\boldsymbol\Phi}^{-}$};
\node[std] (agg)   [right=0.8cm of rotv]   {aggregate\\$\tilde{\boldsymbol V}\hat{\boldsymbol A}^{\top}$};
\node[rfa] (crot)  [right=0.6cm of agg]  {counter-rotate\\$\tilde{\boldsymbol\Phi}^{+}$};
\node[std] (wo)    [right=0.8cm of crot]  {$\boldsymbol W_o$\\{\scriptsize $2d\!\to\!d$}};
\node[std] (out)   [right=0.8cm of wo] {residual update\\$\boldsymbol Z^{+}=\boldsymbol Z+\boldsymbol W_o\bar{\boldsymbol V}$};
 
% ---------------- dynamical model (depends on lag only) ----------------
\node[dyn] (par)   at (8.6,2.55)   {per-head scalars\\$\mu,\ \tilde\sigma^2,\ \eta^2,\ \gamma^2$};
\node[dyn] (E)     [left=0.75cm of par]   {decay kernel\\$\boldsymbol E=e^{-\mu\Delta t}$};
\node[dyn] (P)     [right=0.75cm of par] {DLE precision\\$\boldsymbol P_{\Delta t}=\big(\alpha e^{-2\mu\Delta t}+\beta\big)^{-1}$};
 
% inset: precision as a function of lag in the two regimes
\begin{scope}[shift={(14.6,1.95)}]
  \draw[->, gray!70] (0,0) -- (1.75,0) node[right, font=\scriptsize, text=black, inner sep=1pt] {$\Delta t$};
  \draw[->, gray!70] (0,0) -- (0,1.35) node[left, font=\scriptsize, text=black, inner sep=1pt] {$\boldsymbol P_{\Delta t}$};
  \draw[dashed, gray!60] (0,0.65) -- (1.65,0.65);
  \node[font=\tiny, anchor=west, gray!80!black, inner sep=1pt] at (1.65,0.65) {$1/\beta$};
  \draw[thick, orange!80!black, domain=0:1.6, samples=40] plot (\x, {0.65-0.5*exp(-2.6*\x)});
  \draw[thick, blue!60!black,   domain=0:1.6, samples=40] plot (\x, {0.65+0.55*exp(-2.6*\x)});
  \node[font=\tiny, orange!80!black, anchor=west] at (0.3,0.12) {integrative};
  \node[font=\tiny, blue!60!black,   anchor=west] at (0.3,1.2)  {diffusive};
\end{scope}
 
% ---------------- arrows: features ----------------
\draw[arr] (Z.north) |- (qk.west);
\draw[arr] (Z.south) |- (v.west);
\draw[arr] (qk) -- (rotqk);
\draw[arr] (rotqk) -- node[lab, above] {$\tilde{\boldsymbol Q},\tilde{\boldsymbol K}$} (res);
\draw[arr] (res) -- (logit);
\draw[arr] (logit) -- (sm);
\draw[arr] (sm) -- (A);
\draw[arr] (v) -- (rotv);
\draw[arr] (rotv) -- node[lab, above] {$\tilde{\boldsymbol V}$} (agg);
\draw[arr] (A.south) -- (16.85,-1.3) -- node[lab, above] {$\hat{\boldsymbol A}$} (agg.north |- 0,-1.3) -- (agg.north);
\draw[arr] (agg) -- (crot);
\draw[arr] (crot) -- node[lab, above] {$\bar{\boldsymbol V}$} (wo);
\draw[arr] (wo) -- (out);
\draw[arr] (out.east) -- ++(0.75,0);
 
% skip connection (residual stream)
\draw[arr] (-1.3,-1.3) -- (Z.west);
\fill[black!60] (-0.75,-1.3) circle (1.3pt);
\draw[skip] (-0.75,-1.3) -- (-0.75,-3.75) -- node[lab, above, pos=0.5] {$\boldsymbol Z$ (residual stream)} (out.south |- 0,-3.75) -- (out.south);
 
% ---------------- arrows: dynamical model ----------------
\draw[darr] (par) -- (E);
\draw[darr] (par) -- (P);
\draw[darr] ([xshift=3.5mm]E.south) -- node[lab, left] {$\boldsymbol E$} ([xshift=3.5mm]E.south |- res.north);
\draw[darr] (11.3,0 |- P.south) -- node[lab, right] {$\boldsymbol P_{\Delta t}$} (11.3,0 |- logit.north);
\draw[darr] (E.north) -- (E.north |- 0,3.6) -- node[lab, above, pos=0.5] {$\boldsymbol E$} (16.85,3.6) -- (A.north);
 
% ---------------- legend ----------------
\begin{scope}[shift={(4.6,-4.5)}, font=\scriptsize]
  \node[std, minimum height=0.3cm, minimum width=0.45cm, inner sep=0pt] at (0,0) {};
  \node[anchor=west] at (0.3,0) {as in RoPE attention};
  \node[rfa, minimum height=0.3cm, minimum width=0.45cm, inner sep=0pt] at (3.6,0) {};
  \node[anchor=west] at (3.9,0) {modified by RFA};
  \node[dyn, minimum height=0.3cm, minimum width=0.45cm, inner sep=0pt] at (6.7,0) {};
  \node[anchor=west] at (7.0,0) {dynamical model (depends on lag only)};
\end{scope}
\end{tikzpicture}
}
\caption{\textbf{A single Isotropic RFA head.} The dynamical model (blue)
depends only on the lag and a few per-head scalars, and supplies the decay
kernel $\boldsymbol E$ and the DLE precision $\boldsymbol P_{\Delta t}$,
with $\alpha=\eta^2-\tilde\sigma^2$ and $\beta=\gamma^2+\tilde\sigma^2$.
The robust weight $w$ may be the Student-$t$ or exponential kernel, and
the softmax is causally masked. \emph{Inset:} precision as a function of
lag in the diffusive ($\alpha<0$) and integrative ($\alpha>0$) regimes.}
  \label{fig:rfa_block}
\end{figure}

\newpage

\section{Recovery of Some Common Positional Encodings}
\label{sec:recovery_pos_encodings}

\subsection{The Zero-Decay Limit and ALiBi}
\label{sec:alibi_derivation}

If the queries and keys are normalized, the squared residual norm becomes
\[
\bigl\| \boldsymbol{R}_{qk} \bigr\|^{2}
=
\boldsymbol{1}
+
\boldsymbol{E}^{2}
-
2\,\boldsymbol{E}
\odot
\mathrm{Re}\!\left(
\tilde{\boldsymbol{Q}}^{\dagger}
\tilde{\boldsymbol{K}}
\right).
\]
For zero-decay ($\mu \rightarrow 0$), ($\boldsymbol{E}=\boldsymbol{1}$), the residual becomes the chordal distance on the unit hypersphere:
\[
\bigl\| \boldsymbol{R}_{qk} \bigr\|^{2}
=
2\Big(
\boldsymbol{1}
-
\mathrm{Re}
(
\tilde{\boldsymbol{Q}}^{\dagger}
\tilde{\boldsymbol{K}}
)
\Big).
\]
Using the exponential attention kernel gives
\[
\boldsymbol{L}
=
\log(\boldsymbol{P}_{\Delta t})
-
\frac{2}{\nu}
\boldsymbol{P}_{\Delta t}
+
\frac{2}{\nu}
\boldsymbol{P}_{\Delta t}
\odot
\mathrm{Re}
\!\left(
\tilde{\boldsymbol{Q}}^{\dagger}
\tilde{\boldsymbol{K}}
\right).
\]
In this limit, $\boldsymbol{\Sigma}_{\Delta t}$ grows linearly with temporal lag, 
\[
\boldsymbol{\Sigma}_{\Delta t}
=
\sigma^2\Delta t+\eta^2+\gamma^2,
\]
and therefore
\(
\boldsymbol{P}_{\Delta t}
=
 \frac{1}{\sigma^2\Delta t+\eta^2+\gamma^2}.
\)
The additive positional bias is therefore
\[
\boldsymbol{B}_{\Delta t}
:=
\log(\boldsymbol{P}_{\Delta t})
=
-\log\!\left(1+\frac{\sigma^2}{\eta^2+\gamma^2}\Delta t\right) + \mathrm{const},
\]
which, for $\Delta t \ll (\eta^2+\gamma^2)/\sigma^2$, admits the approximation
\(
\boldsymbol{B}_{\Delta t}
\approx
-
\frac{\sigma^2}{\eta^2+\gamma^2}\Delta t
+
\mathrm{const,}
\)
recovering the linear recency bias used by ALiBi, with slope and intercept determined by $\sigma^2$, $\eta^2$, and $\gamma^2$. At longer lags, the penalty grows only logarithmically, the form of the logarithmic variant of KERPLE \citep{chi2022kerple}, and for $\mu>0$ it
levels off entirely, approaching $-\log(\tilde\sigma^2+\gamma^2)$
(Appendix~\ref{sec:pos_heuristics}), unlike ALiBi's linear penalty.

\subsection{Diffusive and Integrative Filtering Regimes}
\label{sec:pos_heuristics}

\begin{figure}[H]
    \centering
    \includegraphics[width=0.68\linewidth]{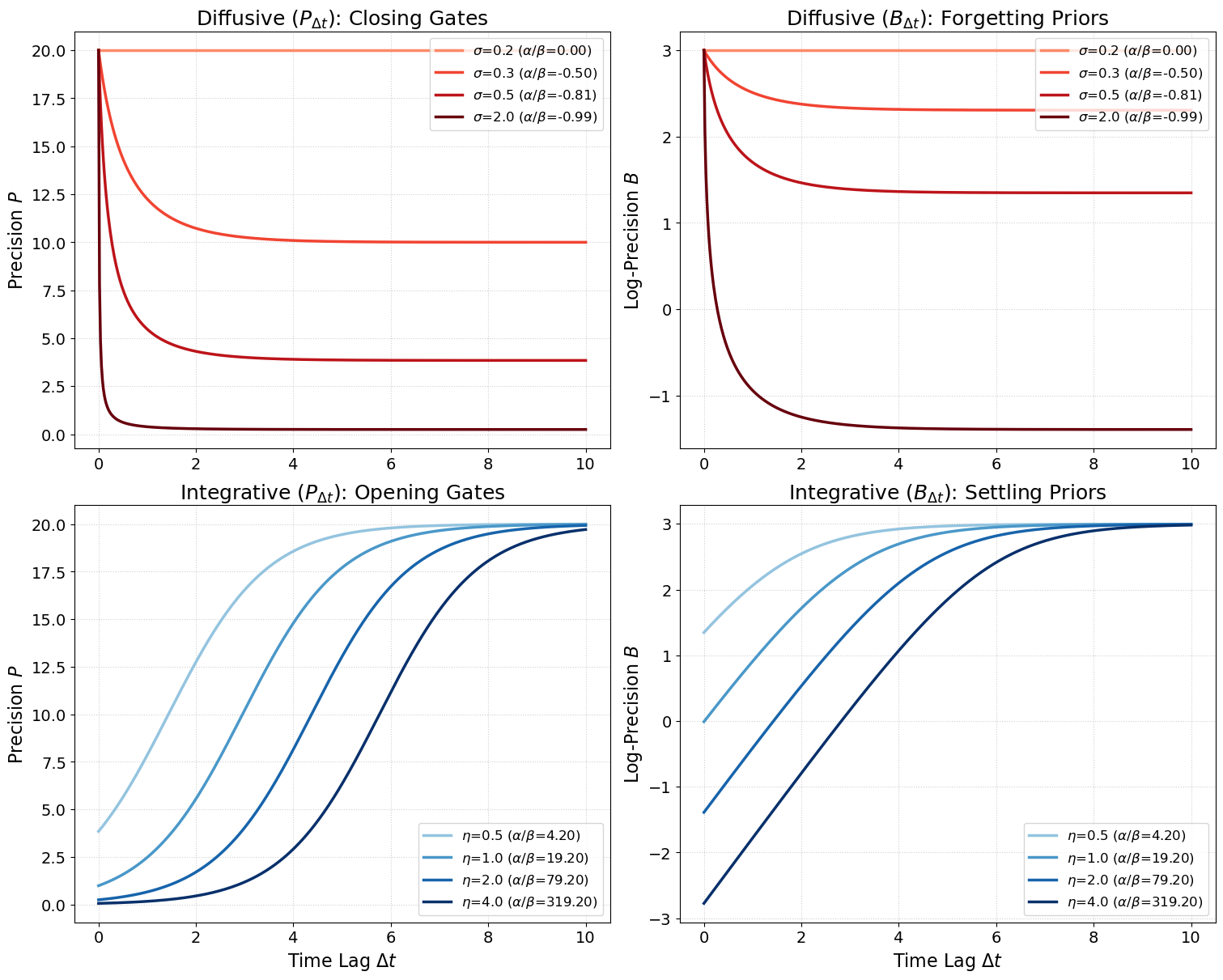}
    \caption{By varying the ratio of steady-state process uncertainty $\tilde{\sigma}^2$ to measurement noise $\eta^2$, RFA heads can specialize into distinct physical regimes: a \textbf{diffusive regime} that favors local recency (top row) and an \textbf{integrative regime} (bottom row) that filters transient noise to identify stable historical trends. The multiplicative gate $\boldsymbol{P}_{\Delta t}$ controls the selectivity (adaptive gain) of the attention, while the additive bias $\boldsymbol{B}_{\Delta t}$ defines the prior budget allocated to tokens at a given temporal lag.}
    \label{fig:diffusive_vs_integrative}
\end{figure}

\begin{figure}[H]
    \centering
    \includegraphics[width=0.68 \linewidth]{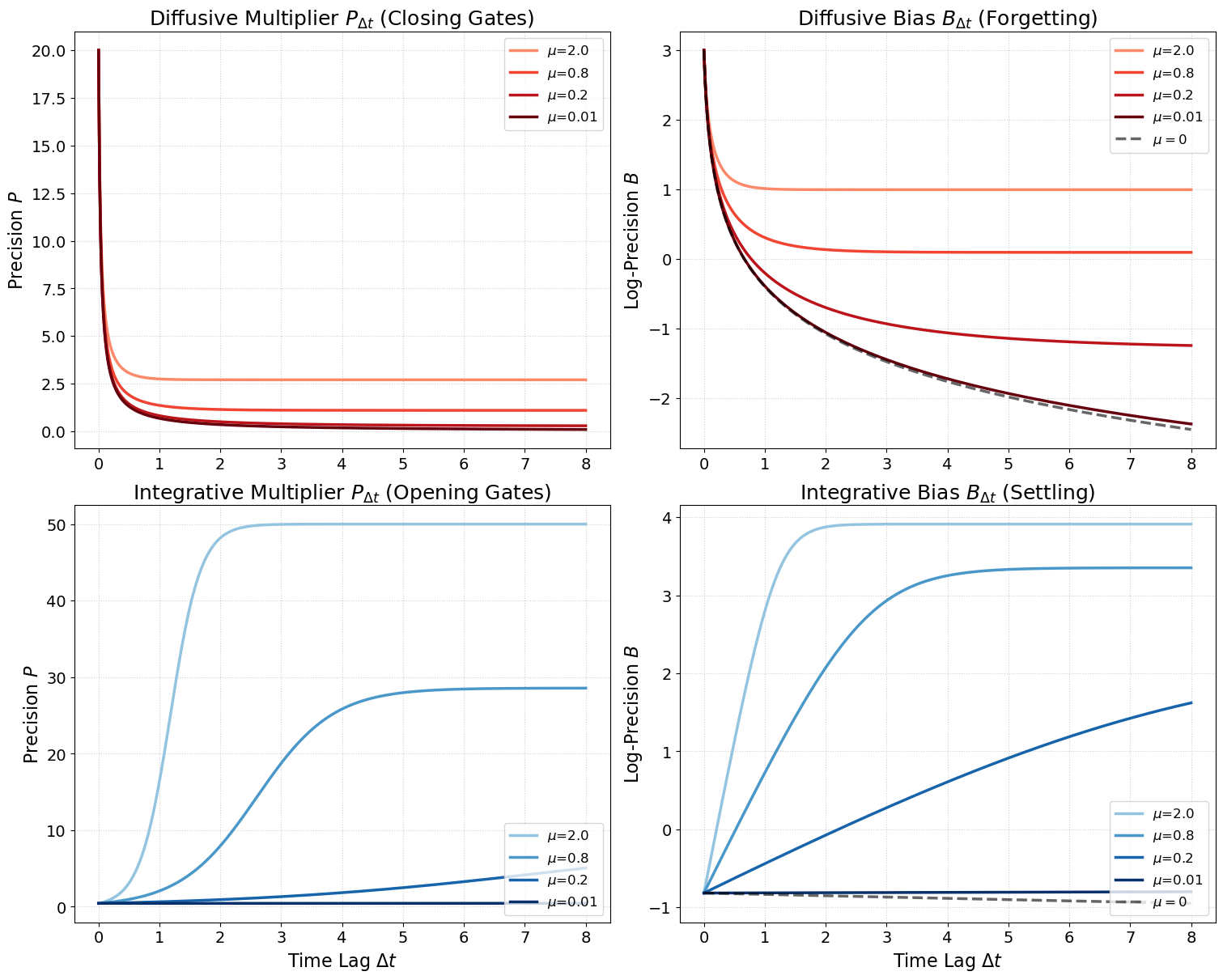}
    \caption{The decay rate $\mu$ dictates the speed of the phase transition. As $\mu \to 0$, the model recovers Brownian diffusion, where variance grows linearly with lag. As $\mu$ increases, the model enforces stationarity, where the attention bias saturates to a learned global noise floor $\beta$, providing a principled mechanism for long-range context retention.}
    \label{fig:bias_and_gate_vs_mu}
\end{figure}

Here, we derive the explicit functional forms of the additive bias 
$\boldsymbol{B}_{\Delta t} := \log(\boldsymbol{P}_{\Delta t})$ and multiplicative gate 
$\boldsymbol{P}_{\Delta t}$ in each regime, determined by the values of $\alpha := \eta^2 - \tilde{\sigma}^2$ and $\beta := \gamma^2 + \tilde{\sigma}^2$. The behavior of each regime is illustrated in Fig.~\ref{fig:diffusive_vs_integrative}, and the effect of $\mu$ on the speed 
of the phase transition is shown in Fig.~\ref{fig:bias_and_gate_vs_mu}.

\paragraph{Diffusive Regime ($\alpha < 0$).}
Letting $\alpha' = -\alpha > 0$, the bias follows a logarithmic decay:
\[
\boldsymbol{B}_{\Delta t}
=
-\log(\beta)
-
\log\!\left(1 - \frac{\alpha'}{\beta} e^{-2\mu \Delta t}\right),
\]
starting at its maximum at $\Delta t = 0$ and decaying toward $-\log(\beta)$. The precision decays as:
\[
\boldsymbol{P}_{\Delta t} = \bigl(\beta - \alpha' e^{-2\mu \Delta t}\bigr)^{-1},
\]
with selectivity maximal near the diagonal and blurring out at longer lags.

\paragraph{Integrative Regime ($\alpha > 0$).}
The bias follows a mirrored softplus:
\[
\boldsymbol{B}_{\Delta t}
=
-\log(\beta)
-
\mathrm{softplus}\!\left(\ln(\alpha/\beta) - 2\mu \Delta t\right),
\]
starting low and curving upward as key-side measurement noise dissipates. The precision  follows a sigmoid:
\[
\boldsymbol{P}_{\Delta t}
=
\frac{1}{\beta}
\cdot
\mathrm{sigmoid}\!\left(2\mu \Delta t - \ln(\alpha/\beta)\right),
\]
with selectivity initially low, opening as the transported observation becomes reliable.

% \newpage

\subsection{Spectrally Coupled RFA}
\label{sec:SC_RFA_appendix}

\begin{figure}[H]
    \centering
    \includegraphics[width=0.48\linewidth]{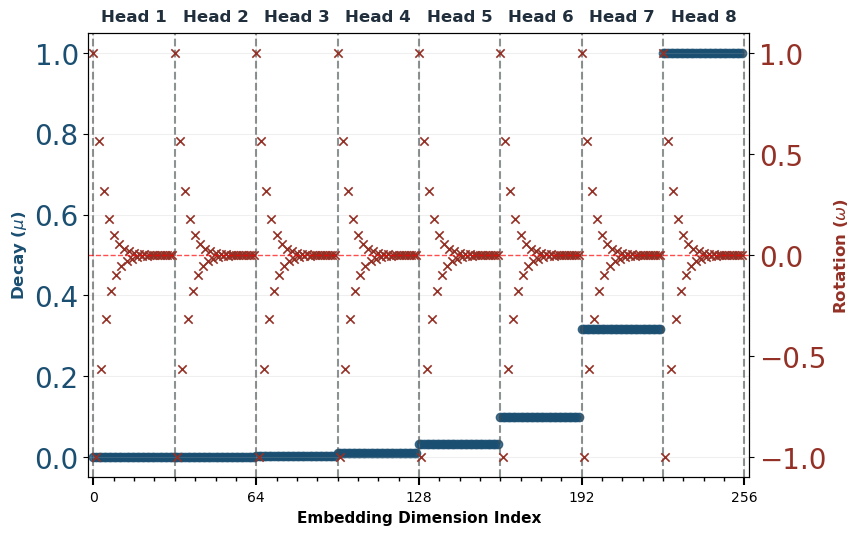}\hfill
    \includegraphics[width=0.48\linewidth]{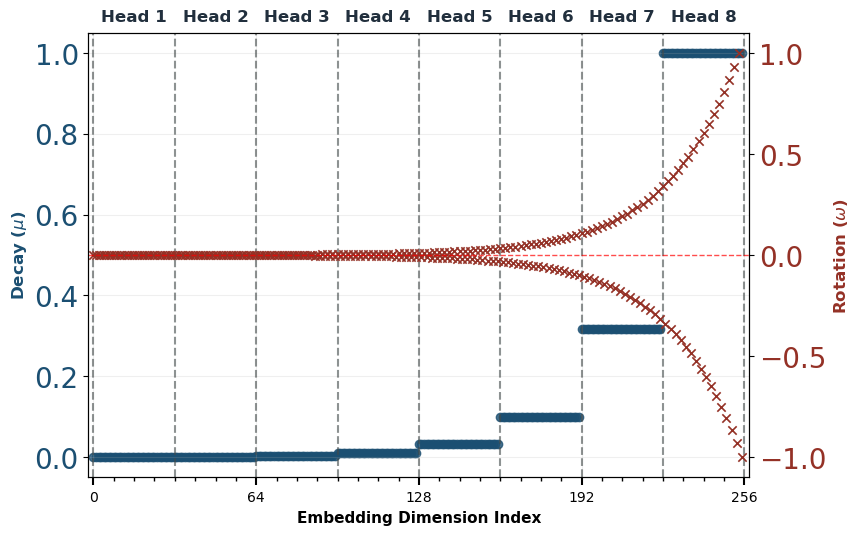}
    \caption{\textbf{Eigenvalue spectra of standard isotropic RFA (left) and
    SC-RFA (right), with $b=1.0$.} Standard RFA uses the full frequency range with
    uniform decay per head. SC-RFA assigns each head a distinct spectral band with
    decay rate coupled to its maximum frequency. Note that eigenvalues appear in complex-conjugate pairs since the dynamics are real-valued.}
    \label{fig:eigenvals_rope_vs_sc_rope}
\end{figure}

\begin{figure}[H]
    \centering
    \includegraphics[width=0.9\linewidth]{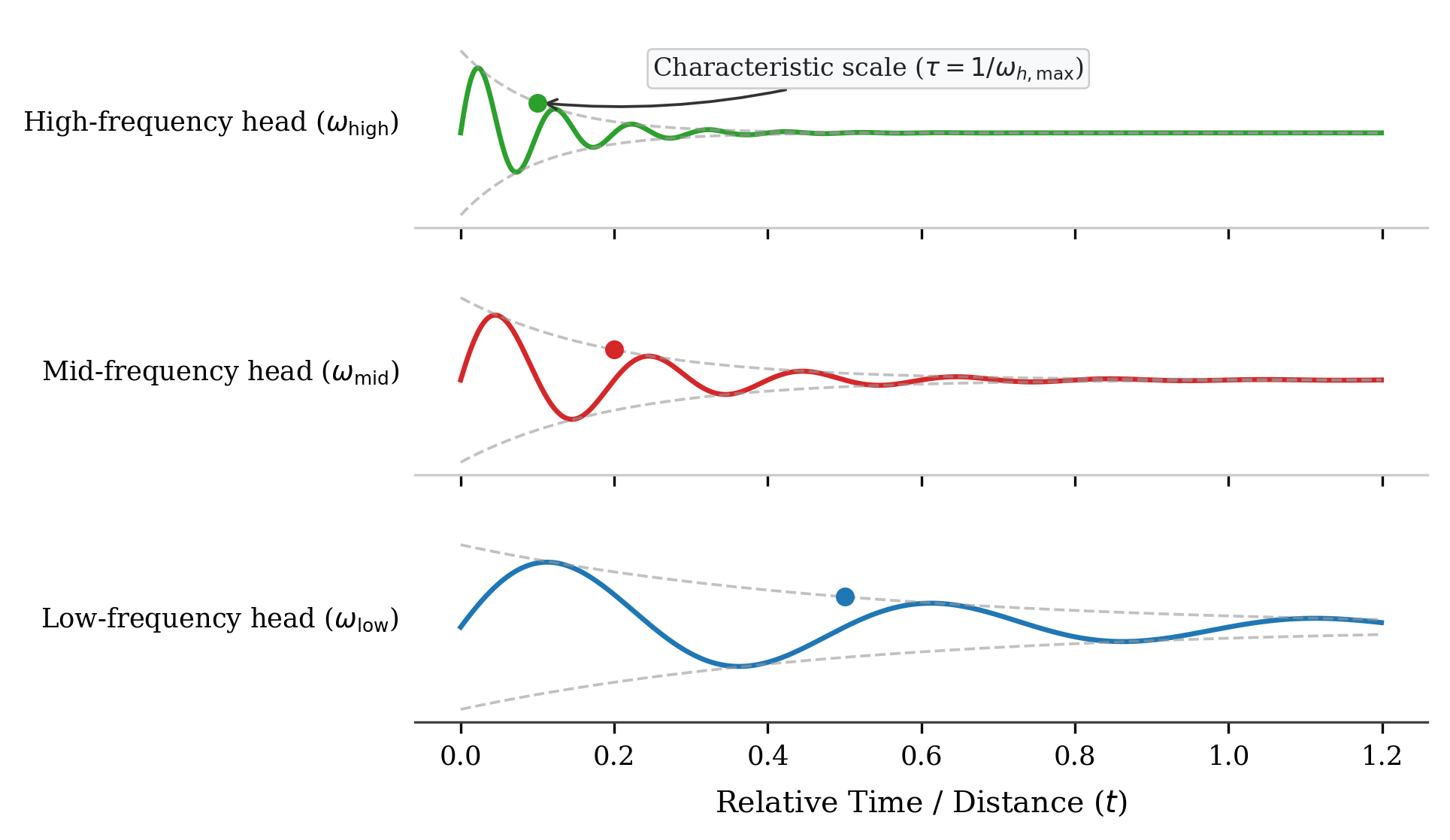}
    \caption{\textbf{Temporal scale separation in SC-RFA.} Coupling each head's decay rate to its maximum frequency ($\mu_h=b\,\omega_{h,\max}$) causes high-frequency heads to attenuate rapidly while low-frequency heads remain active over longer temporal horizons.}
    \label{fig:sc_rfa_scale_separation}
\end{figure}

In standard RFA, all heads share the full frequency range with a uniform decay rate,
so high- and low-frequency components are damped equally within each head
(Fig.~\ref{fig:eigenvals_rope_vs_sc_rope}, left). SC-RFA partitions the spectrum
across heads and couples each head's decay rate to its maximum frequency
(Fig.~\ref{fig:eigenvals_rope_vs_sc_rope}, right). As illustrated in
Fig.~\ref{fig:sc_rfa_scale_separation}, this produces an ordered separation of
temporal scales: high-frequency heads decay rapidly and act as short-range filters,
while low-frequency heads decay more slowly and preserve long-range structure.

% \begin{figure}[H]
%     \centering
%     \includegraphics[width=0.45\linewidth]{images/explanations/eigenvals_sc_rfa_loglog.png}
%     \caption{\textbf{SC-RFA eigenvalue distribution in log-log space ($b=1.0$).}
%     The boundary $\mu_h = b \cdot \omega_{h,\mathrm{max}}$ appears as a straight line of
%     slope $b$. Each head's eigenvalues form a vertical strip, with the
%     highest-frequency eigenvalue on the boundary and lower-frequency eigenvalues
%     falling below it.}
%     \label{fig:sc_rfa_loglog}
% \end{figure}

\newpage

\section{Real-Valued Realization and Initialization}
\label{sec:Implementation}

\subsection{Complex-valued Computations}
\label{sec:Complex-valued Computations}

RFA is formulated in a complex latent space. In practice, this is implemented by lifting real-valued representations into a $2d$-dimensional space using a linear projection (corresponding to $\mathbb{C}^d \cong \mathbb{R}^{2d}$), performing complex rotations and attention in this space, and then projecting the result back to $\mathbb{R}^d$.
% Conceptually, this corresponds to applying attention in a $2d$-dimensional space with structured rotations and inner products that implement complex arithmetic.

A complex-valued linear transformation can be represented in the real domain by operating on paired real and imaginary channels. For an input $\boldsymbol{x} = \left[ \boldsymbol{x}_r, \boldsymbol{x}_i \right]^\top$ with $\boldsymbol{x}_r, \boldsymbol{x}_i \in \mathbb{R}^d$, this corresponds to:
\[
\mathcal{L}(\boldsymbol{x})
=
\begin{bmatrix}
\boldsymbol{W}_r & -\boldsymbol{W}_i \\[4pt]
\boldsymbol{W}_i & \boldsymbol{W}_r
\end{bmatrix}
\boldsymbol{x}
+
\begin{bmatrix}
\boldsymbol{b}_r \\[4pt]
\boldsymbol{b}_i
\end{bmatrix},
\]
Here $\boldsymbol{W}_r, \boldsymbol{W}_i \in \mathbb{R}^{d \times d}$ are the real and imaginary components of the weight matrix and $\boldsymbol{b}_r, \boldsymbol{b}_i \in \mathbb{R}^{d}$ the bias. This is equivalent to multiplication by a complex matrix $\boldsymbol{W} = \boldsymbol{W}_r + i\boldsymbol{W}_i$ with bias $\boldsymbol{b} = \boldsymbol{b}_r + i\boldsymbol{b}_i$.

Assuming the inputs and outputs are purely real, only the real-input columns of the input projections and the real-output columns of the output projections are required:
\[
\mathcal{L}^{d \rightarrow 2d}(\boldsymbol{x}_r)
:=
\begin{bmatrix}
\boldsymbol{W}_r \\[4pt]
\boldsymbol{W}_i
\end{bmatrix}
\boldsymbol{x}_r
+
\begin{bmatrix}
\boldsymbol{b}_r \\[4pt]
\boldsymbol{b}_i
\end{bmatrix}.
\]
\[
\mathcal{L}^{2d \rightarrow d}(\boldsymbol{x})
:=
\begin{bmatrix}
\boldsymbol{W}_r & -\boldsymbol{W}_i
\end{bmatrix}
\boldsymbol{x}
+
\boldsymbol{b}_r.
\]
Hence, both projections may be implemented using standard real-valued linear layers in $\mathbb{R}^{2d}$.

With queries, keys, and values
$\boldsymbol{Q},\boldsymbol{K},\boldsymbol{V}=\mathcal{L}_{q,k,v}^{d\rightarrow 2d}(\boldsymbol{Z})$,
and $\boldsymbol{C}_\omega[k,i]=\cos(\omega_k t_i)$,
$\boldsymbol{S}_\omega[k,i]=\sin(\omega_k t_i)$, complex rotations are applied as:
\[
\tilde{\boldsymbol{Q}}^\top = \begin{bmatrix}
\tilde{\boldsymbol{Q}}^\top_{r} \\
\tilde{\boldsymbol{Q}}^\top_{i}
\end{bmatrix} = (\tilde{\boldsymbol{\Phi}}^{-})^\top \odot \boldsymbol{Q}^\top = \begin{bmatrix}
\boldsymbol{C}_\omega \odot \boldsymbol{Q}^\top_{r} + \boldsymbol{S}_\omega \odot\boldsymbol{Q}^\top_{i} \\
\boldsymbol{C}_\omega \odot \boldsymbol{Q}^\top_{i} - \boldsymbol{S}_\omega \odot\boldsymbol{Q}^\top_{r}
\end{bmatrix},
\]
and likewise for $\tilde{\boldsymbol{K}}$ and $\tilde{\boldsymbol{V}}$. This is algebraically identical to RoPE.

To ensure the underlying system matrix $\boldsymbol{A}$ is real-valued, we enforce that its eigenvalues appear in complex conjugate pairs:
\[
\boldsymbol{\omega} = \big\{ \omega_1, - \omega_1, \dots, \omega_{d/2}, -\omega_{d/2} \big\}, 
\]
Hence, only the positive-frequency half of the spectrum is learned explicitly, with the remaining frequencies determined by conjugation. The lifted $2d$-dimensional representation requires only $d/2$ learned frequencies, rather than the $d$ frequencies that would be required by an unconstrained RoPE parameterization in the same space.

The Mahalanobis distance requires the real part of the complex inner product,
\[
\mathrm{Re}\!\left(
\tilde{\boldsymbol{Q}}^{\dagger}
\tilde{\boldsymbol{K}}
\right)
=
\tilde{\boldsymbol{Q}}_{r}^{\top}\tilde{\boldsymbol{K}}_{r}
+
\tilde{\boldsymbol{Q}}_{i}^{\top}\tilde{\boldsymbol{K}}_{i}
= \Big[ \tilde{\boldsymbol{Q}}_r^\top \, \, \tilde{\boldsymbol{Q}}_i^\top \Big]
\begin{bmatrix} \tilde{\boldsymbol{K}}_r \\ \tilde{\boldsymbol{K}}_i \end{bmatrix} .
\]
This is implemented as a single real matrix multiplication in $\mathbb{R}^{2d}$.

Value aggregation, $\bar{\boldsymbol{V}}$, is computed in the $\mathbb{R}^{2d}$ domain. The real-valued attention matrix $\hat{\boldsymbol{A}}$ is applied identically to both the real and imaginary components of the complex-rotated values:
\[
\boldsymbol{M}^\top
= \hat{\boldsymbol{A}}
\begin{bmatrix}
\tilde{\boldsymbol{V}}_{r}^\top \\
\tilde{\boldsymbol{V}}_{i}^\top
\end{bmatrix}.
\]
The inverse rotation yields:
\[
\bar{\boldsymbol{V}}^\top 
= (\tilde{\boldsymbol{\Phi}}^{+})^\top \odot ( \hat{\boldsymbol{A}} \tilde{\boldsymbol{V}}^\top )
= \begin{bmatrix}
\boldsymbol{C}_\omega \odot \boldsymbol{M}^\top_{ r} - \boldsymbol{S}_\omega \odot\boldsymbol{M}^\top_{i} \\
\boldsymbol{C}_\omega \odot \boldsymbol{M}^\top_{ i} + \boldsymbol{S}_\omega \odot\boldsymbol{M}^\top_{r}
\end{bmatrix}
\]
The estimate $\bar{\boldsymbol{V}}$ is projected back to the real domain using
the $\mathcal{L}^{2d \rightarrow d}$ layer,
$\boldsymbol Z^+= \boldsymbol Z + \mathcal L_o^{2d\to d}(\bar{\boldsymbol V})$.
All components of RFA are therefore implemented using standard real-valued operations.

\subsection{Initialization}
\label{sec:Initializations}

% RFA initializes complex projections and dynamics to cover multiple temporal scales and ensure numerical stability.

\paragraph{Isotropic Complex Projections.}  
Complex weights $\boldsymbol{W} = \boldsymbol{W}_r + i\boldsymbol{W}_i$ are initialized isotropically:
\[
\boldsymbol{W}_{ij} = M_{ij} \begin{bmatrix} \cos(\phi_{ij}) \\ \sin(\phi_{ij}) \end{bmatrix}, \quad
M_{ij} \sim \text{Rayleigh}\Bigl(\sqrt{\frac{1}{d_\text{in}+d_\text{out}}}\Bigr), \quad
\phi_{ij} \sim \mathcal{U}(0,2\pi).
\]
Output projections ($\boldsymbol{W}_o$) are scaled by $1/\sqrt{2}$ to preserve variance when converting back to real space.

\paragraph{Noise and robustness.}
We scale the process noise with the decay rate, $\sigma^2 = 0.1\,\mu$, so
that the steady-state variance $\tilde\sigma^2 = \sigma^2/(2\mu)$ is the
same across heads despite their different temporal persistence. We
initialize the measurement noise with $\eta^2 > \tilde\sigma^2$, so that
the model starts in the integrative regime, preserving long-range gradient
flow early in training. The noise floor $\gamma^2 > 0$ keeps the precision
finite. The Student-$t$ degrees of freedom are set as $\nu = \nu_s d$ with
$\nu_s = 4$ at initialization, a quasi-Gaussian regime that avoids
prematurely rejecting tokens while the query and key representations are
still untrained.

Our experiments learn $\sigma^2$ directly. We recommend instead learning
the steady-state variance $\tilde\sigma^2$, as in Algorithm~\ref{alg:Isotropic_RFA}. The two are equivalent, but learning $\tilde\sigma^2$ decouples it from $\mu$ and is better conditioned when decay rates vary across heads.

Our implementation is written in high-level PyTorch. At $L=512$, it trains about 20\% slower per epoch than our dot-product baselines, which are likewise implemented without fused attention kernels; the overhead grows with context length. RFA's additional terms are elementwise functions of the scores and the lag, so they are compatible with fused (FlashAttention-style) kernels, which we leave to future work.

% \newpage

\subsection{Algorithm}
\label{sec:Algorithm}

Algorithm~\ref{alg:Isotropic_RFA} details the implementation of Isotropic RFA.

\begin{algorithm}[H]
\caption{Robust Filter Attention (isotropic, single head)}
\label{alg:Isotropic_RFA}
\begin{algorithmic}[1]
\REQUIRE Sequence $\boldsymbol Z\in\mathbb R^{d\times N}$ at positions $t_1,\dots,t_N$
\REQUIRE Linear layers $\mathcal L_q,\mathcal L_k,\mathcal L_v$ ($d\to 2d$) and $\mathcal L_o$ ($2d\to d$)
\REQUIRE Learned scalars $\sigma',\eta',\gamma',\nu_s,\beta_s$; decay rate $\mu$; frequencies $\omega'_1,\dots,\omega'_{d/2}$
\ENSURE Updated sequence $\boldsymbol Z^{+}\in\mathbb R^{d\times N}$
\STATE $\boldsymbol\omega \gets (\omega'_1,-\omega'_1,\dots,\omega'_{d/2},-\omega'_{d/2})$ \COMMENT{conjugate symmetry}
\STATE $(\tilde\sigma^2,\eta^2,\gamma^2) \gets \mathrm{softplus}(\sigma',\eta',\gamma')$ \COMMENT{positive variances}
\STATE $\boldsymbol Q,\ \boldsymbol K,\ \boldsymbol V \gets \mathcal L_q(\boldsymbol Z),\ \mathcal L_k(\boldsymbol Z),\ \mathcal L_v(\boldsymbol Z)$ \COMMENT{project}
\STATE $\boldsymbol E[i,j] \gets e^{-\mu|t_i-t_j|},\quad \tilde{\boldsymbol\Phi}^{\pm}[k,i] \gets e^{\pm i\omega_k t_i}$ \COMMENT{decay and rotation}
\STATE $\boldsymbol P_{\Delta t} \gets \big(\tilde\sigma^2(1-\boldsymbol E^{2})+\eta^2\boldsymbol E^{2}+\gamma^2\big)^{-1}$ \COMMENT{DLE precision}
\STATE $\tilde{\boldsymbol Q},\ \tilde{\boldsymbol K},\ \tilde{\boldsymbol V} \gets \tilde{\boldsymbol\Phi}^{-}\odot\boldsymbol Q,\ \tilde{\boldsymbol\Phi}^{-}\odot\boldsymbol K,\ \tilde{\boldsymbol\Phi}^{-}\odot\boldsymbol V$ \COMMENT{rotate}
\STATE $\|\boldsymbol R_{qk}[i,j]\|^2 \gets \|\boldsymbol Q_i\|^2+\boldsymbol E[i,j]^2\,\|\boldsymbol K_j\|^2-2\,\boldsymbol E[i,j]\,\mathrm{Re}\big(\tilde{\boldsymbol Q}_i^{\dagger}\tilde{\boldsymbol K}_j\big)$ \COMMENT{residual}
\STATE $\boldsymbol L \gets \log\boldsymbol P_{\Delta t}-(\nu_s+1)\log\!\big(1+\tfrac{1}{\nu_s d}\,\boldsymbol P_{\Delta t}\odot\|\boldsymbol R_{qk}\|^2\big)$ \COMMENT{robust logits}
\STATE $\boldsymbol A \gets \mathrm{softmax}_j\big(\beta_s\boldsymbol L+\boldsymbol M_{\text{causal}}\big),\quad \hat{\boldsymbol A} \gets \boldsymbol A\odot\boldsymbol E$ \COMMENT{normalize, decay}
\STATE $\bar{\boldsymbol V} \gets \tilde{\boldsymbol\Phi}^{+}\odot\big(\tilde{\boldsymbol V}\hat{\boldsymbol A}^{\top}\big)$ \COMMENT{aggregate, counter-rotate}
\STATE $\boldsymbol Z^{+} \gets \boldsymbol Z+\mathcal L_o(\bar{\boldsymbol V})$ \COMMENT{residual update}
\RETURN $\boldsymbol Z^{+}$
\end{algorithmic}
\end{algorithm}

\newpage

\section{Extensions}
\label{sec:Extensions}

This section presents some natural generalizations of the RFA framework.

\subsection{Distinct Observation Models for Query and Key Streams}
\label{sec:Observation_Models}

The main text assumes a shared observation model $\boldsymbol{z}_i = \boldsymbol{C} \boldsymbol{x}(t_i) + \boldsymbol{\varepsilon}_i$. More generally, the query and key streams may arise from distinct observation models:
\[
\boldsymbol{z}_{i} = \boldsymbol{C}_q \boldsymbol{x}(t_i) + \boldsymbol{\varepsilon}_{q,i}, \quad
\boldsymbol{z}_{i} = \boldsymbol{C}_k \boldsymbol{x}(t_i) + \boldsymbol{\varepsilon}_{k,i},
\]
where $\boldsymbol{\varepsilon}_{q,i} \sim \mathcal{N}(\boldsymbol{0}, \boldsymbol{R}_q)$, 
$\boldsymbol{\varepsilon}_{k,i} \sim \mathcal{N}(\boldsymbol{0}, \boldsymbol{R}_k)$, and 
$\boldsymbol{C}_q, \boldsymbol{C}_k \in \mathbb{R}^{d \times d}$ are invertible observation operators. Under this interpretation, the query and key streams represent distinct noisy observations of a common latent state. The learned projections $\boldsymbol{W}_q$ and $\boldsymbol{W}_k$ can be interpreted as mapping these
observations into a shared latent coordinate system, approximately inverting their respective observation operators. This provides a probabilistic interpretation for asymmetric query and key projections. The latent state $\boldsymbol{x}(t)$ evolves under the same dynamics as in Section~\ref{sec:Setup}. 

Defining:
\[
\boldsymbol{q}_i := \boldsymbol{W}_q \boldsymbol{z}_{i}, \quad
\boldsymbol{k}_j := \boldsymbol{W}_k \boldsymbol{z}_{j}, \quad
\boldsymbol{W}_q := \boldsymbol{S}^{-1}\boldsymbol{C}_q^{-1}, \quad
\boldsymbol{W}_k := \boldsymbol{S}^{-1}\boldsymbol{C}_k^{-1},
\]
the observable residual in the eigenbasis takes the same form as in the main text:
\[
\boldsymbol{r}_{ij} = \boldsymbol{q}_i - e^{\boldsymbol{\Lambda}\Delta t_{ij}} \boldsymbol{k}_j.
\]
The estimator yields the same precision-weighted form as Eq.~\ref{eq:precision_weighted_average}. Since $\boldsymbol{W}_k$ maps keys into the latent eigenbasis, the same transformation applies to values, which represent the same latent state estimates being aggregated. Hence, the model implies $\boldsymbol{W}_v = \boldsymbol{W}_k$, up to diagonal rescaling.

The residual covariance in the eigenbasis is:
\[
\boldsymbol{\Sigma}_{ij}
=
\boldsymbol{R}_{q,s}
+
e^{\boldsymbol{\Lambda}\Delta t_{ij}}
\boldsymbol{R}_{k,s}
e^{\boldsymbol{\Lambda}^{\dagger}\Delta t_{ij}}
+
\boldsymbol{V}_{k,s}(\Delta t_{ij}),
\]
\[
\boldsymbol{R}_{q,s}
=
\boldsymbol{S}^{-1}
\boldsymbol{C}_q^{-1}
\boldsymbol{R}_q
\boldsymbol{C}_q^{-\dagger}
\boldsymbol{S}^{-\dagger}, \quad
\boldsymbol{R}_{k,s}
=
\boldsymbol{S}^{-1}
\boldsymbol{C}_k^{-1}
\boldsymbol{R}_k
\boldsymbol{C}_k^{-\dagger}
\boldsymbol{S}^{-\dagger},
\quad
\boldsymbol{V}_{k,s}(\Delta t)
=
\boldsymbol{S}^{-1}
\boldsymbol{V}(\Delta t)
\boldsymbol{S}^{-\dagger}.
\]
For isotropic measurement noise,
\(
\boldsymbol{R}_{q,s}
=
\gamma^2 \boldsymbol{I}
\)
and
\(
\boldsymbol{R}_{k,s}
=
\eta^2 \boldsymbol{I},
\)
\(
\boldsymbol{\Sigma}_{ij}
\)
remains diagonal in the eigenbasis and efficient computation of the Mahalanobis distance is preserved.

\subsection{Constant Drift and Measurement Bias}
\label{sec:Inhomogeneous}

Here, we show that constant drift and measurement bias induce only deterministic shifts in the propagated mean, which can be absorbed into learned bias terms in the query, key, and value projections.

Consider an inhomogeneous linear SDE with constant drift $\boldsymbol{u}$ and constant measurement bias $\boldsymbol{\beta}$:
\[
d\boldsymbol{x}(t) = \big( \boldsymbol{A} \boldsymbol{x}(t) + \boldsymbol{u} \big)\, dt
+ \boldsymbol{G}\, d\boldsymbol{W}(t), 
\quad
\boldsymbol{z}_k = \boldsymbol{C} \boldsymbol{x}(t_k)
+ \boldsymbol{\beta}
+ \boldsymbol{\varepsilon}(t_k).
\]
The drift term can be eliminated via state augmentation, but this introduces a singular system that breaks simultaneous diagonalization. We therefore treat the inhomogeneous case directly.

We assume $\boldsymbol{A}$ is Hurwitz. Identifying the equilibrium point $\boldsymbol{\mu}$, the SDE can be rewritten as
\[
d\boldsymbol{x}(t)
=
\boldsymbol{A}\big(\boldsymbol{x}(t)-\boldsymbol{\mu}\big)\,dt
+
\boldsymbol{G}\,d\boldsymbol{W}(t),
\qquad
\boldsymbol{\mu}
=
-\boldsymbol{A}^{-1}\boldsymbol{u}.
\]
The solution to the SDE, propagating the state forward from $\boldsymbol{x}(t_j)$ to $\boldsymbol{x}(t_i)$, is
\[
\boldsymbol{x}(t_i)
=
e^{\boldsymbol{A}\Delta t_{ij}}\boldsymbol{x}(t_j)
+
\bigg(
\int_0^{\Delta t_{ij}}
e^{\boldsymbol{A}(\Delta t_{ij}-\tau)}
\,d\tau
\bigg)
\boldsymbol{u}
+
\int_0^{\Delta t_{ij}}
e^{\boldsymbol{A}(\Delta t_{ij}-\tau)}
\boldsymbol{G}\,
d\boldsymbol{W}(\tau).
\]
Letting
\(
\boldsymbol{G}_{u}(\Delta t_{ij})
=
\int_0^{\Delta t_{ij}}
e^{-\boldsymbol{A}\tau}\,d\tau,
\)
the deterministic part is
\(
\hat{\boldsymbol{x}}_{ij}
=
e^{\boldsymbol{A}\Delta t_{ij}}
\Big(
\boldsymbol{x}(t_j)
+
\boldsymbol{G}_{u}(\Delta t_{ij})\boldsymbol{u}
\Big).
\)
Letting $\boldsymbol{u}_s:=\boldsymbol{S}^{-1}\boldsymbol{u}$, the drift term is
\[
\boldsymbol{G}_{u}(\Delta t_{ij})\boldsymbol{u}
=
\boldsymbol{S}
\bigg(
\int_0^{\Delta t_{ij}}
e^{-\boldsymbol{\Lambda}\tau}\,d\tau
\bigg)
\boldsymbol{u}_s
=
\boldsymbol{S}
\bigg(
\frac{\boldsymbol{I}
-
e^{-\boldsymbol{\Lambda}\Delta t_{ij}}}
{\boldsymbol{\Lambda}}
\bigg)
\boldsymbol{u}_s.
\]
Hence, the propagated state in the eigenbasis becomes
\[
\hat{\boldsymbol{x}}_{s,ij}
=
e^{\boldsymbol{\Lambda}\Delta t_{ij}}
\bigg(
\hat{\boldsymbol{x}}_{s,j}
+
\frac{\boldsymbol{I}
-
e^{-\boldsymbol{\Lambda}\Delta t_{ij}}}
{\boldsymbol{\Lambda}}
\boldsymbol{u}_s
\bigg)
=
e^{\boldsymbol{\Lambda}\Delta t_{ij}}
\left(
\hat{\boldsymbol{x}}_{s,j}
+
\frac{\boldsymbol{u}_s}
{\boldsymbol{\Lambda}}
\right)
-
\frac{\boldsymbol{u}_s}
{\boldsymbol{\Lambda}},
\]
where division is element-wise. Since the latent state is recovered from the measurements via
\(
\hat{\boldsymbol{x}}_j
=
\boldsymbol{C}^{-1}
(\boldsymbol{z}_j-\boldsymbol{\beta}),
\)
the diagonalized state estimate is
\[
\hat{\boldsymbol{x}}_{s,j}
=
\boldsymbol{y}_{s,j}
-
\boldsymbol{\beta}_s,
\qquad
\boldsymbol{y}_{s,j}
:=
\boldsymbol{S}^{-1}\boldsymbol{C}^{-1}\boldsymbol{z}_j,
\qquad
\boldsymbol{\beta}_s
:=
\boldsymbol{S}^{-1}\boldsymbol{C}^{-1}\boldsymbol{\beta}.
\]
Substituting this expression into the propagated state yields
\[
\hat{\boldsymbol{x}}_{s,ij}
=
e^{\boldsymbol{\Lambda}\Delta t_{ij}}
\left(
\boldsymbol{y}_{s,j}
+
\frac{\boldsymbol{u}_s}{\boldsymbol{\Lambda}}
-
\boldsymbol{\beta}_s
\right)
-
\frac{\boldsymbol{u}_s}{\boldsymbol{\Lambda}},
\]
so the residual against the current estimate
$\hat{\boldsymbol{x}}_{s,i}=\boldsymbol{y}_{s,i}-\boldsymbol{\beta}_s$ is
\[
\boldsymbol{r}_{s,ij}
=
\left(
\boldsymbol{y}_{s,i}
+
\boldsymbol{b}_s
\right)
-
e^{\boldsymbol{\Lambda}\Delta t_{ij}}
\left(
\boldsymbol{y}_{s,j}
+
\boldsymbol{b}_s
\right),
\qquad
\boldsymbol{b}_s
:=
\frac{\boldsymbol{u}_s}{\boldsymbol{\Lambda}}
-
\boldsymbol{\beta}_s .
\]
Thus, constant drift and measurement bias combine into a single constant offset in the diagonalized coordinates, applied identically to the query and the key.

Since this offset is constant across time, it can be absorbed into the learned linear projections by defining bias terms
$\boldsymbol{b}_q,\boldsymbol{b}_k,\boldsymbol{b}_v\in\mathbb{C}^{d\times1}$
in the input projections defining the queries, keys, and values:
\[
\boldsymbol{Q}_{u}[k,i]
:=
\boldsymbol{Q}[k,i]
+
\boldsymbol{b}_q[k],
\qquad
\boldsymbol{K}_{u}[k,i]
:=
\boldsymbol{K}[k,i]
+
\boldsymbol{b}_k[k],
\qquad
\boldsymbol{V}_{u}[k,i]
:=
\boldsymbol{V}[k,i]
+
\boldsymbol{b}_v[k],
\]
\[
\boldsymbol{b}_{\ell}[k]
:=
\frac{\boldsymbol{u}_{\ell}[k]}
{\boldsymbol{\lambda}_{\ell,k}}
-
\boldsymbol{\beta}_{\ell}[k],
\qquad
\ell\in\{q,k,v\},
\]
where $\boldsymbol{u}_{\ell}[k]$, $\boldsymbol{\beta}_{\ell}[k]$, and $\boldsymbol{\lambda}_{\ell,k}$ denote the $k$th diagonal element of the projected drift, projected measurement bias, and eigenvalue vector associated with projection $\ell$, respectively. These bias terms correspond to the steady-state offset induced by the deterministic dynamics. This allows the residual tensor to maintain the same form as the homogeneous case using the biased tensors:
\[
\mathcal{R}_{qk}[k,i,j]
=
\boldsymbol{Q}_{u}[k,i]
-
\mathcal{E}_{qk}[k,i,j]
\cdot
\boldsymbol{K}_{u}[k,j].
\]
The attention output is
\[
\bar{\boldsymbol{V}}[k,i]
=
\boldsymbol{\Phi}_{v}[k,i]
\cdot
\sum_{j\le i}
\mathcal{A}[k,i,j]
\cdot
\hat{\boldsymbol{V}}_{u}[k,j]
-
\boldsymbol{b}_v[k]
\cdot
\sum_{j\le i}
\mathcal{A}[k,i,j],
\]
which simplifies, using
$\sum_{j\le i}\mathcal{A}[k,i,j]=1$, to
\[
\bar{\boldsymbol{V}}[k,i]
=
\boldsymbol{\Phi}_{v}[k,i]
\cdot
\sum_{j\le i}
\mathcal{A}[k,i,j]
\cdot
\hat{\boldsymbol{V}}_{u}[k,j]
-
\boldsymbol{b}_v[k].
\]
The bias in this final expression can be absorbed into the bias of the output projection,
\(
\boldsymbol{b}_{o}
:=
\boldsymbol{W}_{o}\boldsymbol{b}_{v}.
\)
Hence, the inhomogeneous SDE with constant drift $\boldsymbol{u}$ and constant measurement bias $\boldsymbol{\beta}$ is structurally equivalent to the homogeneous RFA mechanism, provided these deterministic effects are absorbed into constant bias vectors in the input and output projections
$(\boldsymbol{b}_q,\boldsymbol{b}_k,\boldsymbol{b}_v,\boldsymbol{b}_o)$.

\subsection{Content-Dependent Dynamics and Uncertainty Propagation}
\label{sec:content_dependent_dynamics}

The dynamics and uncertainty propagation can both be made dependent on the
content of the tokens encountered along the transport path. Let the
eigenvalues at token \(\ell\) be
\(
\lambda_{k,\ell}
=
-\mu_\ell
+
i\omega_{k,\ell},
\)
where \(\mu_\ell := \mu(\boldsymbol{x}_\ell)\) and
\(\omega_{k,\ell} := \omega_k(\boldsymbol{x}_\ell)\). The dynamics on interval
\([t_\ell,t_{\ell+1})\) are
\[
\boldsymbol{A}_\ell
=
\boldsymbol{S}
\boldsymbol{\Lambda}_\ell
\boldsymbol{S}^{-1},
\quad
\boldsymbol{\Lambda}_\ell
=
\operatorname{diag}
(
\lambda_{1,\ell},
\dots,
\lambda_{d,\ell}
).
\]
Since all intervals share the same eigenbasis, the transition from token
\(j\) to token \(i\) remains diagonalizable:
\[
\boldsymbol{\Phi}_{ij}
=
\prod_{\ell=j}^{i-1}
e^{\boldsymbol{A}_\ell \Delta t_{\ell+1,\ell}}
=
\boldsymbol{S}
\boldsymbol{\Gamma}_{ij}
\boldsymbol{S}^{-1}.
\]
This can be efficiently computed from cumulative damping and phase sums:
\[
M_i
=
\sum_{\ell<i}
\mu_\ell \Delta t_{\ell+1,\ell},
\qquad
W_{k,i}
=
\sum_{\ell<i}
\omega_{k,\ell}\Delta t_{\ell+1,\ell},
\]
\[
\Gamma_{k,ij}
=
\exp
\Big(
-(M_i-M_j)
+
i(W_{k,i}-W_{k,j})
\Big).
\]
Uncertainty can likewise accumulate along the content-dependent dynamics.
Let \(\boldsymbol{V}_{\ell j}\) denote the covariance accumulated from
\(t_j\) to \(t_\ell\), and \(\boldsymbol{Q}_{\ell+1}\) the covariance injected
over \([t_\ell,t_{\ell+1})\). The covariance obeys the discrete-time Lyapunov
recursion
\[
\boldsymbol{V}_{\ell+1,j}
=
\boldsymbol{\Phi}_{\ell+1,\ell}
\boldsymbol{V}_{\ell j}
\boldsymbol{\Phi}_{\ell+1,\ell}^{\dagger}
+
\boldsymbol{Q}_{\ell+1},
\qquad
\boldsymbol{V}_{jj}=\boldsymbol{0}.
\]
When the injected covariance is diagonal in the shared eigenbasis, with
entries \(q_{k,\ell}\), this reduces to independent scalar recursions. Let
\(D_{k,i}\) denote the covariance accumulated from the start of the sequence,
computed by the scan
\[
D_{k,i}
=
e^{-2\mu_{i-1}\Delta t_{i,i-1}}D_{k,i-1}+q_{k,i},
\qquad
D_{k,0}=0 .
\]
The covariance between any pair then follows as
\[
\lambda_{V,k,ij}
=
D_{k,i}-e^{-2(M_i-M_j)}D_{k,j},
\]
which satisfies \(\lambda_{V,k,jj}=0\) and the recursion above. The pairwise
factor is the squared transport decay, so both transport and uncertainty
accumulation are computed from a scan and elementwise operations involving
only decaying exponentials, and remain parallel over all pairs. For constant
\(\mu\) and \(q\), \(D_{k,i}\) saturates at \(\tilde\sigma^2\) and this
reduces to \(\tilde\sigma^2(1-\boldsymbol E^2)\).

\subsection{Stacked Attention Layers as an Unrolled Iterative State Estimator}
\label{sec:EM}

Each layer performs one step of iteratively reweighted least squares (IRLS)
on the robust objective $\sum_{j\le i}\rho(d_{ij}^2)$ of
Appendix~\ref{sec:appendix_robust_estimation}, and a stack of layers unrolls
the iteration. Working in the eigenbasis, let
$\hat{\boldsymbol{x}}_{s,i}^{(k)} :=
\boldsymbol{S}^{-1}\boldsymbol{C}^{-1}\bar{\boldsymbol{z}}_i^{(k)}$ denote the
current state estimate. Transported predictions are
\(
\hat{\boldsymbol{x}}_{s,ij}^{(k)}
=
e^{\boldsymbol{\Lambda}\Delta t_{ij}}
\hat{\boldsymbol{x}}_{s,j}^{(k)},
\)
with observable residual
\[
\boldsymbol r_{s,ij}^{(k)}
=
\hat{\boldsymbol{x}}_{s,i}^{(k)}
-
\hat{\boldsymbol{x}}_{s,ij}^{(k)}.
\]
The weights are evaluated at the current estimate,
\[
w_{ij}^{(k)}
=
\rho'\big(d_{ij}^{2,(k)}\big),
\qquad
d_{ij}^{2,(k)}
=
\boldsymbol r_{s,ij}^{(k)\dagger}
\boldsymbol P_{ij}
\boldsymbol r_{s,ij}^{(k)},
\]
which for the exponential influence function gives
\[
w_{ij}^{(k)}
=
\exp\!\left(
-\lambda_{P,ij}
\lvert\boldsymbol r_{s,ij}^{(k)}\rvert^2/d
\right).
\]
Holding these weights fixed, the weighted least-squares problem is solved in
closed form by the precision-weighted state estimate
\[
\bar{\boldsymbol{x}}_{s,i}^{(k)}
=
\bigg(
\sum_{j\le i}
w_{ij}^{(k)}\lambda_{P,ij}
\bigg)^{-1}
\sum_{j\le i}
w_{ij}^{(k)}\lambda_{P,ij}
\hat{\boldsymbol{x}}_{s,ij}^{(k)},
\]
where $\lambda_{P,ij}$ is the precision at lag $\Delta t_{ij}$
(Appendix~\ref{sec:Derivation}). The estimate is incorporated through the
innovation step, 
\[
\bar{\boldsymbol z}_i^{(k+1)}
=
\bar{\boldsymbol z}_i^{(k)}
+
\boldsymbol\alpha^{(k)}\odot
\boldsymbol C\boldsymbol S
\left(
\bar{\boldsymbol x}_{s,i}^{(k)}
-
\hat{\boldsymbol x}_{s,i}^{(k)}
\right),
\]
where $\boldsymbol\alpha^{(k)}\in(0,1]^d$ controls the correction step.
Residual connections implement partial state updates, allowing a stack of $L$
layers to perform $L$ successive refinements of the state estimate.

For the Student-$t$ penalty, the same iteration is also an EM procedure,
closely related to the EM formulation of probabilistic attention keys
\citep{nguyen2022improving}. There, $\rho$ is the negative log marginal
likelihood of the Gaussian scale mixture, and the weight
$\rho'(d_{ij}^2)=\mathbb E[s_{ij}^{-1}\mid d_{ij}^2]$ is the posterior mean
inverse scale, so the reweighting step is the E-step and the weighted
least-squares solution is the M-step. Unlike probabilistic attention keys,
where the latent variables are mixture-component assignments, the latent
variables here are residual scales, and the estimated quantity is the latent
dynamical state.

\newpage

\section{Experimental Details and Ablations}
\label{sec:Experimental_Details}

\subsection{Experimental Setup}

\paragraph{Architecture and Model Configuration.}
All experiments were conducted using a 6-layer decoder-only Transformer architecture. We set the model dimension to $d_{\text{model}} = 256$ with $h = 8$ attention heads. The attention mechanism maps the model dimension to a total latent dimension of $512$ via the $ d \times 2d$ query, key, and value projections (split into $d_h = 64$ per head), while the $ 2d \times d$ output projection maps back down to $256$.

We employ a Pre-Norm configuration using Layer Normalization. The Feed-Forward Network utilizes an expansion factor of 4. To optimize the parameter budget, we implement weight tying between the token embedding layer and the final linear output head. We use the GPT-2 byte-pair encoding (BPE) tokenizer with a vocabulary size of 50,257 for all language modeling experiments.

To ensure a fair comparison, RFA models and the baselines (RoPE and ALiBi) were designed with near-identical parameter counts. RFA introduces only a small set of scalar coefficients per head to parameterize noise variances ($\tilde{\sigma}^2, \eta^2, \gamma^2$) and robustness ($\nu, \beta_s$). Hence, the RFA models match the baseline parameter count ($19.36$M), with only a 0.02\% increase due to additional scalar coefficients.

\paragraph{Training and Optimization Protocol.}
Models were trained for 15 epochs using the Adam optimizer. We utilized a OneCycleLR scheduler with cosine annealing and a 5\% warmup period, and trained until convergence.

For RFA models, we adopted a decoupled optimization strategy to ensure the stability of the SDE coefficients: Feature weights use a peak LR of $1 \times 10^{-3}$ with $\beta_1=0.9$, while SDE coefficients use $5 \times 10^{-4}$ with $\beta_1=0.0$ and $\epsilon=10^{-7}$. We apply global gradient clipping at $1.0$, with a stricter $1 \times 10^{-4}$ threshold for RFA-specific parameters. These settings were used for all reported results; in later experiments, a
single learning rate for all parameters proved sufficient.

All models were trained on the WikiText-103 and BabyLM-2025 datasets using a standard causal language modeling objective. All results are from single training runs (seed 2025); in-window
perplexity varies by about $\pm0.1$ between runs.

\subsection{Model Variants and Ablation Design}
\label{sec:Model Variants and Ablation Design}

This section presents a series of ablations designed to isolate the contributions of RFA’s core components. The ablations consist of:

\begin{enumerate}
    \item Baseline models;
    \item RFA variants evaluated in Section~\ref{sec:Experiments} (M1–M3); and
    \item Structural diagnostic ablations (M2.1–M2.8), which progressively remove components of the filtering formulation to test necessity and failure modes. These models are not intended as competitive models, but rather as mechanistic probes of stability and extrapolation behavior.
\end{enumerate}

\paragraph{Baselines:}
\begin{itemize}
    \item \textbf{B1: Standard Transformer + RoPE.} Dot-product attention with rotary positional embeddings \citep{su2023roformerenhancedtransformerrotary}. Applies $d \rightarrow 2d \rightarrow d$ projections to match RFA parameterization; 32 frequencies per head, identical across heads.
    
    \item  \textbf{B2: Standard Transformer + ALiBi.}
    Dot-product attention with linear distance bias \citep{press2022trainshorttestlong}. Tests whether static geometric penalties are sufficient for stability. Applies $d \rightarrow 2d \rightarrow d$ projections to match RFA parameterization.
    
    \item  \textbf{B3: Decayed RoPE.}
    Identical to B1 but with an additional exponential decay applied to the normalized attention weights, as in RFA's value transport $(\boldsymbol{A} \odot \boldsymbol{E})$, with RFA's head-wise rates $\mu_h = b \cdot \omega_{h,\text{max}}$ $(b = 0.05)$, testing whether decay alone suffices in the absence of uncertainty modeling.
    
    \item  \textbf{B4: Spectrally Coupled RoPE (SC-RoPE).}
    Identical to B3 but with frequency-partitioned RoPE with head-wise decay schedules, testing whether decay with spectral coupling can recover SC-RFA’s stability; 256 frequencies partitioned across heads (32 per head). Together with M2.8, B4 serves as a spectrum-matched control for SC-RFA.
\end{itemize}

\paragraph{Primary RFA Models:}

\begin{itemize}
    \item \textbf{M1: Isotropic RFA.} Isotropic RFA as described in Algorithm~\ref{sec:Algorithm}, replacing the attention module in a standard Transformer. The first two heads are reserved with $\mu_h = 0$, and $\sigma_h^2$ is initialized to $0.1 \mu_h$.
    
    \item \textbf{M2: Spectrally Coupled RFA (SC-RFA), optimized for near-field performance.}
    Identical to M1 except with explicit coupling between rotation frequencies and decay rates, $\mu_h = b \cdot \omega_{h,\max}$, using light damping ($b=0.05$).
    
    \item \textbf{M3: Spectrally Coupled RFA (SC-RFA), optimized for extrapolation.}
    Identical to M2 but with stronger damping ($b=5.0$), and $\sigma_h^2$ initialized to $0.5 \mu_h$.
\end{itemize}

\paragraph{Structural Diagnostic Ablations:}
These ablations progressively remove components of the filtering formulation, starting from the full SC-RFA model (M2) and simplifying toward standard attention. Their purpose is to isolate which mechanisms are required for stable extrapolation.

\begin{itemize}
    
    \item \textbf{M2.1: Exponential Kernel.}
    M2 with Student's $t$ influence function replaced by an exponential weighting, i.e., $w_{ij} = \exp(-d_{ij}^2 / \nu_s d)$. This isolates the effect of heavy-tailed robust reweighting under the same dynamical precision kernel. 

    \item \textbf{M2.2: Flat Precision Kernel.}
    M2 with noise parameters removed so that $\boldsymbol{P}_{\Delta t}$ is constant across time lag. Tests whether dynamics alone suffice without uncertainty accumulation.

    \item \textbf{M2.3: No Multiplicative Gate.} M2 with the multiplicative gating term $\boldsymbol{P}_{\Delta t}$ set to a constant, to test the impact of the additive bias $\boldsymbol{B}_{\Delta t}$ in isolation.

    \item \textbf{M2.4: No Value Frame Alignment.}
    M2 without value rotation and counter-rotation, testing the necessity of aggregating in a shared temporal frame.

    \item \textbf{M2.5: No Rotational Dynamics.}
    M2 without rotations applied to queries, keys, or values, isolating the effect of decay-only dynamics.

    \item \textbf{M2.6: Unitary Dot-Product Limit.}
    No decay, no process or measurement noise, and no query and key normalization terms, so that
    attention weights reduce to normalized complex dot products between rotated embeddings. This yields a purely rotational positional encoding analogous
    to RoPER \citep{harik2022roper}.

    \item \textbf{M2.7: Minimal SC-RFA.}
    M2 with zero decay, the exponential weight, no multiplicative gate, no value rotation, and no query and key norm terms, but with the noise retained. The logits are then rotary dot products plus a learned per-head bias
    $-\log(1+\tfrac{\sigma^2}{\eta^2+\gamma^2}\Delta t)$, as in the logarithmic variant of
    KERPLE \citep{chi2022kerple}, testing whether SC-RFA's gains reduce to partitioned RoPE plus a learned distance bias.

    \item \textbf{M2.8: Rotary Limit.} M2.6 without value rotation: attention reduces to rotary dot-product attention with conjugate-paired frequencies and a learned temperature, i.e. RoPE implemented within RFA, using M2's spectrum.
    
\end{itemize}

\newpage

\section{Additional Experimental Results}
\label{sec:Additional_Results}

\subsection{Training Dynamics and Extrapolation Behavior}
\label{sec:training_dynamics}

\begin{figure}[H]
    \centering

    \begin{subfigure}{0.65\linewidth}
        \centering
        \includegraphics[width=\linewidth]{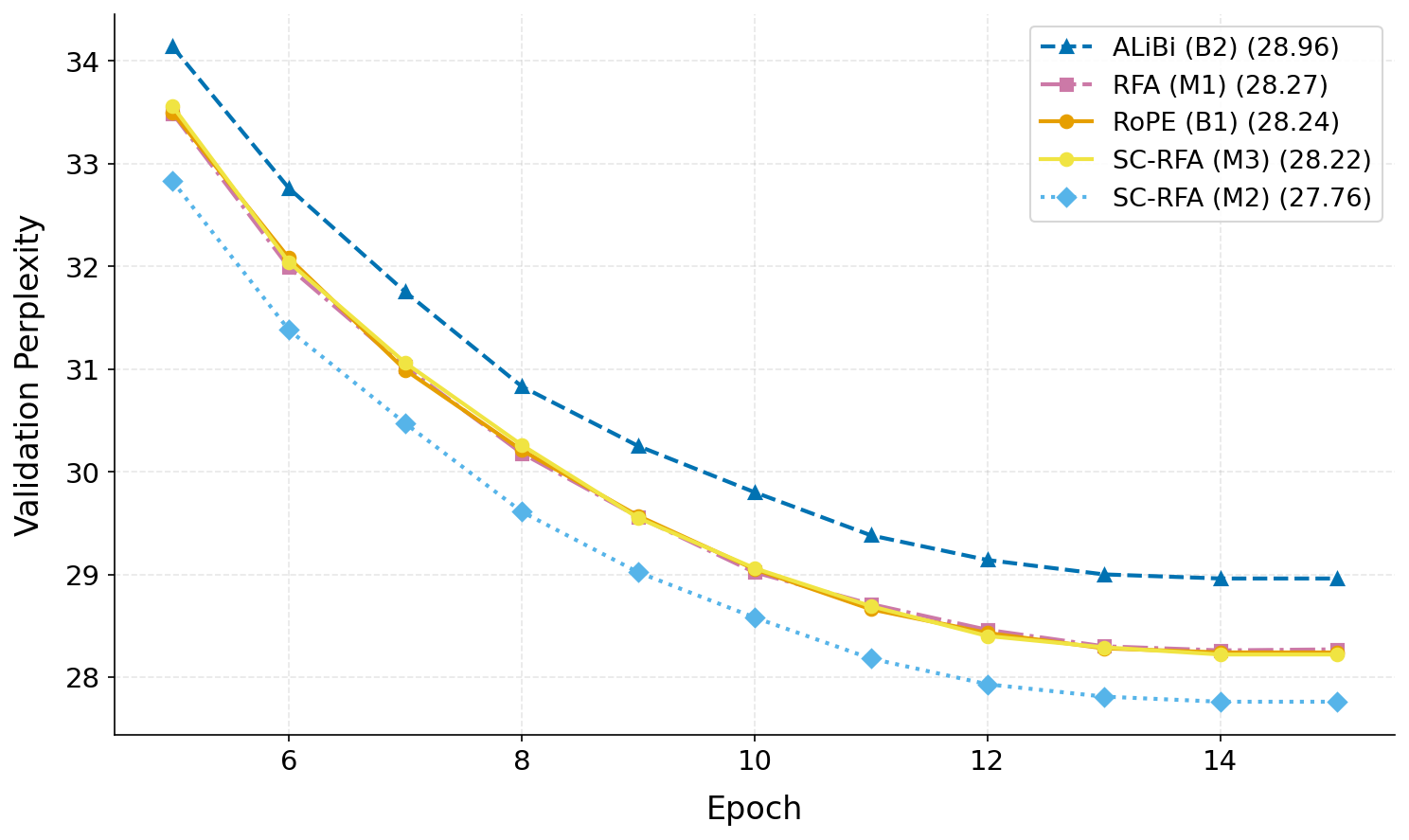}
        \caption{Validation perplexity over training (epochs 5–15).}
        \label{fig:training_andxtrapolation_a}
    \end{subfigure}

    \vspace{0.6em}

    \begin{subfigure}{0.65\linewidth}
        \centering
        \includegraphics[width=\linewidth]{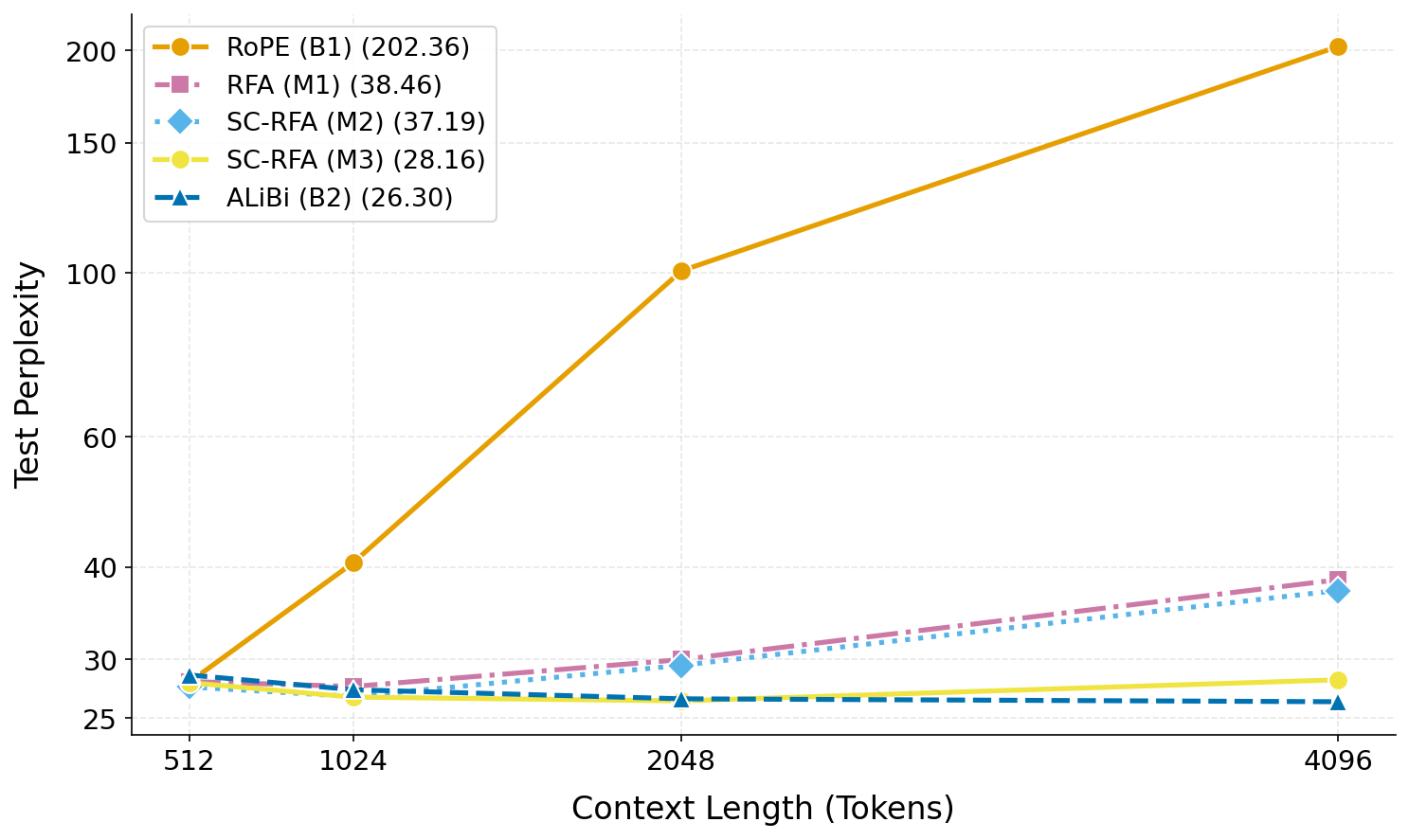}
        \caption{Test perplexity under length extrapolation beyond the training window ($512$ tokens).}
        \label{fig:training_andxtrapolation_b}
    \end{subfigure}

       \caption{\textbf{Training dynamics and length extrapolation on WikiText-103.}
   (a) SC-RFA (M2) stays below both baselines throughout training, while M1 and
   M3 track RoPE. (b) RFA variants degrade far more gradually than RoPE; M3
   extrapolates nearly as stably as ALiBi while matching RoPE in-window.}
    \label{fig:training_andxtrapolation}
\end{figure}

Figure~\ref{fig:training_andxtrapolation_a} tracks validation perplexity
during training. SC-RFA (M2) is below RoPE and ALiBi at every epoch, so its
in-window advantage holds throughout training rather than only at the final
checkpoint, while M1 and M3 closely track RoPE.

Figure~\ref{fig:training_andxtrapolation_b} shows test perplexity under
length extrapolation, corresponding to Table~\ref{tab:extrapolation_wikitext}.
All RFA variants degrade far more gradually than RoPE, and the damping
coefficient sets a trade-off between the two SC-RFA variants: M2 improves on
RoPE within the training window but degrades moderately beyond it, while M3
matches RoPE within the window and extrapolates nearly as stably as ALiBi,
which trails both within the window.

\begin{figure}[H]
    \centering
    \includegraphics[width=0.65 \linewidth]{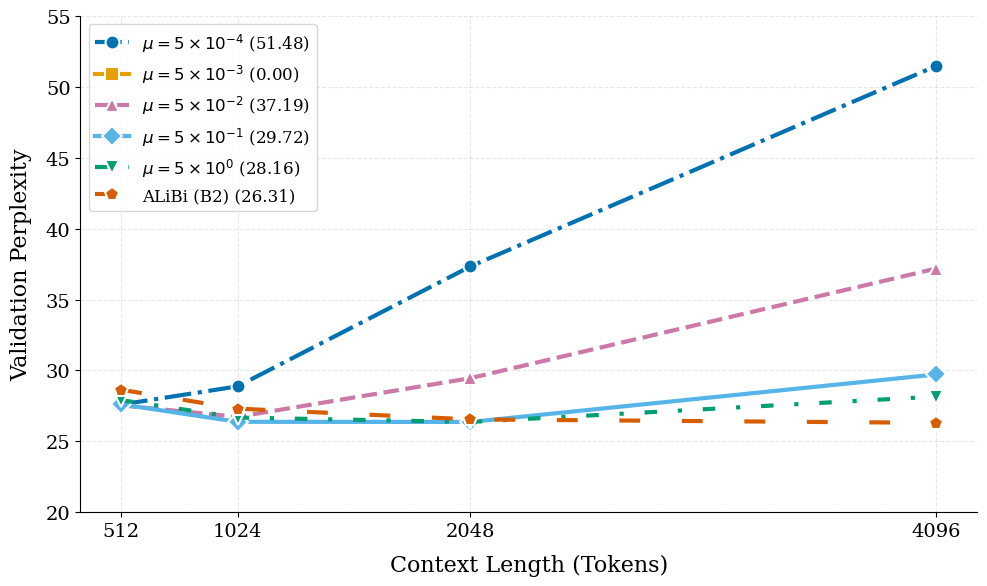}
    \caption{\textbf{Impact of the Damping Coefficient $b$ on Length Extrapolation.} Perplexity curves for varying $b$ demonstrate that higher damping coefficient values effectively stabilize long-range integration.}
    \label{fig:b_sweep}
\end{figure}

Figure~\ref{fig:b_sweep} shows the sensitivity analysis over damping values $b$ reported in Table~\ref{tab:b_sensitivity}. Increasing damping improves long-range stability by suppressing high-frequency propagation, but excessively large damping degrades short-range modeling.

\subsection{Parameter Dynamics in RFA}
\label{app:m1_dynamics}

\begin{figure*}[ht]
    \centering
    % Row 1: Query Noise
    \begin{minipage}{0.48\textwidth}
        \centering
        \includegraphics[width=\linewidth]{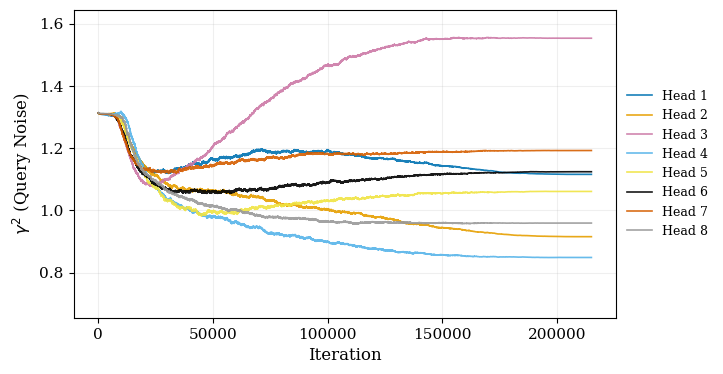}
        \centerline{(a) M1 Query Measurement Noise Variance ($\gamma^2$)}
    \end{minipage}
    \hfill
    \begin{minipage}{0.48\textwidth}
        \centering
        \includegraphics[width=\linewidth]{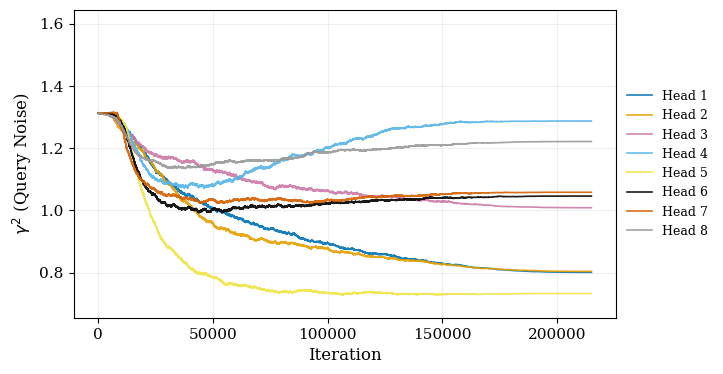}
        \centerline{(b) M2 Query Measurement Noise Variance ($\gamma^2$)}
    \end{minipage}

    \vspace{0.5cm}

    % Row 2: Key Noise
    \begin{minipage}{0.48\textwidth}
        \centering
        \includegraphics[width=\linewidth]{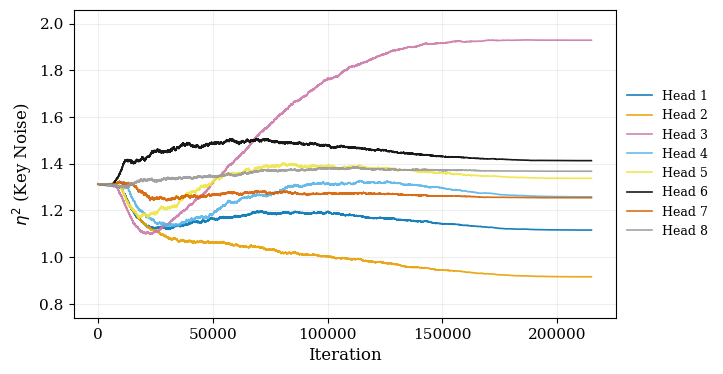}
        \centerline{(c) M1 Key Measurement Noise Variance ($\eta^2$)}
    \end{minipage}
    \hfill
    \begin{minipage}{0.48\textwidth}
        \centering
        \includegraphics[width=\linewidth]{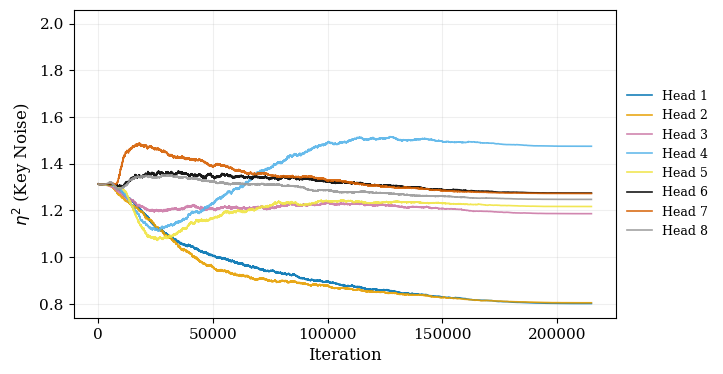}
        \centerline{(d) M2 Key Measurement Noise Variance ($\eta^2$)}
    \end{minipage}

    \vspace{0.5cm}

    % Row 3: Steady State Process Noise
    \begin{minipage}{0.48\textwidth}
        \centering
        \includegraphics[width=\linewidth]{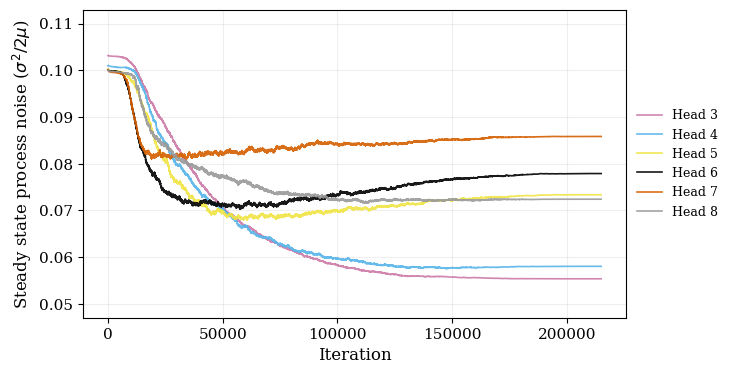}
        \centerline{(e) M1 Steady State Process Variance ($\sigma^2/2\mu$)}
    \end{minipage}
    \hfill
    \begin{minipage}{0.48\textwidth}
        \centering
        \includegraphics[width=\linewidth]{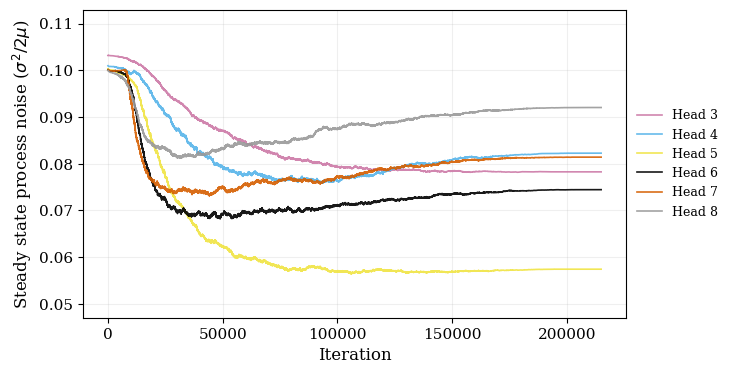}
        \centerline{(f) M2 Steady State Process Variance ($\sigma^2/2\mu$)}
    \end{minipage}

    \caption{\textbf{Measurement and Process Noise Parameters Comparison.} Query and key measurement noise variance and state process variance for M1 and M2, over the course of training. (Note that $\tilde{\sigma}^2$ is undefined for heads 1 and 2, with $\mu=0$.)}
    \label{fig:measurement_comparison}
\end{figure*}

\begin{figure*}[ht]
    \centering

    % Row 1: Robustness (nu/d)
    \begin{minipage}{0.48\textwidth}
        \centering
        \includegraphics[width=\linewidth]{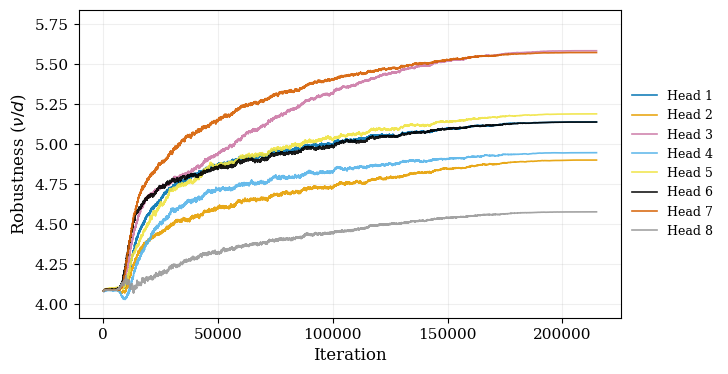}
        \centerline{(a) M1 Robustness Parameter ($\nu_s=\nu/d$)}
    \end{minipage}
    \hfill
    \begin{minipage}{0.48\textwidth}
        \centering
        \includegraphics[width=\linewidth]{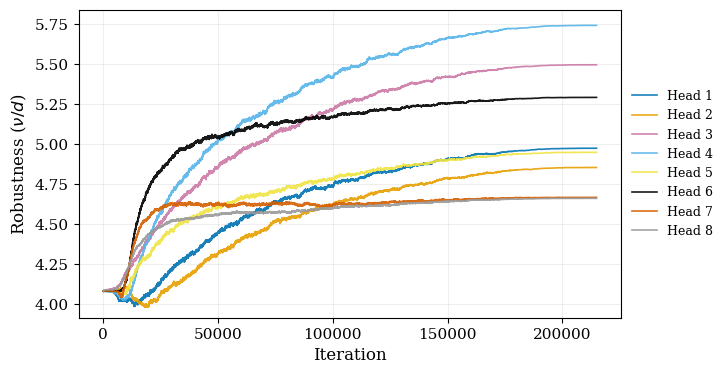}
        \centerline{(b) M2 Robustness Parameter ($\nu_s=\nu/d$)}
    \end{minipage}

    \vspace{0.6cm}

    % Row 2: Inverse Temperature (tau)
    \begin{minipage}{0.48\textwidth}
        \centering
        \includegraphics[width=\linewidth]{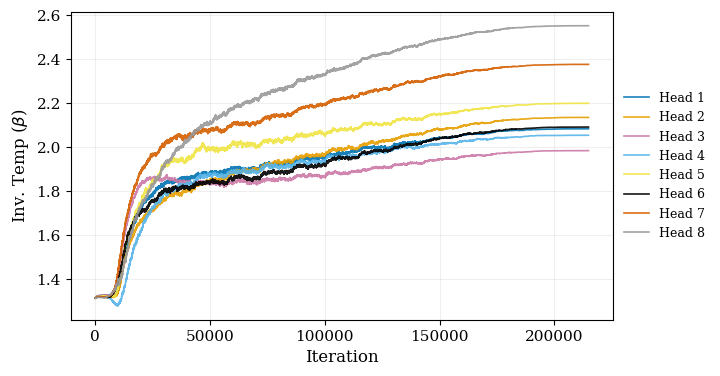}
        \centerline{ (c) M1 Inverse Temperature ($\beta_s$)}
    \end{minipage}
    \hfill
    \begin{minipage}{0.48\textwidth}
        \centering
        \includegraphics[width=\linewidth]{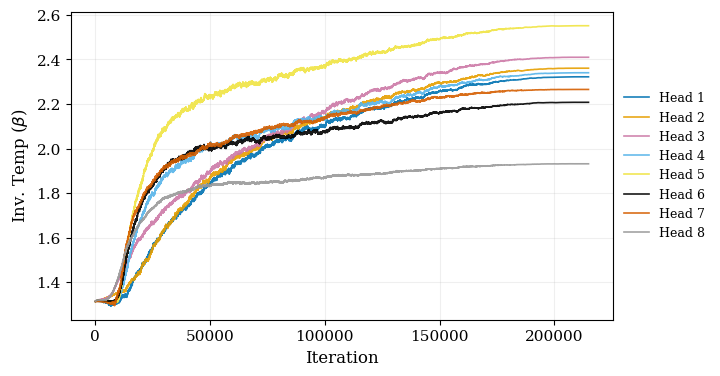}
        \centerline{(d) M2 Inverse Temperature ($\beta_s$)}
    \end{minipage}

    \caption{\textbf{Robustness and inverse temperature.} Robustness parameter and inverse temperature for M1 and M2, over the course of training.
    }
    \label{fig:stats_comparison}
\end{figure*}

Learned measurement and process noise parameters over the course of training are shown in Fig.~\ref{fig:measurement_comparison} for the last layer of both RFA (M1) and SC-RFA (M2). Distinct trajectories in query and key noise parameters indicate that different heads self-organize into separate signal-to-noise regimes. We plot robustness parameter $\nu_s = \nu/d$ and inverse temperature $\beta_s$ in Fig.~\ref{fig:stats_comparison}.

In general, lower-decay heads tend to converge to lower measurement noise variances $\eta^2, \gamma^2$, lower robustness parameter $\nu_s$, and higher inverse temperature $\beta_s$, consistent with stable long-range integration, while higher-decay heads tend to tolerate larger measurement noise.

Intermediate heads tend to converge to the highest measurement noise variance, lowest steady-state process uncertainty, and strongest robustness, consistent with modeling heterogeneous and noisy mid-range structure, while extreme short- and long-range heads tend to remain more tolerant to outliers.

The spectrally coupled model (M2, SC-RFA) exhibits lower average query and key noise variance and more clustered trajectories across heads.

\begin{figure}[ht]
    \centering

    \begin{minipage}{0.48\linewidth}
        \centering
        \includegraphics[width=\linewidth]{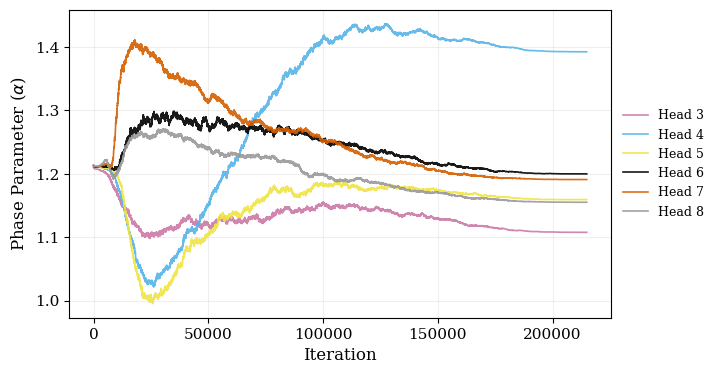}
        \centerline{(a) Phase parameter $\alpha$ by head (M2)}
    \end{minipage}
    \hfill
    \begin{minipage}{0.48\linewidth}
        \centering
        \includegraphics[width=\linewidth]{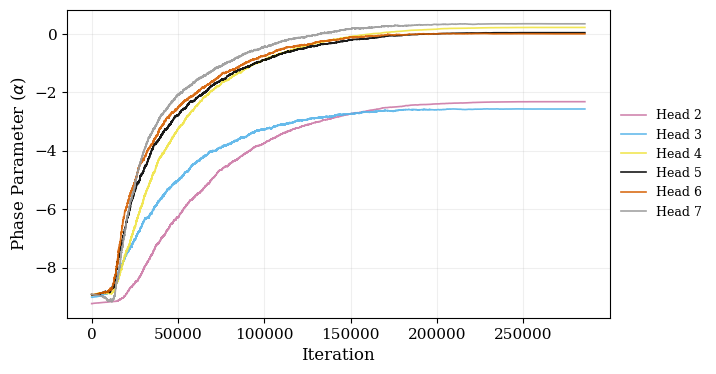}
        \centerline{ (b) Diffusive $\rightarrow$ integrative transition during training}
    \end{minipage}

    \caption{\textbf{Integrative dynamics in SC-RFA.}
    (a) Phase parameter ($\alpha$) under standard initialization, showing specialization across heads.
    (b) When initialized in the diffusive regime ($\alpha < 0$), most heads transitioned into the integrative regime ($\alpha > 0$) during training, while the two lowest-decay heads remained diffusive. (Note that $\alpha$ is undefined for heads 1 and 2, with $\mu=0$.)}
    \label{fig:diffusive_to_integrative}
\end{figure}

When initialized in the diffusive regime, we observed that higher-decay heads consistently transitioned into the integrative regime ($\alpha > 0$), while the lowest-decay heads remained diffusive
(Fig.~\ref{fig:diffusive_to_integrative}).

\clearpage

\subsection{Analysis of Attention Matrices}
\label{sec:attn_mats}

\begin{figure}[H]
    \centering
    \includegraphics[width=0.9\linewidth]{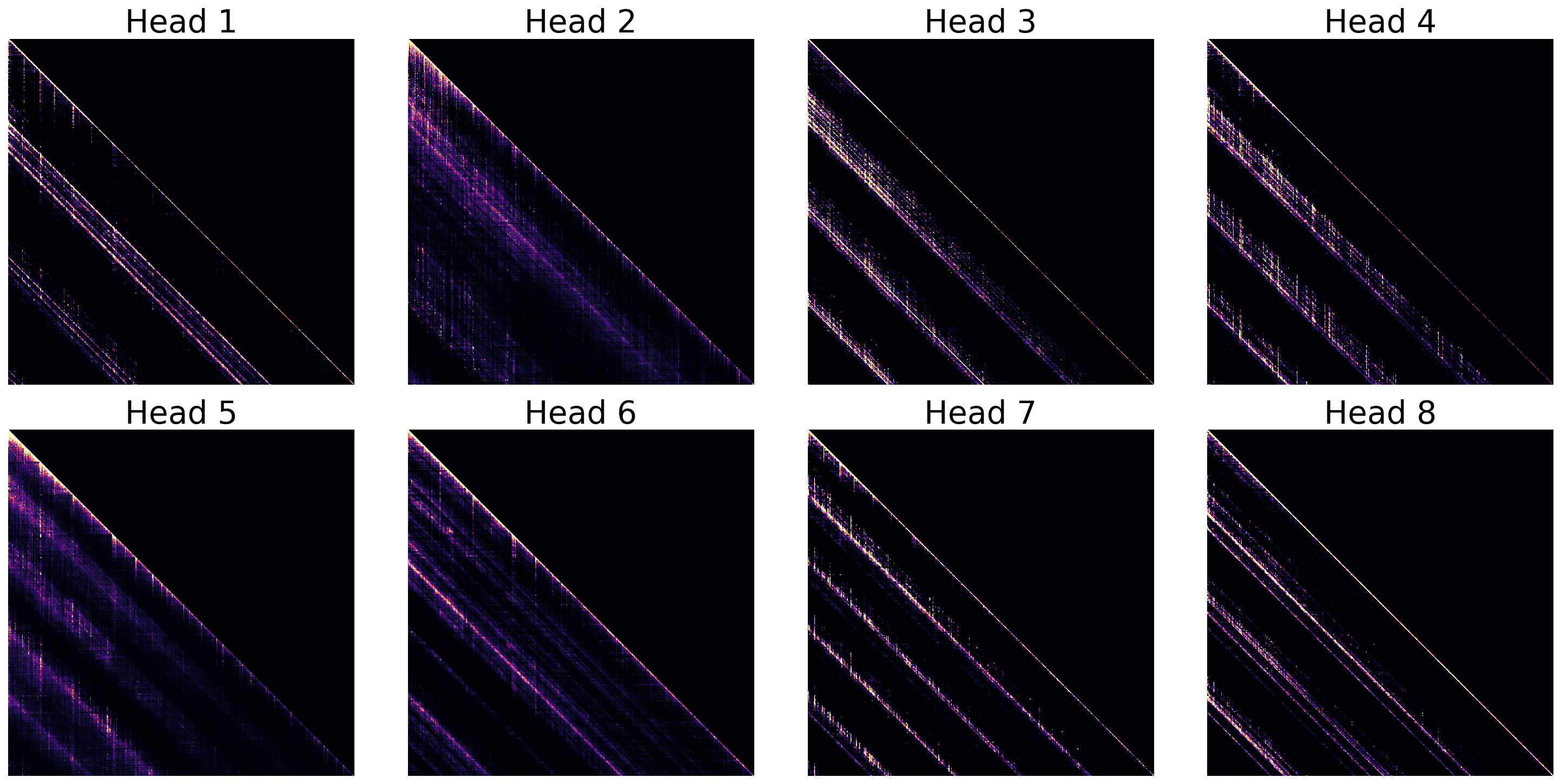}
    \caption{\textbf{Baseline RoPE Transformer (B1) at $L=4096$:} Attention concentrates on bands at fixed lags that persist across the full context, consistent with phase wrap-around of undamped rotations
(Section~\ref{sec:SC_RFA}).}
    \label{fig:attn_mats_B1}
\end{figure}

\begin{figure}[H]
    \centering
    \includegraphics[width=0.9\linewidth]{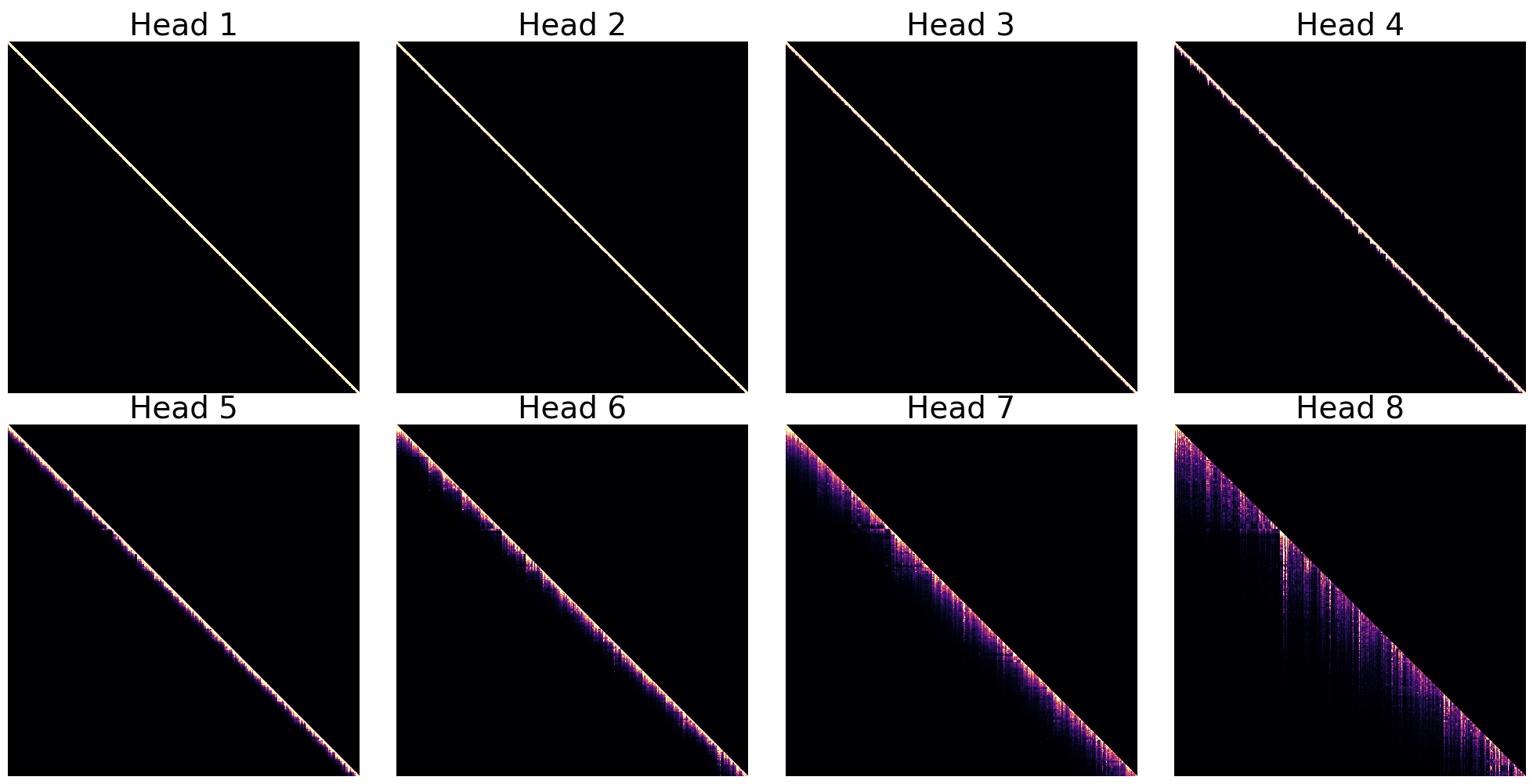}
    \caption{\textbf{ALiBi Transformer (B2) at $L=4096$:} Attention maps remain tightly localized to the diagonal across all heads, with only modest widening in higher heads. Long-range structure is suppressed rather than integrated.
    }
    \label{fig:attn_mats_B2}
\end{figure}

\begin{figure}[H]
    \centering
    \includegraphics[width=0.9\linewidth]{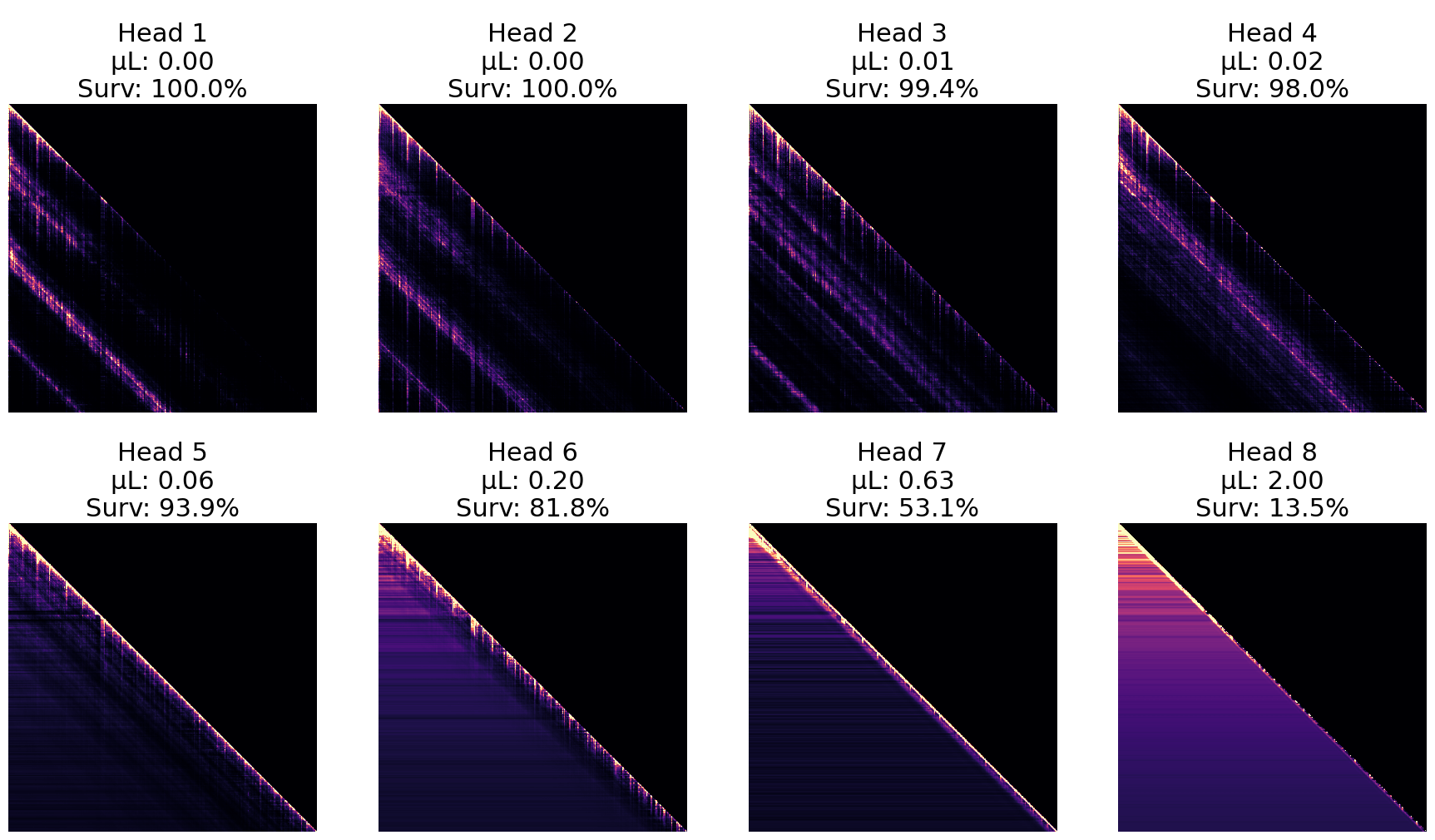}
    \caption{\textbf{Robust Filter Attention (M1) at $L=4096$:} Periodic bands are clearly visible. High-decay heads concentrate focus on the local diagonal, while low-decay heads exhibit the integrative regime: the bottom-right corner near the diagonal is suppressed as the model waits for the SDE dynamics to suppress initial measurement noise before assigning high precision to the state estimate.}
    \label{fig:attn_mats_M1}
\end{figure}

\begin{figure}[h]
\centering
\includegraphics[width=0.9\linewidth]{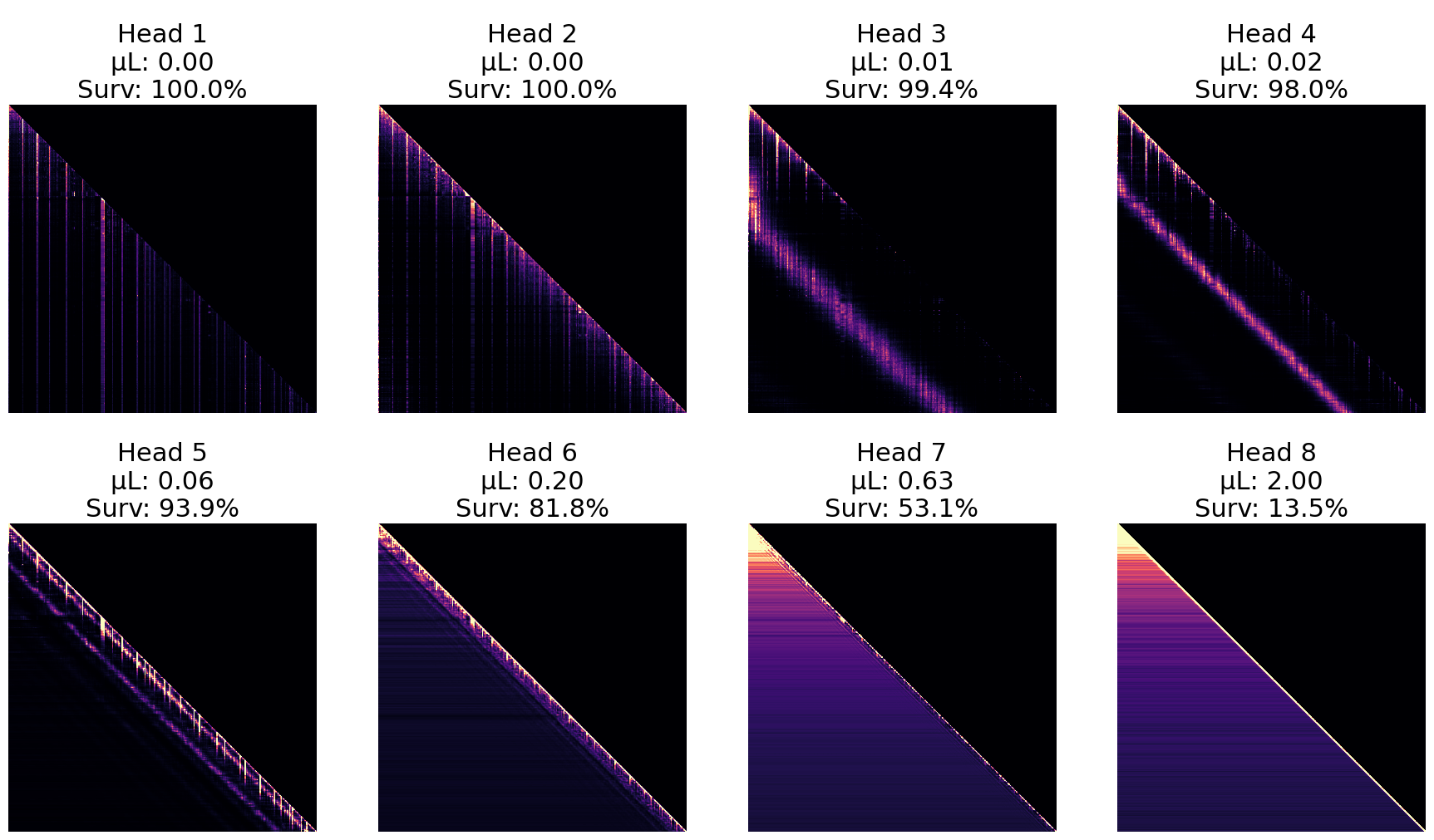}
\caption{\textbf{Spectrally-Coupled RFA (M2, $b=0.05$) at $L=4096$.} Frequency-dependent damping ($\mu_h = b \cdot \omega_{h,\text{max}}$) substantially alters long-range attention structure. SC-RFA has fewer periodic bands than RFA. Heads 3-5 each have only a single band, which become narrower and move closer to the diagonal as decay increases. Heads 1 and 2 act as stable long-range integrators.}
\label{fig:attn_mats_M2}
\end{figure}

We plot attention matrices at a context length of 4096 to visualize long-range behaviors induced by each positional prior: the baselines RoPE (B1) (Fig~\ref{fig:attn_mats_B1}) and ALiBi (B2) (Fig~\ref{fig:attn_mats_B2}); and the RFA (M1) (Fig~\ref{fig:attn_mats_M1}) and SC-RFA (M2) (Fig~\ref{fig:attn_mats_M2}) models. We use attention matrices from the last layer of each model.

\textbf{Note:} For the RFA models, for improved visualization, we plot the
attention matrix $\boldsymbol{A}$ before the decay is applied, rather than
$\hat{\boldsymbol{A}} := \boldsymbol{A}\odot\boldsymbol{E}$. In
$\hat{\boldsymbol{A}}$, bands at long lags are further attenuated.

\clearpage

The attention maps for RoPE show sharp bands at fixed lags that extend across
the entire context. In the absence of decay, a distant token whose phase
matches the query's remains an attention target at every offset.

In ALiBi, attention remains tightly localized to the diagonal across all heads, reflecting its fixed distance-based bias. While higher heads show modestly broader receptive fields, long-range context is suppressed rather than integrated.

RFA (M1) also produces periodic bands. Some heads exhibit an ``opening gate'' behavior, where attention is suppressed near the diagonal and peaks at a characteristic lag, indicating delayed aggregation until the state estimate stabilizes.

Spectral coupling in SC-RFA (M2) sharpens and organizes this structure. Coupling decay to frequency ($\mu_h = b \cdot \omega_{h,\text{max}}$) induces a specialization of heads to temporal lags: high-decay heads concentrate near the diagonal, while low-decay heads shift toward longer lags, forming distinct, narrow bands consistent with $\Delta t^* \propto 1/\mu_h$ (Sec.~\ref{sec:SC_RFA}). Decay alone, which falls monotonically with lag, cannot produce such a band;
its location is set by the precision rising with lag in the integrative
regime.

In SC-RFA, the first two (zero-decay) heads exhibit vertical structures corresponding to stable long-range retrieval. These heads effectively perform global key–value lookup rather than temporal filtering, assigning similar weight to salient tokens across all query positions and producing clean, vertically aligned patterns.

Together, these patterns support the view that RFA learns a structured multi-scale filtering behavior rather than relying on fixed geometric positional biases.

%%%%%%%%%%%%%%%%%%%%%%%%%%%%%%%%%%%%%%%%%%%%%%%%%%%%%%%%%%%%%%%%%%%%%%%%%%%%%%%
%%%%%%%%%%%%%%%%%%%%%%%%%%%%%%%%%%%%%%%%%%%%%%%%%%%%%%%%%%%%%%%%%%%%%%%%%%%%%%%

\end{document}